\documentclass[10pt]{article} % For LaTeX2e
\usepackage[accepted]{tmlr}
\usepackage{amsmath,amsfonts,bm}

\def\eqref#1{equation~\ref{#1}}
\def\1{\bm{1}}

\DeclareMathAlphabet{\mathsfit}{\encodingdefault}{\sfdefault}{m}{sl}
\SetMathAlphabet{\mathsfit}{bold}{\encodingdefault}{\sfdefault}{bx}{n}

\usepackage{hyperref}
\usepackage{url}
\hypersetup{hidelinks} 
\usepackage{booktabs}

\let\AND\relax
\usepackage{algorithm}
\usepackage{algorithmic}
\usepackage{natbib}
\usepackage[capitalize,noabbrev]{cleveref}
\usepackage[switch]{lineno}
\usepackage{nicefrac}       % compact symbols for 1/2, etc.
\usepackage{microtype}      % microtypography
\usepackage{xcolor}         % colors
\usepackage{multicol}
\usepackage{subcaption}
\usepackage{hyperref}
\usepackage{enumitem}
\usepackage{bm}
\usepackage{bbm}
\usepackage{comment}
\usepackage{amsmath}
\usepackage{amssymb}
\usepackage{mathtools}
\usepackage{amsthm}
\usepackage{makecell}
\usepackage[capitalize,noabbrev]{cleveref}

\theoremstyle{plain}
\newtheorem{theorem}{Theorem}

\newtheorem{lemma}{Lemma}

\theoremstyle{definition}
\newtheorem{definition}{Definition}
\newtheorem{assumption}{Assumption}

\theoremstyle{definition}
\newtheorem{remark}{Remark}

\title{Beyond Ordinary Lipschitz Constraints: Differentially Private Stochastic Optimization with Tsybakov Noise Condition }

\author{\name Difei Xu \email difei.xu@kaust.edu.sa \\
      \addr   Department of  Statistics\\
      King Abdullah University of Science and Technology \\
      \AND
      \name Meng Ding \email mengding@buffalo.edu \\
      \addr Department of Computer Sciecne\\
      State University of New York at Buffalo\\
      \AND
      \name Zihang Xiang \email zihang.xiang@kaust.edu.sa\\
      \addr Department of Computer Sciecne\\
      King Abdullah University of Science and Technology \\
      \AND
      \name Jinhui Xu \email jhxu@ustc.edu.cn \\
      \addr  School of Information Science and Technology\\
      University of Science and Technology of China \\
      \AND
      \name Di Wang\thanks{Corresponding author}  \email di.wang@kaust.edu.sa\\
      \addr   Department of Computer Science\\
      King Abdullah University of Science and Technology 
      }

\def\month{09}  % Insert correct month for camera-ready version
\def\year{2025} % Insert correct year for camera-ready version
\def\openreview{\url{https://openreview.net/forum?id=SZCygcrGng}} % Insert correct link to OpenReview for camera-ready version

\begin{document}

\maketitle

\begin{abstract}
We study Stochastic Convex Optimization in the Differential Privacy model (DP-SCO). Unlike previous studies, here we assume the population risk function satisfies
the Tsybakov Noise Condition (TNC) with some parameter $\theta>1$, where the Lipschitz constant of the loss could be extremely large or even unbounded, but the $\ell_2$-norm gradient of the loss has bounded $k$-th moment with $k\geq 2$. 
For the Lipschitz case with $\theta\geq 2$, we first propose an $(\varepsilon, \delta)$-DP algorithm whose utility bound is $\Tilde{O}\left(\left(\tilde{r}_{2k}(\frac{1}{\sqrt{n}}+(\frac{\sqrt{d}}{n\varepsilon}))^\frac{k-1}{k}\right)^\frac{\theta}{\theta-1}\right)$ in high probability, where $n$ is the sample size, $d$ is the model dimension, and $\tilde{r}_{2k}$ is a term that only depends on the $2k$-th moment of the gradient.  It is notable that such an upper bound is independent of the Lipschitz constant. We then extend to the case where 
 $\theta\geq \bar{\theta}> 1$ for some known constant $\bar{\theta}$. Moreover, when the privacy budget $\varepsilon$ is small enough, we show an upper bound of $\tilde{O}\left(\left(\tilde{r}_{k}(\frac{1}{\sqrt{n}}+(\frac{\sqrt{d}}{n\varepsilon}))^\frac{k-1}{k}\right)^\frac{\theta}{\theta-1}\right)$ even if the loss function is not Lipschitz. For the lower bound, we show that for any $\theta\geq 2$, the private minimax rate for $\rho$-zero Concentrated Differential Privacy is lower bounded by $\Omega\left(\left(\tilde{r}_{k}(\frac{1}{\sqrt{n}}+(\frac{\sqrt{d}}{n\sqrt{\rho}}))^\frac{k-1}{k}\right)^\frac{\theta}{\theta-1}\right)$. 
\end{abstract}

\section{Introduction}
% The application of machine learning in daily life is becoming increasingly common, with the volume of data used growing larger nowadays. However, these data manany also contain sensitive information. 
% To address the privacy issue, many current regulations such as GDPR, enforce machine learning algorithms to not only learn effectively from the training data but also provide a certain level of guarantee on
% privacy preservation. As a
% rigorous notion for statistical data privacy, differential privacy (DP) \citep{dwork2006calibrating} has received much attention
% in the past few years and has become a de facto technique
% for private data analysis.
Machine learning is increasingly being integrated into daily life, driven by an ever-growing volume of data. This data often includes sensitive information, which raises significant privacy concerns. In response, regulations such as the GDPR mandate that machine learning algorithms not only effectively extract insights from training data but also uphold stringent privacy standards. Differential privacy (DP) \citep{dwork2006calibrating}, a robust framework for ensuring statistical data privacy, has garnered substantial attention recently and has emerged as the leading methodology for conducting privacy-preserving data analysis.

% As one of the most fundamental problems in machine learning and differential privacy communities, DP Stochastic Convex Optimization (DP-SCO) and its empirical form, DP Empirical Risk Minimization (DP-ERM), have received great attention in the past decade starting from \citep{chaudhuri2011differentially},  such as \citep{bassily2014private, wang2017differentially,wang2019differentially,wu2017bolt,kasiviswanathan2016efficient,kifer2012private,smith2017interaction,wang2018empirical,wang2019noninteractive,asi2021private}. For example, \citep{bassily2019private} provides the near-optimal rates of DP-SCO with convex and strongly convex loss function. \citep{feldman2020private} provides algorithms with linear time complexity. \citep{su2023differentially} considers non-Euclidean spaces instead. 
Differential Privacy Stochastic Convex Optimization (DP-SCO) and its empirical form, DP Empirical Risk Minimization (DP-ERM), stand as core challenges within the machine learning and differential privacy communities. These methodologies have been the focus of significant research over the past decade, beginning with seminal works like those by Chaudhuri et al. \citep{chaudhuri2011differentially} and followed by numerous influential studies \citep{bassily2014private, wang2017differentially, wang2019differentially, wu2017bolt, kasiviswanathan2016efficient, kifer2012private, smith2017interaction, wang2018empirical, wang2019noninteractive, huai2020pairwise,xue2021differentially,asi2021private}. For instance, Bassily et al. \citep{bassily2019private} have provided near-optimal rates for DP-SCO across both convex and strongly convex loss functions. Feldman et al. \citep{feldman2020private} have developed algorithms that boast linear time complexity, and Su et al. \citep{su2023differentially} have expanded the discussion to non-Euclidean spaces.
% However, most of the existing theoretical results focus on the case where the loss function is $O(1)$-Lipschitz for all data, i.e., they need to assume either the underlying data distribution is bounded or sub-Gaussian. These assumptions are particularly important to 
% privacy guarantees for those output perturbation-based \citep{chaudhuri2011differentially} and objective or gradient perturbation-based 
% \citep{bassily2014private} DP methods.
% However, such assumptions may not always hold when dealing with real-world datasets, especially those from biomedicine and finance, which are often heavy-tailed \citep{woolson2011statistical,biswas2007statistical,ibragimov2015heavy}, implying that those algorithms may fail to guarantee the DP property. To fill in the gap, several recent works initiated the study of DP-SCO with heavy-tailed data, where the Lipschitz constant of the loss could be extremely large or unbounded \citep{wang2020differentially,kamath2022improved,hu2022high,lowy2023private,tao2022private}. Specifically, they assume that the gradient of loss has bounded $k$-th moment for some $k>0$, which is significantly weaker than the assumption that the loss is $O(1)$-Lipschitz, i.e., the gradient is uniformly lower bounded by some constant. 

However, the majority of existing theoretical frameworks primarily focus on scenarios where the loss function is \(O(1)\)-Lipschitz across all data, necessitating assumptions that the underlying data distribution is either bounded or sub-Gaussian. Such assumptions are crucial for the effectiveness of differential privacy methods based on output perturbation \citep{chaudhuri2011differentially} and objective or gradient perturbation \citep{bassily2014private}. Yet, these assumptions may not be valid for real-world datasets, particularly those from fields like biomedicine and finance, which are known to exhibit heavy-tailed distributions \citep{woolson2011statistical,biswas2007statistical,ibragimov2015heavy,chen2023quantizing}. This discrepancy can compromise the effectiveness of the algorithms in maintaining differential privacy. To bridge this gap, recent research has begun exploring DP-SCO in the context of heavy-tailed data, where the Lipschitz constant for the loss may be significantly higher or even unbounded \citep{wang2020differentially,kamath2022improved,hu2022high,lowy2023private,tao2022private,tian2025differentially,dingnearly,ding2024revisiting}. These studies typically assume that the gradient of the loss is bounded only in terms of its \(k\)-th moment for some \(k > 0\), a much less stringent requirement than \(O(1)\)-Lipschitz continuity.

% While DP-SCO with heavy-tailed data has been intensively studied, most of them focus on either general convex or strongly convex functions.  However, there are also many problems that are even stronger than strongly convex functions, or fall between convex and strongly convex functions. In the non-private counterpart, various studies have attempted to get faster rates by imposing additional assumptions on the loss functions. And it has been shown that it is indeed possible to achieve rates that are faster than the rates of general convex loss functions \citep{yang2018simple,koren2015fast,van2015fast}, or it could even achieve the same rate as in the strongly convex case even if the function is not strongly convex \citep{karimi2016linear,liu2018fast,xu2017stochastic}.  This is also true for the private case \citep{asi2021adapting}. Motivated by the above facts, one natural question is 

Although DP-SCO with heavy-tailed data has been extensively studied, most research has concentrated on general convex or strongly convex functions. Yet, numerous other problems exist that exceed the complexity of strongly convex functions or do not neatly fit within the convex-to-strongly convex spectrum. In non-private settings, several studies have managed to achieve faster convergence rates by introducing additional constraints on the loss functions. It has been demonstrated that it is possible to exceed the convergence rates of general convex functions \citep{yang2018simple,koren2015fast,van2015fast}, and some approaches have even matched the rates typical of strongly convex functions without the function actually being strongly convex \citep{karimi2016linear,liu2018fast,xu2017stochastic}. Similar advancements have been observed in the context of privacy-preserving algorithms \citep{asi2021adapting,su2022faster,su2024faster}. This leads to a compelling question:

% {\bf For the problem of DP-SCO with heavy-tailed data and special classes of population risk functions, is it possible to achieve faster rates of excess population risk than the optimal ones of general convex and (or) strongly convex cases? }

{\bf For the problem of DP-SCO with heavy-tailed data and special classes of population risk functions, is it possible to achieve faster rates of excess population risk than the optimal ones of general convex and (or) strongly convex cases? }

% In this paper, we provide an affirmative answer by studying some classes of population risk functions. Particularly, we  mainly focus on the case where the population risk function has a large or even infinite Lipschitz constant and satisfies the Tsybakov Noise Condition (TNC) \footnote{In some related work, it is also called the Error Bound Condition or the Growth Condition \citep{liu2018fast,xu2017stochastic}.}, which includes strongly convex functions, SVM, $\ell_1$-regularized stochastic optimization and linear regression with heavy-tailed data as special cases. Our contributions can be summarized as follows (see  Table \ref{tab:1} for details).  

In this paper, we affirmatively respond by examining certain classes of population risk functions. Specifically, we focus on the case where the population risk function possesses a large or potentially infinite Lipschitz constant and meets the Tsybakov Noise Condition (TNC) \footnote{This is also referred to as the Error Bound Condition or the Growth Condition in related literature \citep{liu2018fast,xu2017stochastic}.}, encompassing strongly convex functions, SVM, \(\ell_1\)-regularized stochastic optimization, and linear regression with heavy-tailed data as notable examples. 

Our contributions are detailed as follows (refer to Table \ref{tab:1} for details).
\begin{enumerate}
    \item We study DP-SCO where the population risk satisfies $(\theta, \lambda)$-TNC with $\theta>1$. Here, the loss function is $L_f$-Lipschitz, and the $k$-th moment of the loss gradient is small, where $L_f<\infty$ could be extremely large and $k\geq 2$.  Based on our newly developed localization method, we propose an $(\varepsilon, \delta)$-DP algorithm whose utility bound, with high probability, is $\Tilde{O}((\tilde{r}_{2k}(\frac{1}{\sqrt{n}}+(\frac{\sqrt{d}}{n\varepsilon}))^\frac{k-1}{k})^\frac{\theta}{\theta-1})$ when $\theta\geq 2$. Here, $n$ is the sample size, $d$ is the model dimension and $\tilde{r}_{2k}$ is a term that only depends on the $2k$-th moment of the gradient. It is notable that such an upper bound is independent of the Lipschitz constant.
    
    \item  We further relax the assumption that $\theta\geq 2$ to $\theta\geq \bar{\theta}>1$ for some known $\bar{\theta}$ and propose an algorithm that could achieve asymptotically the same bound as the previous one. Moreover, when the privacy budget $\varepsilon$ is small enough,  we show that even if the loss function is not Lipschitz, we can still get an upper bound of $\tilde{O}((\tilde{r}_{k}(\frac{1}{\sqrt{n}}+(\frac{\sqrt{d}}{n\varepsilon}))^\frac{k-1}{k})^\frac{\theta}{\theta-1})$. 
    
    \item On the lower bound side,  for any $\theta\geq 2$, we show that there exists a population risk function satisfying TNC with parameter $\theta$, whose  minimax population risk under $\rho$-zero Concentrated Differential Privacy  is always lower bounded by 
   $\Omega((\tilde{r}_{k}(\frac{1}{\sqrt{n}}+(\frac{\sqrt{d}}{n\sqrt{\rho}}))^\frac{k-1}{k})^\frac{\theta}{\theta-1})$. 
    
    %Finally, we investigate the case where $\theta=2$, i.e., the population risk function is strongly convex. Under the conditions where the loss function is non-negative and the optimal value of the population is small enough, we demonstrate that it is possible to achieve an upper bound of $O(\tilde{r}_{2k}(\frac{d}{n^2\varepsilon^2})^\frac{k-1}{k}+(\frac{1}{n^\tau})^\frac{k-1}{k})$ for any $\tau>1$, given that $n$ is sufficiently large. This result presents an improvement over the previously established upper bound. %Moreover, when the loss function is non-Lipschitz, we also develop an algorithm whose population risk is $O(\tilde{r}_{2k}(\frac{1}{n}+\frac{d}{n^2\varepsilon^2}))$, which matches our above bound with $k=2$. 
\end{enumerate}

\begin{table*}[!tb]
\centering 
\caption{Comparion with previous results on DP-SCO with different assumptions in $(\varepsilon, \delta)$-DP (we always assume the loss is smooth). All results omit the term of $\log \frac{1}{\delta}$, smoothness and strong convexity. $\dagger$ means the result is for $\rho$-zCDP.  $\star$ indicated the result holds when $\varepsilon=\tilde{O}(\frac{1}{n})$ \label{tab:1}.}
\resizebox{\textwidth}{!}{%
\begin{tabular}{lllp{4cm}}
% \hline
  & Upper Bound& Lower Bound & Assumption \\ [3ex] \hline
\citep{bassily2019private} & $O\left(\frac{1}{\sqrt{n}}+\frac{\sqrt{d}}{n\varepsilon}\right)$ & $\Omega\left(\frac{1}{\sqrt{n}}+\frac{\sqrt{d}}{n\varepsilon}\right)$ & $O(1)$-Lipschitz   \\ [2ex] \hline 
\citep{bassily2019private} & $O\left(\frac{1}{n}+\frac{d}{n^2\varepsilon^2}\right)$ & $\Omega\left(\frac{1}{\sqrt{n}}+\frac{d}{n^2\varepsilon^2}\right)$ & $O(1)$-Lipschitz   \\  [2ex]  \hline 
\citep{kamath2021improved} & $\tilde{O}\left(\frac{d}{\sqrt{n}}+\frac{d^2}{n\varepsilon}(\frac{\varepsilon n}{d^\frac{3}{2}})^\frac{1}{k}\right)$ & $\Omega(\sqrt{\frac{d}{n}}+\sqrt{d}(\frac{\sqrt{d}}{n\sqrt{\rho}})^\frac{k-1}{k})^\dagger $ &
$O(1)$-Lipschitz and bounded $k$-th moment ($k\geq 2$) \\  [2ex]  \hline 
\citep{kamath2021improved} & $\tilde{O}\left(\frac{d}{n}+d(\frac{\sqrt{d}}{n\varepsilon})^\frac{2(k-1)}{k}\right)$ & $\Omega\left(\frac{d}{n}+d(\frac{\sqrt{d}}{n\sqrt{\rho}})^\frac{2(k-1)}{k}\right)^\dagger $ &
$O(1)$-Lipschitz, strongly convex and bounded $k$-th moment ($k\geq 2$) \\  [2ex]  \hline 
\citep{asi2021private,su2022faster}    & $\tilde{O}\left(\left(\frac{1}{\sqrt{n}}+\frac{\sqrt{d}}{n\varepsilon}\right)^{\frac{\theta}{\theta-1}}\right)$ & $\Omega \left(\left(\frac{1}{\sqrt{n}}+\frac{\sqrt{d}}{n\varepsilon}\right)^{\frac{\theta}{\theta-1}}\right)$  when $\theta\geq 2$  & $O(1)$-Lipschitz under TNC with $\theta>1 $         
 \\  [2ex]  \hline
\citep{lowy2023private}  & $O\left(\widetilde{R}_{2 k, n} \left(\frac{1}{\sqrt{n}}+\left(\frac{\sqrt{d }}{\varepsilon n}\right)^{\frac{k-1}{k}}\right)\right)$                             & $\Omega\left(\widetilde{r}_k \left(\frac{1}{\sqrt{n}}+\left(\frac{\sqrt{d}}{\sqrt{\rho} n}\right)^{\frac{k-1}{k}} \right)\right)^\dagger$        & (large) Lipschitz,  bounded $k$-th moment ($k\geq 2$)         \\ \hline
\citep{lowy2023private}  & $\tilde{O}\left(\widetilde{r}_k \left[\frac{1}{\sqrt{n}}+\max \left\{\left(\left(\frac{1}{\widetilde{r}_k}\right)^{1 / 4} \frac{\sqrt{d}}{\varepsilon n}\right)^{\frac{4(k-1)}{5 k-1}},\left(\frac{\sqrt{d}}{\varepsilon n}\right)^{\frac{k-1}{k}}\right\}\right]\right)$       & $\Omega\left(\widetilde{r}_k \left(\frac{1}{\sqrt{n}}+\left(\frac{\sqrt{d}}{\sqrt{\rho} n}\right)^{\frac{k-1}{k}} \right)\right)^\dagger$     & bounded $k$-th moment ($k\geq 2$)    \\ \hline

\citep{lowy2023private}  & $\tilde{O}\left(\widetilde{R}^2_{2 k, n} \left(\frac{1}{n}+\left(\frac{\sqrt{d }}{\varepsilon n}\right)^{\frac{2(k-1)}{k}}\right)\right)$                          & $\Omega\left(\widetilde{r}_k^2 \left(\frac{1}{n}+\left(\frac{\sqrt{d}}{\sqrt{\rho} n}\right)^{\frac{2(k-1)}{k}} \right)\right)^\dagger$        &  strongly convex, bounded $k$-th moment ($k\geq 2$)         \\ \hline
This paper & $\tilde{O}\left(\left(\widetilde{R}_{2k,n}( \frac{1}{\sqrt[]{n}}+\left(\frac{\sqrt[]{d}}{\varepsilon  n}\right)^{\frac{k-1}{k}})\right)^{\frac{\theta}{\theta-1}}\right)$                                                   &  $\Omega \left(\left(\tilde{r}_k( \frac{1}{\sqrt[]{n}}+\left(\frac{\sqrt[]{d}}{\varepsilon  n}\right)^{\frac{k-1}{k}})\right)^{\frac{\theta}{\theta-1}}\right)$   when $\theta\geq 2$& (large) Lipschitz function under TNC with $\theta>1$              \\ \hline
This paper & $\tilde{O}\left({\widetilde{r}_k^{\frac{\theta }{\theta-1}}}\left(\frac{1}{\sqrt[]{n}}+\left(\frac{\sqrt[]{d }}{\varepsilon  n}\right)^{\frac{k-1}{k}}\right)^{\frac{\theta}{\theta-1}}\right)^\star$ & $\Omega \left(\left(\tilde{r}_k( \frac{1}{\sqrt[]{n}}+\left(\frac{\sqrt[]{d}}{\varepsilon  n}\right)^{\frac{k-1}{k}})\right)^{\frac{\theta}{\theta-1}}\right)$   when $\theta\geq 2$ &  TNC with $\theta>1$ \\ \hline

\end{tabular}
}%

\end{table*}

\section{Related Work}
\noindent {\bf DP-SCO with Heavy-tailed Data.} As we mentioned previously, there is a long list of work for DP-SCO from various perspectives. Here we only focus on the work related to DP-SCO with heavy-tailed data. Generally speaking, there are two ways of modeling heavy-tailedness: The first one considers each coordinate of the loss gradient has bounded moments, while the second one assumes the norm of the loss gradient has bounded moments, which is stronger than the first one.  For the first direction, \citep{wang2020differentially} provides the first study under the assumption of bounded $k$-th moment ($k\geq 2$) and proposes three different ways for both convex and strongly convex loss. The bounds were later improved by  \citep{kamath2021improved}. Specifically, \citep{kamath2021improved} provides improved upper bounds for convex loss and optimal rates for strongly convex loss. Later, there are some works that consider different extensions. For example, \citep{hu2022high} extends to the high-dimensional and polyhedral cases, \citep{tao2022private} extends to the case where the gradient only has $(1+v)$-th moment with $v\in (0, 1]$, and \citep{wang2022differentially,wang2025private} considers the $\ell_1$-regression. For the second direction, \citep{lowy2023private} provides a comprehensive study for both convex and strongly convex loss. In detail, for Lipschitz loss whose gradient has $k$-th moment, they provide upper bounds that are independent of the Lipschitz constant. In contrast to the work of \citep{lowy2023private}, our approach builds upon Theorem 6 from \citep{lowy2023private}. We then derive the high-probability upper bound and extend these results to population risks satisfying the Tsybakov Noise Condition (TNC) in subsequent sections. Notably, when $\theta=2$, our findings align with their results for strongly convex losses, while simultaneously generalizing their underlying assumptions. Moreover, the results in  \citep{lowy2023private} are in expectation form while we provide new algorithms, and our results are in the high probability form. What is notable is that \citep{asi2024private} proposes a reduction-based method to deal with data with heavy-tailed gradients, deriving  results  analogous to ours. Nevertheless, we apply different techniques under distinct assumptions. Specifically, their framework relies on a uniform Lipschitz assumption, which imposes more restrictive conditions compared to our method. Furthermore, our analysis extends to TNC, and subsequent theoretical investigation demonstrates that the Lipschitz requirement can be entirely eliminated under certain conditions.

\noindent {\bf DP for Heavy-tailed Data.} In addition to DP-SCO, there is also some work on DP for heavy-tailed data.  \citep{barber2014privacy} provided the first study on private mean estimation for distributions with a bounded moment, which has been extended by \citep{kamath2020private,liu2021robust,brunel2020propose} recently. However, these methods cannot be applied to our problem as these results are all in the expectation form. Motivated by \citep{wang2020differentially}, we later consider statistical guarantees of DP Expectation Maximization and apply them to the Gaussian Mixture Model. \citep{wu2023differentially,tao2022optimal,wu2024private} considers private reinforcement learning and bandit learning where the reward follows a heavy-tailed distribution.  However, since the reward is a scalar, these methods are not applicable to our problem. 

\noindent {\bf Loss functions with TNC.} While most of this paper focuses on loss functions that are either convex or strongly convex, many loss functions fall between these two categories. That is, they are not strongly convex, but their statistical rate is better than purely convex losses. For TNC and Lipschitz loss functions, the best-known current rate is $O((\frac{1}{\sqrt{n}})^\frac{\theta}{\theta-1})$ \citep{liu2018fast}, which corresponds to the first term in our upper bounds. In Theorem 5, we demonstrate that this upper bound is tight for $\theta \geq 2$. A comparison to the non-private setting will be included in the final version of the paper.

Our methods introduce novel technical challenges compared to non-private approaches. The key innovation lies in our analysis, which is based on algorithmic stability and a newly developed localized and clipped algorithm (Algorithm 3), which has not been previously studied. Specifically, Algorithm 4 is inspired by Algorithm 2 in \citep{liu2018fast}. However, while the base algorithm in \citep{liu2018fast} is a simple averaged version of projected SGD, our Algorithm 3 is significantly more complex. One major technical challenge is that Algorithm 2 in \citep{liu2018fast} assumes a Lipschitz loss function with a fixed Lipschitz constant. Consequently, their bounds rely on this constant. In contrast, we address scenarios where the Lipschitz parameter can be extremely large. Therefore, we developed a new base algorithm that removes dependence on this parameter and instead utilizes moments.

\section{Preliminaries}
	\begin{definition}[Differential Privacy \citep{dwork2006calibrating}]\label{def:3.1}
	Given a data universe $\mathcal{X}$, we say that two datasets $S,S'\subseteq \mathcal{X}$ are neighbors if they differ by only one entry, which is denoted as $S \sim S'$. A randomized algorithm $\mathcal{A}$ is $(\varepsilon,\delta)$-differentially private (DP) if for all neighboring datasets $S,S'$ and for all events $E$ in the output space of $\mathcal{A}$, the following holds
	$$\mathbb{P}(\mathcal{A}(S)\in E)\leqslant e^{\varepsilon} \mathbb{P}(\mathcal{A}(S')\in E)+\delta.$$ 
 If $\delta=0$, we call algorithm $\mathcal{A}$ $\varepsilon$-DP. 
\end{definition}

\begin{definition}[zCDP \citep{bun2016concentrated}]
    A randomized algorithm $\mathcal{A}$ is $\rho$-zero-concentrate-differentially private (zCDP) if for all neighboring datasets $S,S'$ and $\alpha\in (1, \infty)$, we have $D_\alpha(\mathcal{A}(S)\| \mathcal{A}(S'))\leqslant \rho \alpha $, where  $D_\alpha$ is the $\alpha$-R\'{e}nyi divergence between $\mathcal{A}(S)$ and $\mathcal{A}(S')$.
\end{definition}

\begin{remark}
    In this paper, we  focus on $(\varepsilon, \delta)$-DP for upper bounds and $\rho$-zCDP for lower bounds, and we mainly use the Gaussian mechanism to guarantee the DP property. For Algorithms 1-5, which are based on stability analysis and the Gaussian mechanism, they operate as one-pass algorithms without sub-sampling. As a result, they can easily meet the requirements for CDP. However, a challenge arises with Algorithm 6. In this case, we employ privacy amplification via shuffling to reduce the noise. Currently, privacy amplification via shuffling is only applicable to $\varepsilon$ and $(\varepsilon, \delta)$-LDP, and no version exists for zCDP. To maintain consistency throughout the paper, we use $(\varepsilon, \delta)$-DP for all our upper bounds.
\end{remark}
	\begin{definition}[Gaussian Mechanism]
	Given any function $q: \mathcal{X}^n\rightarrow \mathbb{R}^d$, the Gaussian mechanism is defined as  $q(S)+\xi$ where $\xi\sim \mathcal{N}(0,\frac{16\Delta^2_2(q)\log(1/\delta)}{\varepsilon^2}\mathbb{I}_d)$,  where $\Delta_2(q)$ is the $\ell_2$-sensitivity of the function $q$,
{\em i.e.,}
		$\Delta_2(q)=\sup_{S\sim S'}\|q(S)-q(S')\|_2.$	Gaussian mechanism preserves $(\varepsilon, \delta)$-DP for $0<\varepsilon, \delta\leqslant 1$.
	\end{definition}
 	\begin{definition}[DP-SCO \citep{bassily2014private}]\label{def:3}
		Given a dataset $S=\{x_1,\cdots,x_n\}$ from a data universe $\mathcal{X}$ where $x_i$ are i.i.d. samples from some unknown distribution $\mathcal{D}$, a convex loss function $f(\cdot, \cdot)$, and a convex constraint set  $\mathcal{W} \subseteq \mathbb{R}^d$, Differentially Private Stochastic Convex Optimization (DP-SCO) is to find $w^{\text{priv}}$ so as to minimize the population risk, {\em i.e.,} $F (w)=\mathbb{E}_{x\sim \mathcal{D}}[f(w, x)]$
		with the guarantee of being differentially private.
		 The utility of the algorithm is measured by the \textit{(expected) excess population risk}, that is  $$\mathbb{E}_{\mathcal{A}}[F (w^{\text{priv}})]-\min_{w\in \mathbb{\mathcal{W}}}F(w),$$
where the expectation of $\mathcal{A}$ is taken over all the randomness of the algorithm. Besides 
%Instead of 
the population risk, we may also measure the \textit{empirical risk} of dataset $S$: $\bar{F}(w, S)=\frac{1}{n}\sum_{i=1}^n f(w, x_i).$
	\end{definition} 
 	\begin{definition}[Lipschitz]\label{def:4}
		A function $f:\mathcal{W}\mapsto \mathbb{R}$ is L-Lipschitz over the domain $\mathcal{W}$ if for all $w,w^{\prime}\in \mathcal{W}$,
		$|f(w)-f(w^{\prime})|\leqslant L\|w-w^{\prime}\|_2.$
	\end{definition}
	\begin{definition}[Smoothness]\label{def:5}
		A function $f:\mathcal{W}\mapsto \mathbb{R}$ is $\beta$-smooth over the domain $\mathcal{W}$ if for all $w,w' \in \mathcal{W}$, 
		$f(w)\leqslant f(w')+\langle \nabla f(w'), w-w'\rangle +\frac{\beta}{2}\|w-w'\|_2^2.$
	\end{definition}
	\begin{definition}[Strongly Convex]\label{def:6}
		A function $F:\mathcal{W}\mapsto \mathbb{R}$ is $\lambda$-strongly convex over the domain $\mathcal{W}$ if, for all $w,w^{\prime}\in \mathcal{W}$, $
		F(w)+\langle\nabla F(w),w^{\prime}-w\rangle+\frac{\lambda}{2}\|w^{\prime}-w\|_2^2\leqslant F(w^{\prime}). 
		$
	\end{definition} 
     Previous work on DP-SCO only focused on cases where the loss function is either convex or strongly convex \citep{bassily2019private,feldman2020private}. In this paper, we mainly examine the case where the population risk satisfies the Tsybakov Noise Condition (TNC) \citep{ramdas2012optimal,liu2018fast,ramdas2013algorithmic}, which has been extensively studied and has been shown that it could achieve faster rates than the optimal one of general convex loss functions in the non-private case.  Below, we introduce the definition of TNC. 
     \begin{definition}[Tsybakov Noise Condition]\label{def:8}
         For a convex function $F(\cdot)$, let $\mathcal{W}_*=\arg\min_{w\in \mathcal{W}} F(w)$ denote the optimal set and for any $w\in \mathcal{W}$, let $w^*=\arg\min_{u\in \mathcal{W}_*}\|u-w\|_2$ denote the projection of $w$ onto the optimal set $\mathcal{W}_*$. 
        The function $F$ satisfies $(\theta, \lambda)$-TNC for some $\theta>1 $ and $\lambda>0$ if, for any $w\in \mathcal{W}$, the following inequality holds:
        \begin{equation}\label{eq:2}
		F(w)- F(w^*)\geq \lambda\|w-w^*\|_2^{\theta}. 
		\end{equation}
     \end{definition}
   From the definition of TNC and Definition \ref{def:6}, we can see that a $\lambda$-strong convex function is $(2, \frac{\lambda}{2})$-TNC. In Examples after Theorem 2, we deomenstrate how TNC can be verified in practical situations. Moreover, if a function is $(\theta, \lambda)$-TNC, then it is also $(\theta', \lambda)$-TNC for any $\theta<\theta' 
    $.  Throughout the paper, we assume that $\theta$ is a constant and thus we omit the term of $c^{\theta}$ in the Big-$O$ notation if $c$ is a constant.

%For a TNC function, we have the following property. 
	\begin{lemma}[Lemma 2 in \citep{ramdas2012optimal}]\label{lemma:1}
	If the function $F(\cdot)$ is $(\theta, \lambda)$-TNC and $L_f$-Lipschitz, then we have $\|w-w^*\|_2\leqslant(L_f\lambda^{-1})^{\frac{1}{\theta-1}}$  and $F(w)-F(w^*)\leqslant (L_f^{\theta}\lambda^{-1})^{\frac{1}{\theta-1}}$ for all $ w\in\mathcal{W}$, where $w^*$ is defined as in Definition \ref{def:8}.
	\end{lemma} 
As mentioned earlier, our primary focus here is on cases where the loss function's Lipschitz constant is sufficiently large or even infinite. In such cases, we may seek alternative terms to replace the Lipschitz constant. Motivated by previous work on DP-SCO with heavy-tailed gradients, we consider the moments of the gradient. Specifically, we assume that the stochastic gradient distributions have bounded $k$-th moment for some $k \geqslant 2$:
\begin{assumption}\label{1}
     There exists $k \geqslant 2$ and $\widetilde{r}^{(k)}>0$ such that $\mathbb{E}\left[\sup _{w \in \mathcal{W}}\|\nabla f(w, x)\|_2^k\right] \leqslant \widetilde{r}^{(k)}$, where $\widetilde{r}_k:=\left(\widetilde{r}^{(k)}\right)^{1 / k}$. Moreover, we assume the constrained set $\mathcal{W}$ is bounded with diameter $D$. 
\end{assumption}
\iffalse 
\begin{assumption}
    There exists $k \geqslant 2$, and $r^{(k)}>0$ such that $\sup _{w\in \mathcal{W}}\mathbb{E}\left\| \nabla f(w,x) \right\| _2^k \leqslant r^{(k)}$, for all gradients $\nabla f(w,x_i) \in \partial _wf(w,x_i)$. Denote $r_k:=(r^{(k)})^{1/k}$  
\end{assumption}
\fi 
If the loss function is $L_f$-Lipschitz, we can always observe that  $\widetilde{r}_k \leqslant L_f=\sup _{w, x}\|\nabla f(w, x)\|_2$. Moreover, $\widetilde{r}_k$ could be far less than the Lipschitz constant.

To state our subsequent theoretical results more clearly, we introduce some additional notations. For a batch of data $X \in \mathcal{X}^m$, we define the $k$-th empirical moment of $f(w,\cdot )$, by
\begin{equation}\nonumber
    \widehat r_m(X)^{(k)}=\sup\limits_{w\in\mathcal{W}} \frac1m\sum\limits_{i=1}^m\|\nabla f(w,x_i)\|_2^k. 
\end{equation}
For $X \sim \mathcal{D}^m$, we denote the $k$-th expected empirical moment by
\begin{equation}\nonumber
    \widetilde{e}_m^{(k)}:=\mathbb{E}[\widehat{r}_m(X)^{(k)}],
\end{equation}
and let 
\begin{equation}\nonumber
    \widetilde{r}_{k,m}:=(\widetilde{e}_m^{(k)})^{1/k}.
\end{equation}
Note that $\widetilde{r}_{k,1}=\widetilde{r}_k.$
We define $\widetilde{R}_{k,n}:=\sqrt{\sum_{i=1}^l2^{-i}\widetilde{r}_{k,n_i}^2}$,  where $n_i=2^{-i}n$ and $l=\log_{2} n $. Actually, $\widetilde{R}_{k,n}$, a weighted average of the expected empirical moments for distinct batch sizes, is a constant used in the excess risk upper bounds, where we give more weight to $\widetilde{r}_m$ for large $m$. The following lemma indicates that it is smaller than $\widetilde{r}_k$. 
\begin{lemma}[\citep{lowy2023private}]\label{lem:2}
    Under Assumption \ref{1}, we have: $\widetilde{r}^{(k)}=\widetilde{e}_1^{k} \geqslant \widetilde{e}_2^{(k)}\geqslant \widetilde{e}_4^{(k)} \geqslant \cdots \geqslant r^{(k)} $. Thus, we have $\widetilde{R}_{k,n} \leqslant \widetilde{r}_k$. 
\end{lemma}

\section{Large Lipschitz Constant Case}\label{sec:large}
\begin{algorithm}
	\renewcommand{\algorithmicrequire}{\textbf{Input:}}
	\renewcommand{\algorithmicensure}{\textbf{Return}}
	\caption{ClippedMean$(\{z_i\}_{i=1}^n,n,C)$}
	\label{alg:1}
	\begin{algorithmic}[1]
		\REQUIRE $Z=\{z_i\}_{i=1}^n,C>0$,
		% \ENSURE $U^{p}$, $V^{p}$, $b^{p}$
		\STATE Compute $\widetilde{v}:=\frac{1}{n}\sum\limits_{i=1}^{n} \prod_{C}^{}(z_i)$, where $\prod_{C}(z):= \text{argmin}_{y\in \mathbb{B}_C}\left\| y-z \right\|_2^2$ denotes the projection onto the $\ell _2$ ball $\mathbb{B}_C$. 
		\ENSURE $\widetilde{v}$
	\end{algorithmic}  
\end{algorithm}
In this section, we will focus on the population risk function satisfying $(\theta, \lambda)$-TNC, and the Lipschitz constant of the loss is extremely large (but finite). Before that, we first propose a novel localized noisy stochastic gradient method whose excess population risk is independent of the Lipschitz constant for general convex loss. See Algorithm \ref{alg:3} for details. 

In Algorithm \ref{alg:3}, we first partition the dataset into $O(\log_2 n)$ subsets where the $i$-th set has $O(2^{-i} n)$ samples. In the $i$-th iteration, we use the $i$-th set and construct an $\ell_2$-regularized empirical risk function $F_i$ with hyperparameter $\lambda_i$ in step 5. Moreover, based on the current model $w_{i-1}$, we construct the constrained set $\mathcal{W}_i$ with diameter exponential decay $D_i$. To handle a large Lipschitz constant and to solve the $\ell_2$-regularized empirical risk, we adopt a clipped gradient descent method (Algorithm \ref{alg:2}) with a clip threshold $C_i$, where we use clipped gradients (Algorithm \ref{alg:1}) to update our model instead of the original gradient. After $T_i$ iterations, we add Gaussian noise based on the stability of our clipped gradient descent to ensure $(\varepsilon, \delta)$-DP. In the following, we show Algorithm \ref{alg:3} could achieve a rate $\tilde{O}(\max\{\frac{1}{\sqrt{n}}, (\frac{d\log \frac{1}{\delta}}{\varepsilon n})^\frac{k-1}{k}\})$ with specific parameters $\lambda_i, T_i$ and $C_i$.

\begin{algorithm}
    \renewcommand{\algorithmicrequire}{\textbf{Input:}}
	\renewcommand{\algorithmicensure}{\textbf{Return}}
    \caption{Clipped Regularized Gradient Method}
	\label{alg:2}
	\begin{algorithmic}[1]
		\REQUIRE Dataset $S \in \mathcal{X}^n$, iteration number $T$, stepsize $\eta $, clipping threshold $C$, regularization $\lambda \geqslant 0$, constraint set $\mathcal{W}$
        and initialization $w_0\in \mathcal{W}$.
		% \ENSURE $U^{p}$, $V^{p}$, $b^{p}$
		\FORALL{$t \in [T-1]$} 
        \STATE $\widetilde{\nabla }F_t(w_t):=$ClippedMean$(\{\nabla f(w_t,x_i)\}_{i=1}^n;C)$ for gradients $\nabla f(w_t,x_i)$.
        \STATE $w_{t+1}=\prod_{\mathcal{W}}^{}[w_t-\eta (\widetilde{\nabla }F_t(w_t)+\lambda (w_t-w_0))]$
        \ENDFOR
		\ENSURE $w_T$
	\end{algorithmic}  
\end{algorithm}
%Here, with the help of clipping, as shown in Algorithm \ref{alg:2}, we use iterative localization (in Algorithm \ref{alg:3}) to cope with the situation with large Lipschitz constant.

\begin{algorithm}
    \renewcommand{\algorithmicrequire}{\textbf{Input:}}
	\renewcommand{\algorithmicensure}{\textbf{Output:}}
    \caption{Localized Noisy Clipped Gradient Method for DP-SCO(LNC-GM)$(w_0,\eta,n,\mathcal{W})$}
    \label{alg:3}
    \begin{algorithmic}[1]
        \REQUIRE   Dataset $S\in \mathcal{X}^n$, stepsize $\eta $, clipping threshold $\{C_i\}_{i=1}^{\log_{2} n}$, privacy parameter $\varepsilon, \delta$, hyperparameter $p$, initialization $w_0 \in \mathcal{W}$.
        \STATE   Let $l=\log_{2} n$.
        \FORALL{$i \in [l]$} 
        \STATE  Set $n_i=2^{-i}n,\eta_i=4^{-i}\eta$, $\lambda_i=\frac{1}{\eta_in_i^p}$ when $i \geqslant 2$, and $\lambda _1=\frac{1}{\eta_1n_1^{2p}}$, $T_i=\widetilde{\Theta}\left(\frac{1}{\lambda_i\eta_i}\right),\text{ and }D_i=\frac{2L_f}{\lambda_i}.$ 
        \STATE Draw a new batch $\mathcal{B}_i$ of $n_i=|\mathcal{B}_i|$ samples from $S$ without replacement.
        \STATE Denote $\widehat{F}_i(w):=\frac{1}{n_i}\sum_{j\in\mathcal{B}_i}f(w,x_j)+\frac{\lambda_i}{2}\|w-w_{i-1}\|^2.$
        \STATE Use Algorithm \ref{alg:2} with initialization $w_{i-1}$  to minimize $\widehat{F}_i$ over $\mathcal{W}_i:=\{w \in \mathcal{W}| \left\| w-w_{i-1} \right\| \leqslant D_i\}$ for $T_i$ iterations with clipping  threshold $C_i=\tilde{r}_{2k,n_i}(\frac{\varepsilon n_i}{\sqrt{d\log (1/\delta)\log(n)}})^{1/k}$ and stepsize $\eta_i$.
        Let $\hat{w}_i$ be the output of Algorithm \ref{alg:2}. 
        \STATE Set $\xi _i \sim \mathcal{N}(0,\sigma _i^2\mathbb{I}_d)$ where $\sigma _i = \frac{8C_i \sqrt{\log \frac{1}{\delta}}}{n_i\lambda_i\varepsilon }$%\sqrt[]{\log_{} (1/ \delta )}
        \STATE Set $w_i=\widehat{w}_i+\xi _i$. 
        \ENDFOR
        \STATE \textbf{Return} the final iterate $w_l$
    \end{algorithmic}
\end{algorithm}

\begin{theorem}\label{thm:1}
   Under Assumption \ref{1}, suppose that $f(\cdot ,x)$
   is $\alpha$-smooth and $L_f$-Lipschitz with $L_f < \infty$ for every $x$. Then, for any $0<\varepsilon  \leqslant \sqrt[]{\log (1/\delta)}, 0<\delta<1$ and $\eta_i \leqslant \frac{1}{\alpha}$ for all $i$,  Algorithm \ref{alg:3} is $(\varepsilon ,\delta )$-DP. Let $p\geq 1$ such that  $L_f \leqslant n^{p/2}\widetilde{R}_{2k,n}(\frac{1}{\sqrt[]{n}}+(\frac{\sqrt[]{d\log (1/\delta) \log_{} n}}{\varepsilon  n})^{\frac{k-1}{k}}) $. For any $0< \beta \leqslant \dfrac{1}{n}$, with probability at least $1-\beta $, it holds that
    \begin{equation}\nonumber
        \begin{aligned}
            F(w_l)-F(w^*)\leqslant & \tilde{O}\left(\widetilde{R}_{2k,n}D( \frac{1}{\sqrt[]{n}}+(\frac{\sqrt[]{d\log (1/\delta)}}{\varepsilon  n})^{\frac{k-1}{k}})+\frac{D\sqrt{\log (1/\beta )}}{2^{p+1}\sqrt{n}}\right), 
        %+& \frac{D\sqrt[]{\log_{} (1/\beta )}}{2^{p+1}\sqrt[]{n}}\right),
        \end{aligned}
    \end{equation}
    where the Big-$\tilde{O}$ notation omits all logarithmic terms (it is the same for other upper bounds). 
\end{theorem}
\begin{remark}
    Previous work on DP-SCO such as \citep{wang2017differentially,bassily2014private}, Lipschitz is still required for the loss function; though, it disappears in the final excess risk upper bound. And due to the property of worst-case stability and our assumption that $L_f$ can be controlled by $n^{p/2}\widetilde{R}_{2k,n}(\frac{1}{\sqrt[]{n}}+(\frac{\sqrt[]{d\log(1/\delta)log_{} n}}{\varepsilon n})^{\frac{k-1}{k}})$ for sufficiently large $p$, we reach the upper bound with high probability without $L_f$ in the final result. Compared to \citep{lowy2023private}, the main difference is that our result is in the high probability form while  \citep{lowy2023private} is only in the expectation form.
    Specifically, to achieve a high probability result, instead of adding Gaussian noise to the gradient,  we use the stability of the gradient descent. However, we cannot directly use the stability result in \citep{hardt2015train} here, which depends on the Lipschitz constant, making a large noise. We show that by using clipping, the stability now only depends on the clipping threshold. 
    % Previous work on DP-SCO, such as \citep{wang2017differentially,bassily2014private} still requires the loss function to be Lipschitz, even though the Lipschitz constant does not appear in the final excess risk upper bound. However, our proposed method (\cref{alg:3}) employs the clipped gradient so that our excess population risk is independent of the Lipschitz constant, which is beneficial for handling heavy-tailed data. Moreover, \cref{thm:1} offers the upper bound in the high probability form instead of the expectation form in \citep{lowy2023private}. To obtain a high-probability result, we consider the property of on-average stability, where we {\color{blue} how to use the property of stability} rather than directly adding Gaussian noise to the gradient. It is noted that the stability method in \citep{hardt2015train} can not be applied in our case, as its results/noise depend on the Lipschitz constant, which is quite large here and will lead to high bias. Therefore, we employ a clipping method so that the noise will only depend on the clipping threshold.
    
     %What we also need to note is that our requirement of the Lipschitz constant is weak, namely, $L_f$ can be very large but under control of some $n^{p/2}\widetilde{R}_{2k,n}(\frac{1}{\sqrt[]{n}}+(\frac{\sqrt[]{d \log_{} n}}{\varepsilon  n})^{\frac{k-1}{k}})$, which can be adapted to the data with outliers.
\end{remark}

Based on our novel locality algorithm, we then apply it to TNC functions. See Algorithm \ref{alg:4} for details. Specifically, we partition the dataset into several subsets of equal size. As the iteration number increases, we consider a constrained set centered at the current parameter with a smaller diameter and learning rate in Algorithm \ref{alg:3}. 

%Algorithm 4 helps us divide the diameter more adaptive to each stage. What we need to consider is how we project onto the ball given in step 4.  \di{add some introductions or details  of Algorithm 4} 
%A general requirement means low accuracy. Therefore, adding some restrictions on $f$ is appropriate in many situations. With the help of TNC, we can reach better bounds. Actually, strongly convex (see Definition\ref{def:6}) is a special case of TNC. So, bearing this in mind and combining the above Localized Noisy Clipped method with the Growth condition, we have the following algorithm.
\begin{algorithm}
\renewcommand{\algorithmicrequire}{\textbf{Input:}}
\renewcommand{\algorithmicensure}{\textbf{Return}}
\caption{Private Stochastic Approximation$(w_1,n,R_0)$}
\label{alg:4}
\begin{algorithmic}[1]
\REQUIRE	Dataset $S\in \mathcal{X}$,  initial point $w_1 \in \mathcal{W}$, privacy parameter $\varepsilon $ and $\delta $, hyperparameter $p$, initial diameter $R_0$.
\STATE Set $\hat{w}_0=w_1$, $m=\lfloor \frac{1}{2}\log_{2} \frac{2n}{\log_{2} n} \rfloor-1 $, $n_0= \lfloor \frac{n}{m}\rfloor $. Then partition the dateset $S$ into $m$ disjoint subsets, namely, $\{S_1,\cdots,S_m\}$ with each $|S_i|=n_0$.
\FORALL{$l\in [m]$}
\STATE Set $\gamma _l=\frac{R_{l-1}}{n_0^{\frac{p}{2}}}\min \{\frac{1}{L_f},\frac{1}{\widetilde{R}_{2k,n}n_0^{\frac{p+1}{2}}}(\frac{\varepsilon n_0}{\sqrt[]{d \log_{} n}})^{\frac{k-1}{k}}$, $ \frac{1}{n_0^{\frac{p-1}{2}} L_f^2 \sqrt[]{\log_{} n_0 \log_{} (1/\beta )}}\} $ and $R_l=\frac{R_{l-1}}{2}$. 
\STATE Denote $\hat{w}_l =$ LNC-GM$(\hat{w}_{l-1},\gamma_l,n_l,\mathcal{W})$,  and constrained set $ \mathcal{W}\cap \mathbb{B}(\hat{w}_{l-1},R_{l-1})$.
\ENDFOR
\ENSURE $\hat{w}_m$ 
\end{algorithmic}
\end{algorithm}
\begin{theorem}\label{thm:2}
   Under Assumption \ref{1} and suppose that the population risk function $F(\cdot )$ is $(\theta ,\lambda )$-TNC with $\theta\geq 2$, and $f(\cdot ,x )$ is  $\alpha  $-smooth and $L_f$-Lipschitz for each $x$. Additionally, take  $p\geq 1$ such that  $L_f \leqslant n^{p/2}\widetilde{R}_{2k,n}(\frac{1}{\sqrt[]{n}}+(\frac{\sqrt[]{d \log (1/\delta)\log_{} n}}{\varepsilon  n})^{\frac{k-1}{k}}) $, then algorithm \ref{alg:4} is $(\varepsilon ,\delta )$-DP. %Morever based on different stepsizes $\{\gamma_k \}_{k=1}^m$ and noises if $\gamma _k \leqslant \frac{1}{\alpha }$. 
    Moreover, for sufficiently large $n$ such that $\gamma_l \leqslant \frac{1}{\alpha}$, with probability at least $1-\beta $, we have
    % \begin{equation}\nonumber
    %    \begin{aligned}
    %     &F\left(\hat{w}_m\right)-F\left(w^*\right)\\
    %     \leqslant &\tilde{O}\left(\frac{1}{\lambda^{\frac{1}{\theta-1}}} (\widetilde{R}_{2k,n}( \frac{1}{\sqrt[]{n}}+(\frac{\sqrt[]{d}}{\varepsilon  n})^{\frac{k-1}{k}}))^{\frac{\theta}{\theta-1}}\right).
    %    \end{aligned}
    % \end{equation}
    \begin{equation}\nonumber
       % \begin{aligned}
        F\left(\hat{w}_m\right)-F\left(w^*\right)
        \leqslant \tilde{O}\left(\frac{1}{\lambda^{\frac{1}{\theta-1}}} (\widetilde{R}_{2k,n}( \frac{1}{\sqrt[]{n}}+(\frac{\sqrt[]{d \log (1/\delta)}}{\varepsilon  n})^{\frac{k-1}{k}})++\frac{\sqrt[]{\log_{} n \log_{} (1/\beta )}}{2^{p+1}\sqrt[]{n}})^{\frac{\theta}{\theta-1}}\right).
       % \end{aligned}
    \end{equation}
\end{theorem}
We note that there is no dependence on p in the final bound in Theorem \ref{thm:1} and \ref{thm:2}. $p$ is used to control the Lipschitz constant thus we can remove the Lipschitz constant from the final bound. We can see that in the proof of Theorem \ref{thm:1}, there exists a term with $n^p$ both in the numerators and denominators. By assuming that $L_f$ is controlled by the $O(n^p/2)$ and choosing specific $\eta$, we can eliminate the $p$ in the final bound. A similar result holds for Theorem $2$.
\begin{remark}
  In the case of $O(1)$-Lipschitz loss under TNC, compared with the optimal rate $\Theta((( \frac{1}{\sqrt[]{n}}+(\frac{\sqrt[]{d}}{\varepsilon n})^{\frac{k-1}{k}}))^{\frac{\theta}{\theta-1}})$ in \citep{asi2021adapting}, our improved result gets rid of the dependence of Lipschitz constant, which could be extremely large. Moreover, when $\theta=2$, i.e., the population risk is strongly convex, our result covers the result in \citep{lowy2023private}. Thus, our result is a generalized upper bound. It is also notable that our upper bound is independent of the diameter of the constrained set and the Lipschitz-smoothness parameter. In Algorithm \ref{alg:4}, one need the projection onto the ball $ \mathcal{W}\cap \mathbb{B}(\hat{w}_{l-1},R_{l-1})$ in each iteration of the Phased-SGD in each phase. This could be solved using Dykstra's algorithm \citep{dykstra1983algorithm,boyle1986method}. 
\end{remark}
\noindent {\bf Example.}  We consider the $\ell_1$ constrained $\ell_4$-norm linear regression, which has been studied in \citep{xu2017stochastic} and satisfies TNC with $\theta=4$ \citep{liu2018fast}. Specifically, it can be written as the following.  
 \begin{equation}\label{eq25}
     \min \limits_{\|w\|_1\leqslant 1}F(w)\overset{\Delta}{=} \mathbb{E}[(\langle w, x\rangle-y)^4 ].
 \end{equation}
 When $y$ is bounded by $O(1)$ and $x$ follows a truncated normal Gaussian distribution at $[-n, n]^d$. Then we can see that the loss function is $\text{Poly}(n)$-Lipschitz, but its $2k$-th moment is $O(1)$. In this case, our bound in  \eqref{thm:2} is much smaller than the previous results in \citep{asi2021adapting,su2024faster}. 

\noindent \textbf{Example}. We also investigate the $\ell_2$-norm regularized logistic regression problem with regularization parameter $\lambda$ subject to the unit $\ell_2$-norm ball constraint. This formulation exhibits $\lambda$-strong convexity and consequently satisfies the TNC with $\theta=2$. Specifically, let $h_w(x) = \frac{1}{1+e^{-\langle x,w \rangle}}$ denote the logistic function and $y \in \{0, 1\}$ represent the binary response variable. The optimization problem can be formulated as
    \begin{equation}\label{eg2}
\min_{\|w\|_2 \leq 1} F(w) \triangleq \mathbb{E}[-y \log h_w(x) - (1-y) \log(1 - h_w(x))] + \frac{\lambda}{2}\|w\|_2^2.
\end{equation}

Suppose that $\|x\|_2 \leq R$ or x is sub-Gaussian and $y$ is bounded by some constant. Under these conditions, we can derive the $2k$-th moment for \eqref{eg2}, which is $O(1)$ for fixed $k$ and $R$. Under these circumstances, our approach yields improvements over the previous results established in \citep{liu2018fast}.

So far, we have proposed an algorithm for TNC. Nevertheless, we also find that Theorem \ref{thm:2} needs a strong assumption on $\theta $, i.e. $\theta \geqslant 2$. Thus, a direct question that occurs to us is whether we can further improve the upper bound. To conquer the disadvantage of the above algorithm, we propose the following. We assume $\theta $ is unknown but bigger than some definite $\bar{\theta }>1$. Then we divide the whole dataset into subsets with distinct elements, detailly $l=\left\lfloor\left(\log _{\bar{\theta }} 2\right) \cdot \log \log n\right\rfloor$ with $n_i=\left\lfloor 2^{i-1} n /(\log n)^{\log _{\bar{\theta }}^2 2}\right\rfloor$ samples for each subset. Then we run the Algorithm \ref{alg:1} for $l$ times while each phase implements on the $i$-th subset and is initialized at the output of the previous one.
\begin{algorithm}
    \renewcommand{\algorithmicrequire}{\textbf{Input:}}
    \renewcommand{\algorithmicensure}{\textbf{Return}}
    \caption{Iterated Localized Noisy Clipped Gradient Method}
    \label{alg:5}
    \begin{algorithmic}[1]
        \REQUIRE Dataset $S\in \mathcal {X}^n$, initial point $w_0 \in \mathcal{W}$,
        privacy parameter $\varepsilon$ and  $\delta$, parameter $p$, initial diameter $R_0$. 
        \STATE Partite the data $S$ into $l$ disjoint subsets $\left\{S_1, \cdots, S_l\right\}$, where $l=\left\lfloor\left(\log _{\bar{\theta }} 2\right) \cdot \log \log n\right\rfloor$
        and for each $i \in[l],\left|S_i\right|=n_i=\left\lfloor 2^{i-1} n /(\log n)^{\log _{\bar{\theta }}^2 2}\right\rfloor$.
        \FORALL{$t=1, \cdots, l$}
        \STATE	Let $w_t=$ LNC-GM$\left(S_i, w_{t-1}, \eta_t,  \mathcal{W}\right)$, where $  \eta_t=\frac{R_{t-1}}{n_0^{\frac{p}{2}}}\min \{\frac{1}{L_f},\frac{1}{\widetilde{R}_{2k,n}n_i^{\frac{p+1}{2}}}(\frac{\varepsilon n_i}{\sqrt[]{d \log_{} n}})^{\frac{k-1}{k}}$, $ \frac{1}{n_i^{\frac{p-1}{2}} L_f^2 \sqrt[]{\log_{} n_i \log_{} (1/\beta )}}\} $  and $R_l=\frac{R_{l-1}}{2}$. 
        \ENDFOR
        \ENSURE	$w_l$ 
    \end{algorithmic}
\end{algorithm}
\begin{theorem}\label{thm:3}
   Under Assumption \ref{1} and   assume that the loss function $F(\cdot )$ satisfies  $(\theta ,\lambda )$-TNC with parameter $\theta  \geqslant  \bar{\theta }>1$ for some definite constant $\bar{\theta}$, and $f(\cdot ,x)$ is convex, $\alpha $ smooth and $L_f$-Lipschitz for each $x$. Algorithm \ref{alg:5} is $(\varepsilon ,\delta )$-DP for any $\varepsilon \leqslant 2 \log_{} (1/\delta )$, and take  $p\geq 1$ such that  $L_f \leqslant n^{p/2}\widetilde{R}_{2k,n}(\frac{1}{\sqrt[]{n}}+(\frac{\sqrt[]{d \log (1/\delta)  \log_{} n}}{\varepsilon  n})^{\frac{k-1}{k}}) $. Moreover, if the sample size $n$ is sufficiently large such that $\bar{\theta }\geqslant 2^{\frac{\log_{} \log_{} n}{\log_{} n-1}}$  and $\eta_t  \leqslant \frac{1}{\alpha}$, we  have with probability at least $1-\beta$ 

% \begin{equation}\nonumber
%     \begin{aligned}
%         &F\left(w_l \right)-F\left(w^*\right)\\
%     \leqslant 
%      &\tilde{O}\left((\frac{1}{\lambda})^{\frac{1}{\theta-1}}(\widetilde{R}_{2k,n}( \frac{1}{\sqrt[]{n}}+(\frac{\sqrt[]{d }}{\varepsilon  n})^{\frac{k-1}{k}}))^{\frac{\theta}{\theta-1}}\right).
%         %& \leqslant 2^{\frac{3 \theta}{2(\theta-1)}} \cdot C\left(\frac{c^\theta }{\lambda}\right)^{\frac{1}{\theta-1}}\left(\frac{1}{n}+\frac{d \log (1 / \delta)}{\varepsilon^2 n^2}\right)^{\frac{\theta}{2(\theta-1)}}.
%     \end{aligned}
% \end{equation}
\begin{equation}\nonumber
    % \begin{aligned}
        F\left(w_l \right)-F\left(w^*\right)
    \leqslant 
     \tilde{O}\left((\frac{1}{\lambda})^{\frac{1}{\theta-1}}(\widetilde{R}_{2k,n}( \frac{1}{\sqrt[]{n}}+(\frac{\sqrt[]{d \log (1/\delta) }}{\varepsilon  n})^{\frac{k-1}{k}})+\frac{\sqrt[]{\log_{} (1/\beta )}}{2^{p+1}\sqrt[]{n}})^{\frac{\theta}{\theta-1}}\right).
        %& \leqslant 2^{\frac{3 \theta}{2(\theta-1)}} \cdot C\left(\frac{c^\theta }{\lambda}\right)^{\frac{1}{\theta-1}}\left(\frac{1}{n}+\frac{d \log (1 / \delta)}{\varepsilon^2 n^2}\right)^{\frac{\theta}{2(\theta-1)}}.
    % \end{aligned}
\end{equation}
%where we use $\tilde{O}$ to denote the result with $\log n$ and omit $\log n$ for short.
%\di{change the results related to $n$, rather than $n_l$} 
\end{theorem}
\begin{remark}
    We pause to have another glimpse of Algorithm \ref{alg:4} and Algorithm \ref{alg:5}. Note that they have a similar procedure to take the dataset apart, while the number of each subset is the same in Algorithm \ref{alg:5} and increases in Algorithm \ref{alg:5} as the iteration grows. And the set we project on also varies between Algorithm \ref{alg:4} and \ref{alg:5}.
\end{remark}
\section{Lower Bounds}
In this section, we will show that the above upper bound 
is nearly optimal (if $\tilde{r}_{2k}$ and $\tilde{r}_{k}$ are asymptotically the same) by providing lower bounds of the private minimax rate for $\rho$-zCDP. Specifically, for a sample space $\mathcal{X}\subseteq \mathbb{R}^d$ and collection of distributions $\mathcal{P}$ over $\mathcal{X}$, we define the function
class $\mathcal{F}_k^\theta( \mathcal{P}, \tilde{r}^{(k)})$ as the set of population risk functions from $\mathbb{R}^d\mapsto \mathbb{R}$ that satisfy $(\theta, 1)$-TNC and their loss satisfies Assumption \ref{1}. We define the constrained minimax risk
\begin{align*}
    \mathcal{M}(\mathcal{W}, \mathcal{P}, \mathcal{F}_k^\theta( \mathcal{P}, \tilde{r}^{(k)}), \rho)=  
    \inf_{\mathcal{A}\in \mathcal{Q}(\rho) }\max_{F\times P \in  \mathcal{F}_k^\theta( \mathcal{P}, \tilde{r}_k) \times \mathcal{P}}\mathbb{E}_{\mathcal{A}, D\in P^n} [F(\mathcal{A}(D))-\min_{w\in \mathcal{W}} F(w)],
\end{align*}
where $\mathcal{Q}(\rho)$ is the set of all $\rho$-zCDP algorithms. We will show the following two results for different sample spaces and constraint sets. 

\begin{theorem}\label{thm:low1}
    For any $\theta, k\geq 2, \tilde{r}^{(k)}>0$, denote $\mathcal{X}=\{\pm p^{-\frac{1}{k}}\frac{\tilde{r}_k}{2\sqrt{d}}\}^d \cup \{0\}$ with $\tilde{r}_k=(\tilde{r}^{(k)})^\frac{1}{k}$, and $\mathcal{W}=\mathbb{B}_r$ with $r= (\frac{p^{-\frac{1}{k}}\tilde{r}_k}{2})^\frac{1}{\theta-1}$ and $p=\frac{d}{n\sqrt{\rho}}$, then, if $n$ is large enough such that $n\geq \Omega(\frac{\sqrt{d}}{\sqrt{\rho}})$, we have the following lower bound
    \begin{equation*}
         \mathcal{M}(\mathcal{W}, \mathcal{P}, \mathcal{F}_k^\theta( \mathcal{P}, \tilde{r}^{(k)}), \rho)\geq \Omega \left( (\tilde{r}_k((\frac{\sqrt{d}}{\sqrt{\rho}n})^\frac{k-1}{k}))^\frac{\theta}{\theta-1}\right). 
    \end{equation*}
\end{theorem}

\begin{theorem}\label{thm:low2}
    For any $\theta, k\geq 2, \tilde{r}_k>0$, denote $\mathcal{X}=\{\pm \frac{\tilde{r}_k}{2\sqrt{d}}\}^d $, and $\mathcal{W}=\mathbb{B}_r$ with $r= (\frac{\tilde{r}_k}{2})^\frac{1}{\theta-1}$, then, if $n\geq \Omega({\sqrt{d}})$, we have the following lower bound 
    \begin{equation*}
         \mathcal{M}(\mathcal{W}, \mathcal{P}, \mathcal{F}_k^\theta( \mathcal{P}, \tilde{r}^{(k)}), \rho)\geq \Omega \left( (\frac{\tilde{r}_k }{\sqrt{n}})^\frac{\theta}{\theta-1}\right). 
    \end{equation*}
\end{theorem}
\begin{remark}
    First, it is notable that although the upper bounds in Section \ref{sec:large} are for $(\varepsilon, \delta)$-DP, we can easily extend to the $\rho$-zCDP case as we used the Gaussian mechanism and parallel theorem to guarantee DP, which also holds for zCDP \citep{bun2016concentrated}. The only difference is replacing the term $O(\frac{\sqrt{\log \frac{1}{\delta}}}{\varepsilon})$ by $O(\frac{1}{\sqrt{\rho}})$. Thus, from this side, combining with  Theorem \ref{thm:low1} and \ref{thm:low2}, we can see the upper bound is nearly optimal for $\rho$-zCDP in the general case if $\tilde{r}_{2k}$ (since $\tilde{R}_{2k,n}\leqslant \tilde{r}_{2k}$) and $\tilde{r}_{k}$ are asymptotically the same. Secondly, in the Lipschitz case for $(\varepsilon,\delta)$-DP, \citep{asi2021private} proved the lower bound result via
a reduction to the ERM problem for general convex loss. However, their reduction cannot be applied to our problem as their proof heavily relies on the $O(1)$-Lipschitz condition, which is not satisfied for our loss. For $\varepsilon$-DP, \citep{asi2021private} considered the empirical risk and used the packing argument for the lower bound, which cannot be applied to our problem as our loss is not constant Lipschitz.   In our proof, we directly considered the population risk $F_P(w)=-\langle w, \mathbb{E}_P[x] \rangle+\frac{1}{\theta}\|w\|_2^\theta$  for some data distribution $P$ and used private Fano's lemma to prove the lower bound.   
\end{remark}
\begin{algorithm}
    \renewcommand{\algorithmicrequire}{\textbf{Input:}}
    \renewcommand{\algorithmicensure}{\textbf{Return}}
    \caption{Permuted Noisy Clipped Accelerated SGD for Heavy-Tailed DP SCO (PNCA-SGD) }
    \label{alg:6}
    \begin{algorithmic}[1]
        \REQUIRE	Data $S \in \mathcal{X}^n$, iteration number $T$, stepsize parameters $\left\{\eta_t\right\}_{t \in[T]}$, $ \left\{\alpha_t\right\}_{t \in[T]}$ with $\alpha_1=1$, private paratemter $\varepsilon, \delta$, initialization $w_0$.
        \STATE Randomly permute the data and denote the permuted data as $\{x_1, \cdots  , x_n\}$.
        \STATE Initialize $w_0^{a g}=w_0$.
        \FORALL{$t \in[T]$}
        \STATE	$w_t^{m d}:=\left(1-\alpha_t\right) w_{t-1}^{a g}+\alpha_t w_{t-1}$.
        \STATE Draw new batch $\mathcal{B}_t$ (without replacement) of $n / T$ samples from $S$.
        \STATE $\widetilde{\nabla} F_t\left(w_t^{m d}\right):=$ ClippedMean $\left(\left\{\nabla f\left(w_t^{m d}, x\right)\right\}_{x \in \mathcal{B}_t} ; \frac{n}{T} ; C\right)+\zeta_i$, where $\zeta_i\sim \mathcal{N}(0, \sigma^2\mathbb{I}_d)$, $\sigma^2=O(\frac{C^2 T\log \frac{1}{\delta} }{n^2\varepsilon^2})$ and $C=\widetilde{r}_k\left(\frac{\varepsilon n}{\sqrt{d \log_{} (1/\delta )}}\right)^{1 / k}$. 
        \STATE $w_t:=\underset{w \in \mathcal{W}}{\arg}\left\{\alpha_t\left\langle\tilde{\nabla} F_t\left(w_t^{m d}\right), w\right\rangle+\frac{\eta_t}{2}\left\|w_{t-1}-w\right\|^2\right\}$.
        \STATE $w_t^{a g}:=\alpha_t w_t+\left(1-\alpha_t\right) w_{t-1}^{a g}$.
        \ENDFOR 
        \ENSURE	$w_T^{a g}$
    \end{algorithmic}
\end{algorithm}
\section{Relax the Lipschitz Assumption}

In the previous section, we have considered the Lipschitz case and show that under the TNC, compared to that for the general convex loss, it is possible to get improved excess population risk that is independent of the Lipschitz constant. There are still two questions left: (1) Compared to the previous studies on DP-SCO with heavy-tailed gradient such as \citep{wang2020differentially,kamath2021improved}, our above upper bounds still need the finite Lipschitz condition; (2) We can see our upper bounds depend on $\tilde{R}_{2k, n}\leqslant \tilde{r}_{2k}$ while the lower bounds only depend on $\tilde{r}_k$. Thus, there is a gap for the moment term. In this section, we aim to address these two issues. Specifically, we will show that even if the loss function is not Lipschitz, it is still possible to get the same upper bound as in the above section when $\varepsilon$ is small enough. Moreover, we can improve the dependency from $\tilde{R}_{2k, n}$ to $\tilde{r}_k$. 

Specifically, our main method, Algorithm \ref{alg:7}, shares a similar idea as in Algorithm \ref{alg:5} with different parameters and base algorithm. Specifically, rather than using Algorithm \ref{alg:3}, here we propose Algorithm \ref{alg:6} as our base algorithm, which is a shuffled, clipped, and private version of the accelerated SGD. Specifically, in step 1 we randomly shuffle the data for privacy amplification \citep{feldman2022hiding}. Then, in each iteration, we clip the gradients and add Gaussian noise to ensure DP. We can show that with some parameters, the output could achieve an upper bound similar to Theorem \ref{thm:1} even if the loss is not Lipschitz. 

\begin{algorithm}
    \renewcommand{\algorithmicrequire}{\textbf{Input:}}
    \renewcommand{\algorithmicensure}{\textbf{Return}}
    \caption{Iterated PNCA-SGD $\left(w_0, n, \mathcal{W}, \bar{\theta}\right)$ }
    \label{alg:7}
    \begin{algorithmic}[1]
        \REQUIRE Dataset $S\in \mathcal{X}^n$, initial point $w_0 \in \mathcal{W}$, privacy parameter $\varepsilon$ and $\delta$. 
        \STATE Partite the data $S$ into $k$ disjoint subsets $\left\{S_1, \cdots, S_k\right\}$, where $k=\left\lfloor\left(\log _{\bar{\theta }} 2\right) \cdot \log \log n\right\rfloor$,
        and for each $i \in[k],\left|S_i\right|=n_i=\left\lfloor 2^{i-1} n /(\log n)^{\log _{\bar{\theta }}^2 2}\right\rfloor$.
        \FORALL{$t=1, \cdots, k$}
        \STATE	Let $w_t=$ PNCA-SGD $\left(w_{t-1}, \eta_t, n_t, \mathcal{W}\right)$, where the AC-SA runs on the $t$-th subset $S_i$.For $(\varepsilon, \delta)$-DP, $  \eta_t=\frac{4\eta}{t(t+1)} $, $\alpha_t= \frac{2}{t+2}$  and $R_l=\frac{R_{l-1}}{2}$. 
        \ENDFOR
        \ENSURE	$w_k$ 
    \end{algorithmic}
\end{algorithm}  
%Based on AC-SA algorithm and the restrict on $\varepsilon$, i.e. $\varepsilon<\frac{1}{\sqrt{n}}$, we can improve the result by randomly permuting the batches. Furthermore, given the strongly convex assumption, we can handle the case when the Lipschitz constant is infinity.

\begin{theorem}\label{thm:SC1}
For any $\varepsilon=O(\sqrt{\frac{{\log n/\delta}}{n}})$, and $0<\delta<1$, Algorithm \ref{alg:6} is $(\varepsilon, \delta)$-DP. Moreover, under 
Assumption \ref{1} and assuming function $F$ is $\beta $-smooth with the diameter $D$ over $w\in \mathcal{W}$, then the output of Algorithm \ref{alg:6}, by selecting the following $T$,
\begin{equation}\nonumber
        T = \min \{\sqrt[]{\frac{\beta D}{\widetilde{r}_k}}\cdot (\frac{\varepsilon n}{\sqrt[]{d \log_{} (1/\delta )}})^{\frac{k-1}{2k}}\!,\ \sqrt[]{\frac{\beta D}{\widetilde{r}_k}}\cdot n^{1/4} \},
\end{equation}
we have
    \begin{equation}\nonumber
        \mathbb{E} F\left(w_T^{a g}\right)-F^*  \leqslant O\left(\widetilde{r}_k D(\frac{1}{\sqrt{n}}+(\frac{\sqrt{d \log_{} (1/\delta )}}{\varepsilon n})^{\frac{k-1}{k}})\right). 
    \end{equation}
    % \di{what is $C$? what is $T$?}
\end{theorem}
Note that \citep{lowy2023private} also proposes a private accelerated SGD. However, their bound is sub-optimal (see the second row in Table \ref{tab:1}). Here, we leverage privacy amplification via shuffling to reduce the noise added to each iteration. Thus, we can get the optimal rate here. We note that this is also the first result that can achieve the optimal rate for the general convex function without the Lipschitz assumption. Based on this result, we have the following theorem for Algorithm \ref{alg:7}.
%We can see that in \citep{lowy2023private}, accelerated algorithm has been proposed to improve the result of convergence. Based on Algorithm \ref{alg:6}, we extend the Iterated AC-SA algorithm to Strongly Convex case, which is described in Theorem \ref{thm:SC1}.

\begin{theorem}\label{thm:SC2}
   For any $\varepsilon=O(\sqrt{\frac{{\log n/\delta}}{n}})$, and $0<\delta<1$, Algorithm \ref{alg:7} is $(\varepsilon, \delta)$-DP. Moreover, under 
Assumption \ref{1} and assuming function $F$ is $\beta $-smooth, then we have 
\begin{equation}
    \mathbb{E} F(\hat{w}_m)-F(w^*) \leqslant 
         \tilde{O}\left(\frac{1}{\lambda^{\frac{1}{\theta-1}}} (\widetilde{r}_k(\frac{1}{\sqrt[]{n}}+(\frac{\sqrt{d \log (1/\delta) }}{\varepsilon  n})^{\frac{k-1}{k}}))^{\frac{\theta}{\theta-1}}\right).
\end{equation}

\end{theorem}
%Compared with the results in the above section,  we can see the dependency on the moment is $\widetilde{r}_k$ while it is $\widetilde{R}_{2k, n}$ in the above results, which are incomparable in general. In the following, we show that the dependency of $\widetilde{r}_k$ is optimal. However, 
Compared with the results in the above section, we can see the result in Theorem \ref{thm:SC2} is in the expectation form, which is due to the noisy clipped gradient in Algorithm \ref{alg:6}. Moreover, the constraint of $\varepsilon=O(\sqrt{\frac{{\log n/\delta}}{n}})$ comes from the results of privacy amplification via shuffling \citep{feldman2022hiding}. We leave these two assumptions to be relaxed for future research. Moreover, the improvement from $\widetilde{R}_{2k,n}$ to $\widetilde{r}_k$ is due to the different results between Theorem \ref{thm:SC1} and \ref{thm:1}. 

\section{Numerical Experiments}
In this section, we conduct a series of  numerical experiments with three different datasets to show the performance of our algorithms.

\subsection{Experimental Settings}

For instances satisfying TNC, we investigate two representative examples that have been thoroughly analyzed in existing literature. We first consider the $\ell_4$-norm  linear regression. This setting satisfies TNC with parameter $\theta=4$. More precisely, we have
\begin{equation}\nonumber
    \min_{w \in \mathcal{D}} F(w) \triangleq \mathbb{E}[(\langle w, x \rangle - y)^2].
\end{equation}

We also investigate the $\ell_2$-norm regularized logistic regression problem with regularization parameter $\lambda$. This formulation exhibits $\lambda$-strong convexity and consequently satisfies the $(2, \lambda)$-TNC condition. Specifically, let $h_w(x) = \frac{1}{1+e^{-\langle x,w \rangle}}$ denote the logistic function and $y \in \{0, 1\}$ represent the binary response variable. The optimization problem can be formulated as follows:
\begin{equation}\nonumber
\min_{w \in \mathcal{D}} F(w) \triangleq \mathbb{E}\left[-y \log h_w(x) - (1-y) \log(1 - h_w(x))\right] + \frac{\lambda}{2}\|w\|_2^2. 
\end{equation}
\paragraph{Baselines}
Although our investigation covers both $(\varepsilon, \delta)$-DP and $\varepsilon$-DP, in practice, we prefer $(\varepsilon, \delta)$-DP. Therefore, this section is dedicated to the performance of $(\varepsilon, \delta)$-DP. For the problem mentioned before, we adopt DP-SGD as a baseline.

\begin{itemize}
\item \textbf{DP-SGD} \citep{abadi2016deep}. The foundational DP-SGD algorithm was originally proposed by  \citep{bassily2014private}. However, its practical efficacy in the initial formulation proved inadequate, as evidenced by \citep{wang2017differentially}. To overcome the deficiencies, we adopt the batched and clipped variant as proposed by \citep{abadi2016deep}, which demonstrates substantial improvement.  Although the algorithm of \citep{abadi2016deep} with arbitrary clipping thresholds does not provide theoretical convergence guarantees for excess population risk, it achieves superior empirical performance.  Our experimental framework incorporates systematic hyperparameter tuning to optimize results, with findings reported using some selected parameters.

\item \textbf{LNC-GM} (Algorithm 3). LNC-GM could be considered as the state-of-the-art method for DP-SCO problem with smooth convex loss functions, especilally with large Lipschitz constant. 

\item \textbf{PNCA-SGD} (Algorithm 6). PNCA-SGD could be regarded as the state-of-the-art algorithm for DP-SCO problem withour requirement of Lipschitzness for the smooth loss functions.
\end{itemize}

\paragraph{Dataset and Parameter Settings}
We will implement the LNC-GM algorithm on three real-world datasets from the libsvm website, namely a8a ($n = 22,696$, $d = 123$ for training, and $n = 9,865$ for testing), a9a ($n = 32,561$, $d = 123$ for training, and $n = 16,281$ for testing),and w7a ($n = 24,692$, $d = 300$ for training, and $n = 25,057$ for testing). 

We study the above-mentioned TNC problem and their corresponding testing errors with various sample sizes and privacy budgets $\varepsilon$. When performing the results for different sample sizes, we will fix $\varepsilon=8$ and consider different sample sizes n that are at most $3.5 \times 10^4$. When performing the results for different privacy budgets $\varepsilon$, we will use $n=10^4$ samples and choose $\varepsilon=\{0.5, 1.0, 1.5, 2.0, 3.0, 4.0, 5.0\}$ respectively. We will fix $\delta=\frac{1}{n^{1.1}}$ for all experiments.

\subsection{Experiment Results} 

Figures are attached in the Appendix. Figures \ref{fig:a8a}, \ref{fig:a9a}, and \ref{fig:w7a} present the results of $\ell_4$-norm linear regression for our proposed methods (LNC-GM and PNCA-SGD) compared to the baseline DP-SGD algorithm. Our base algorithm demonstrates performance comparable to or superior to DP-SGD in most cases, with closely matched results on the w7a dataset. Without dataset normalization, the uniform Lipschitz assumption no longer holds. Our LNC-GM algorithm can accommodate large Lipschitz constants, which explains why its performance is not inferior to DP-SGD and even outperforms it on the a8a and a9a datasets. The relaxation of the Lipschitz requirement represents a key characteristic of PNCA-SGD, albeit requiring a small privacy budget $\varepsilon$. As illustrated in Figures \ref{fig:a8aa} and \ref{fig:a9aa}, a small value of $\varepsilon$ ensures that PNCA-SGD achieves superior performance compared to DP-SGD.

\begin{figure}[htbp]
    \centering
    \begin{subfigure}{0.45\textwidth}
        \centering
        \includegraphics[width=\textwidth]{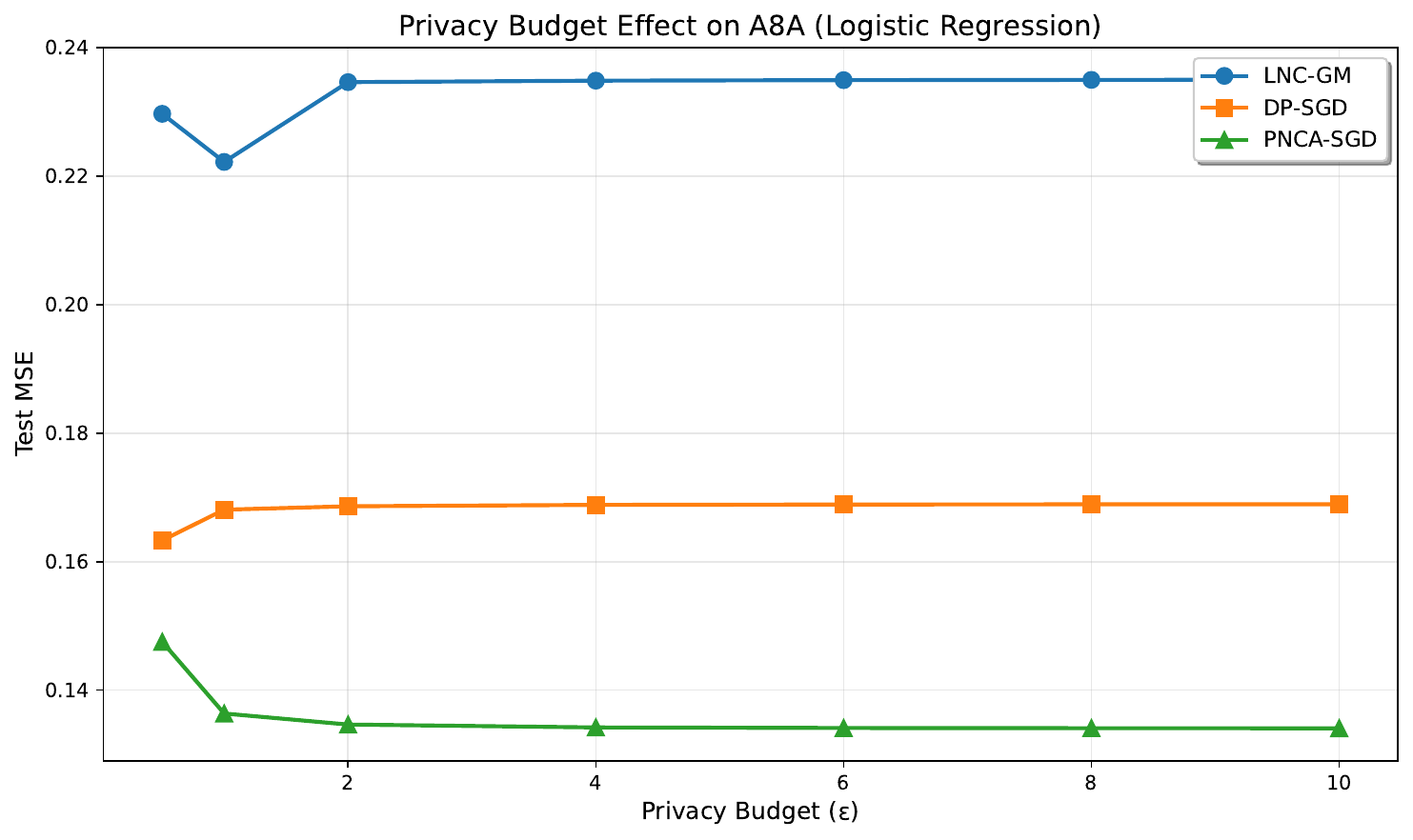}
        \caption{Results of logistic regression with different privacy budget $\varepsilon$ on  a8a}
        \label{fig:loga8aa}
    \end{subfigure}
    \hfill
    \begin{subfigure}{0.45\textwidth}
        \centering
        \includegraphics[width=\textwidth]{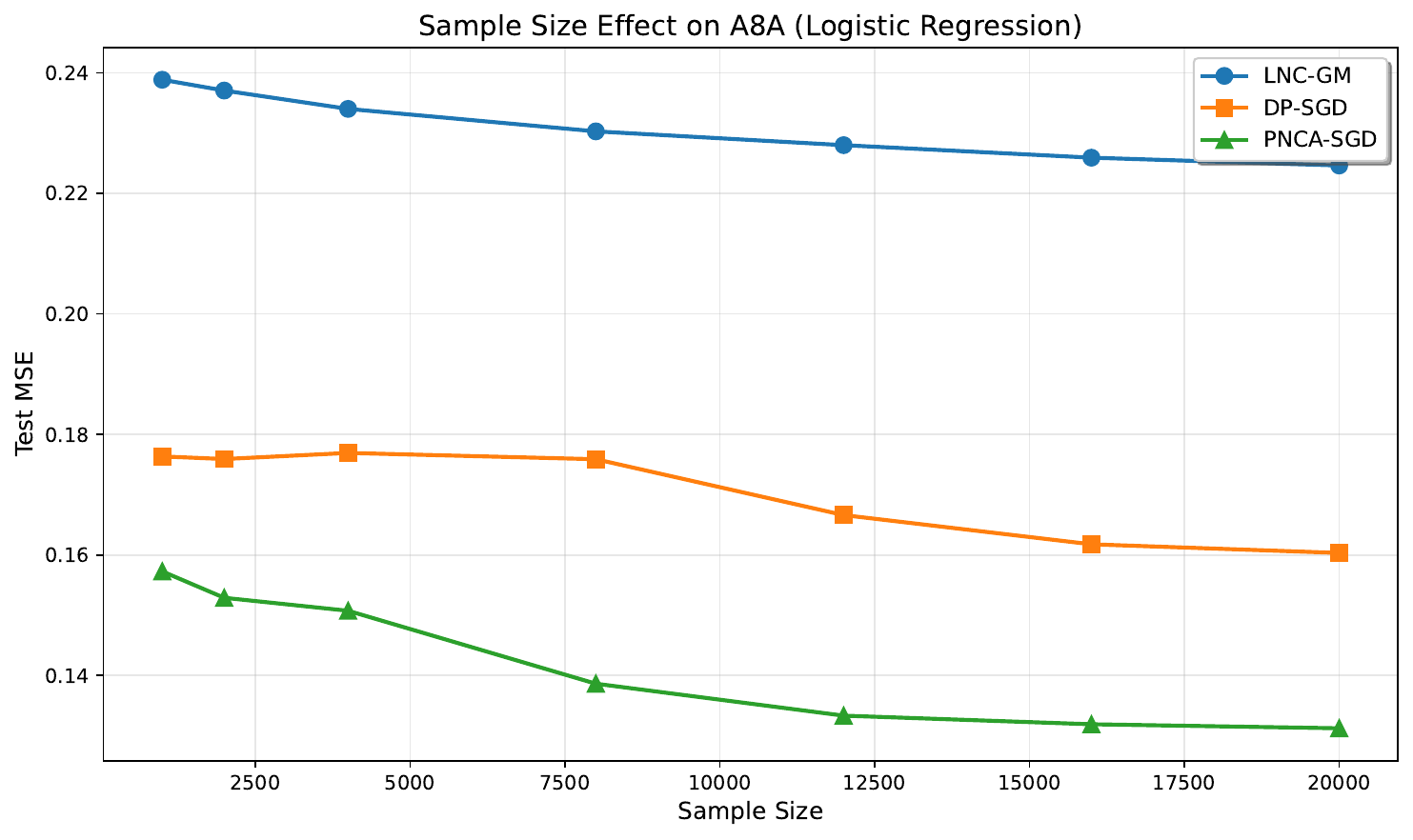}
        \caption{Results of logistic regression with different training sample size on  a8a}
        \label{fig:loga8ab}
    \end{subfigure}
    \caption{Two experiments on logistic regression with a8a }
    \label{fig:loga8a}
\end{figure}

\begin{figure}[htbp]
    \centering
    \begin{subfigure}{0.45\textwidth}
        \centering
        \includegraphics[width=\textwidth]{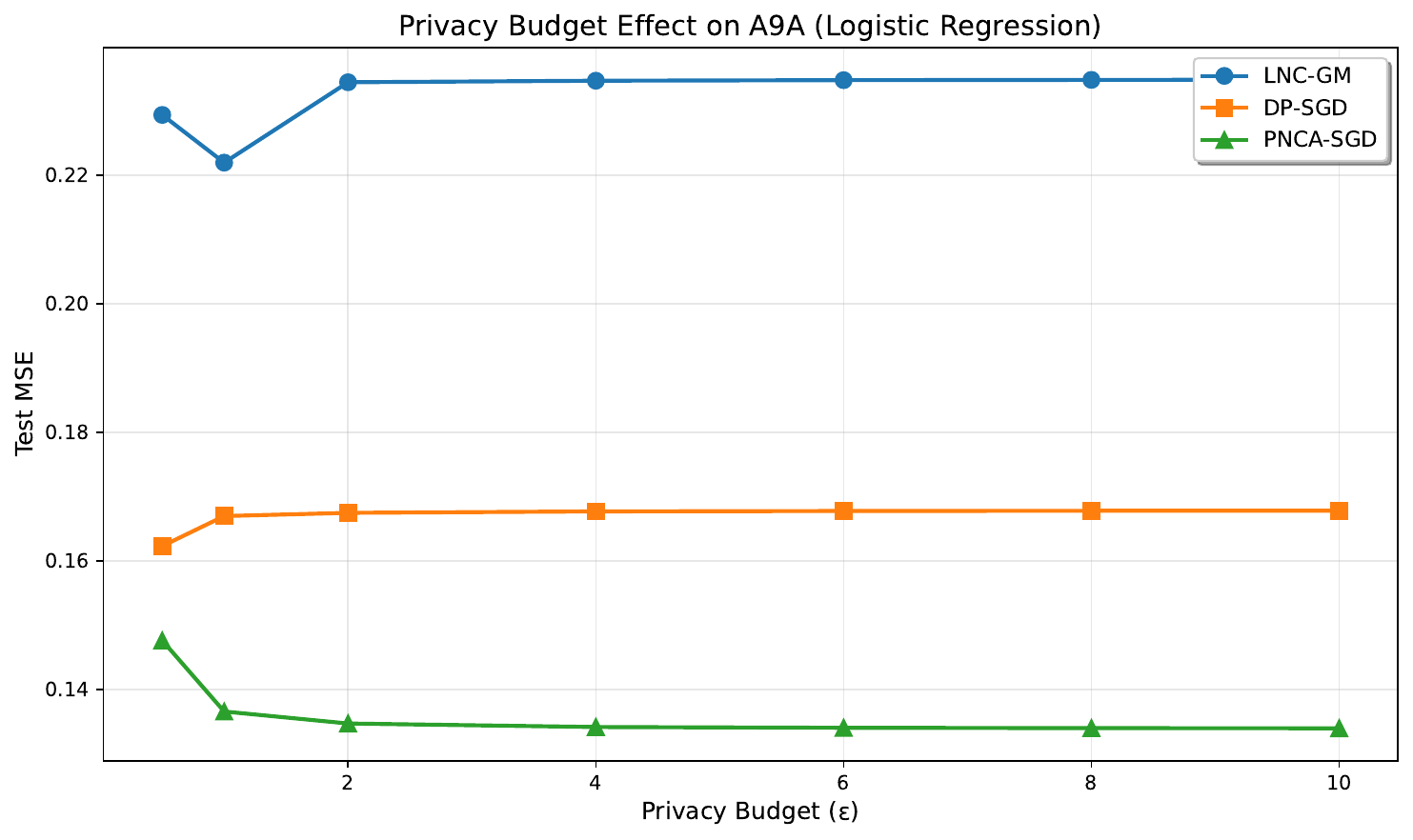}
        \caption{Results of logistic regression with different privacy budget $\varepsilon$ on  a9a}
        \label{fig:loga9aa}
    \end{subfigure}
    \hfill
    \begin{subfigure}{0.45\textwidth}
        \centering
        \includegraphics[width=\textwidth]{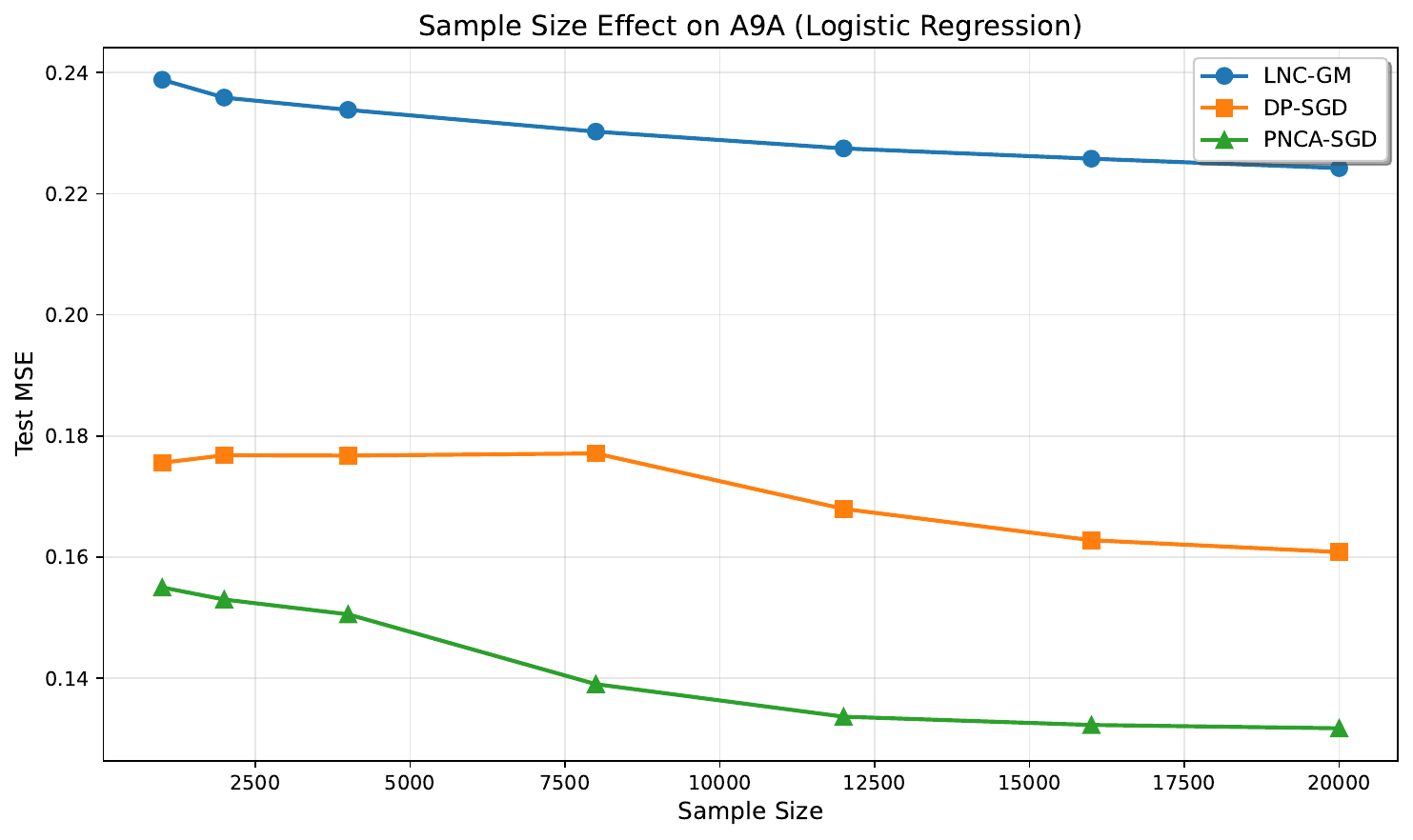}
        \caption{Results of logistic regression with different training sample size on a9a}
        \label{fig:a9ab}
    \end{subfigure}
    \caption{Two experiments on logistic regression with a9a }
    \label{fig:loga9a}
\end{figure}

\begin{figure}[htbp]
    \centering
    \begin{subfigure}{0.45\textwidth}
        \centering
        \includegraphics[width=\textwidth]{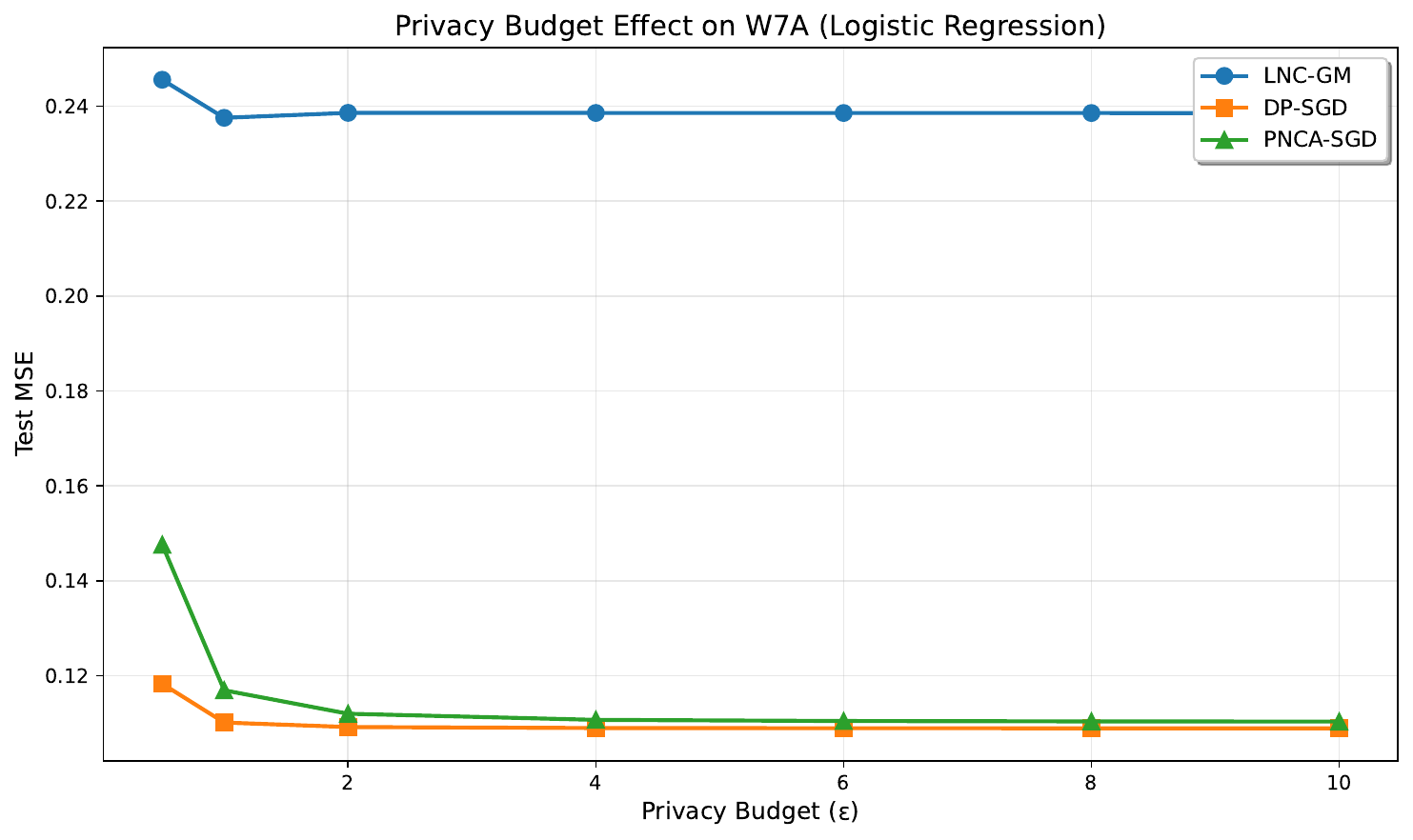}
        \caption{Results of logistic regression with different privacy budget $\varepsilon$ on  w7a}
        \label{fig:logw7aa}
    \end{subfigure}
    \hfill
    \begin{subfigure}{0.45\textwidth}
        \centering
        \includegraphics[width=\textwidth]{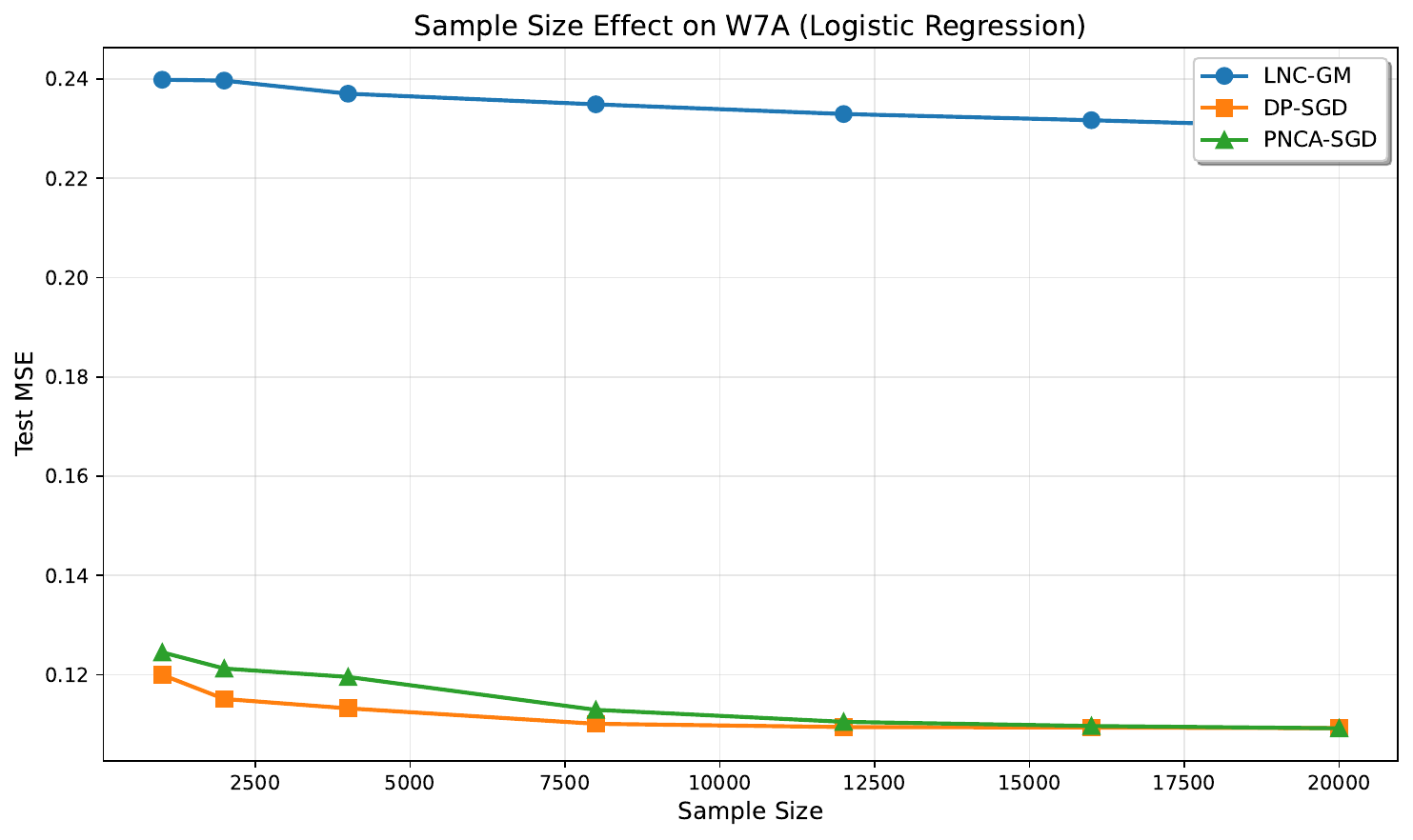}
        \caption{Results of logistic regression with different training sample size on  w7a}
        \label{fig:logw7ab}
    \end{subfigure}
    \caption{Two experiments on logistic regression with w7a }
    \label{fig:logw7a}
\end{figure}

Figure \ref{fig:loga8a} \ref{fig:loga9a} \ref{fig:logw7a} shows the results of $\ell_2$ norm regularized  logistic regression.  DP-SGD is better on the two datasets a8a and a9a, where the gap for logistic regression between DP-SGD and LNC-GM is acceptable. However, we can see through Figure \ref{fig:logw7a}, the performance on w7a of our two methods is better than DP-SGD, both for varying privacy budget $\varepsilon$ and different sample sizes, where PNCA-SGD converges faster than the others given large sample size as expected.  Although LNC-GM does not achieve the same convergence rate as DP-SGD, the trade-off is acceptable given that our approach operates under considerably more relaxed conditions. In contrast, PNCA-SGD not only eliminates the Lipschitz requirement but also demonstrates superior convergence performance compared to DP-SGD. When compared to DP-SGD, PNCA-SGD achieves comparable test MSE performance and demonstrates superior performance given sufficiently large sample sizes.

\section{Conclusion}

% Within this paper, we focus on the problem in Differentially Private Stochastic Convex Optimization with heavy tailed data. We provide the bound without Lipschitz constant using $k$-th moments for Lipschitz loss functions. Another notable contribution of our proposed method is the successful elimination of the Lipschitz continuity requirement for loss functions. Additionally, we introduce the Tsybakov Noise Condition as a generalization framework for our analysis. Through rigorous mathematical derivation presented in the Appendix, we demonstrate the fundamental trade-off between privacy preservation and utility, providing comprehensive insights into the relationship between privacy guarantees and data quality.
In this paper, we address the challenge of DP-SCO with heavy-tailed data. We establish bounds for Lipschitz loss functions using the $k$-th moments, without relying on the Lipschitz constant. A key contribution of our work is the elimination of the Lipschitz requirement for loss functions. Furthermore, we introduce the Tsybakov Noise Condition as a unifying framework for our analysis. We reveal the fundamental trade-off between privacy preservation and utility, offering comprehensive insights into the interplay between privacy guarantees and data quality.

\section*{Acknowledgments}
Di Wang and Difei Xu are supported in part by the funding BAS/1/1689-01-01, URF/1/4663-01-01,  REI/1/5232-01-01,  REI/1/5332-01-01,  and URF/1/5508-01-01  from KAUST, and funding from KAUST - Center of Excellence for Generative AI, under award number 5940.

%\section{Conclusions}
%Whthin this paper, we focus on DP SCO problem under different situations. Firstly, we propose our foundation clipping algorithm, which reaches the bound at $\tilde{O}\left(\widetilde{R}_{2k,n}D( \frac{1}{\sqrt[]{n}}+(\frac{\sqrt[]{d}}{\varepsilon  n})^{\frac{k-1}{k}})\right)$ for $(\varepsilon ,\delta )$-DP. What is notable is that we eliminate the Lipschitz constant from the final result. Then after introducing TNC, we can show that the upper bound turns to be $\tilde{O}\left(\frac{1}{\lambda^{\frac{1}{\theta-1}}} \left(\widetilde{R}_{2k,n}( \frac{1}{\sqrt[]{n}}+\left(\frac{\sqrt[]{d}}{\varepsilon  n}\right)^{\frac{k-1}{k}})\right)^{\frac{\theta}{\theta-1}}\right)$. Next, it is demonstrated that the lower bound $\Omega \left( (\tilde{r}_k((\frac{\sqrt{d}}{\sqrt{\rho}n})^\frac{k-1}{k}))^\frac{\theta}{\theta-1}\right)$ and $\Omega \left( (\frac{\tilde{r}_k }{\sqrt{n}})^\frac{\theta}{\theta-1}\right)$ for $n \geqslant \Omega(\frac{\sqrt{d}}{\sqrt{\rho}})$ and $n \geqslant \Omega({\sqrt{d}})$  respectively. In the second part of the paper,  we can further remove the requirement of the Lipschitz property of the loss functions. After doing permutation for the batches, the accelerated SGD achieves the upper bound $O\left(\widetilde{r}_k D\left[\frac{1}{\sqrt{n}}+\left(\frac{\sqrt{d \log_{} (1/\delta )}}{\varepsilon n}\right)^{\frac{k-1}{k}}\right]\right)$.

% \newpage 
\bibliography{reference}

\begin{thebibliography}{60}
\providecommand{\natexlab}[1]{#1}
\providecommand{\url}[1]{\texttt{#1}}
\expandafter\ifx\csname urlstyle\endcsname\relax
  \providecommand{\doi}[1]{doi: #1}\else
  \providecommand{\doi}{doi: \begingroup \urlstyle{rm}\Url}\fi

\bibitem[Abadi et~al.(2016)Abadi, Chu, Goodfellow, McMahan, Mironov, Talwar, and Zhang]{abadi2016deep}
Martin Abadi, Andy Chu, Ian Goodfellow, H~Brendan McMahan, Ilya Mironov, Kunal Talwar, and Li~Zhang.
\newblock Deep learning with differential privacy.
\newblock In \emph{Proceedings of the 2016 ACM SIGSAC conference on computer and communications security}, pp.\  308--318, 2016.

\bibitem[Acharya et~al.(2021)Acharya, Sun, and Zhang]{acharya2021differentially}
Jayadev Acharya, Ziteng Sun, and Huanyu Zhang.
\newblock Differentially private assouad, fano, and le cam.
\newblock In \emph{Algorithmic Learning Theory}, pp.\  48--78. PMLR, 2021.

\bibitem[Asi et~al.(2021{\natexlab{a}})Asi, Duchi, Fallah, Javidbakht, and Talwar]{asi2021private}
Hilal Asi, John Duchi, Alireza Fallah, Omid Javidbakht, and Kunal Talwar.
\newblock Private adaptive gradient methods for convex optimization.
\newblock In \emph{International Conference on Machine Learning}, pp.\  383--392. PMLR, 2021{\natexlab{a}}.

\bibitem[Asi et~al.(2021{\natexlab{b}})Asi, L{\'e}vy, and Duchi]{asi2021adapting}
Hilal Asi, Daniel L{\'e}vy, and John~C Duchi.
\newblock Adapting to function difficulty and growth conditions in private optimization.
\newblock \emph{Advances in Neural Information Processing Systems}, 34:\penalty0 19069--19081, 2021{\natexlab{b}}.

\bibitem[Asi et~al.(2024)Asi, Liu, and Tian]{asi2024private}
Hilal Asi, Daogao Liu, and Kevin Tian.
\newblock Private stochastic convex optimization with heavy tails: Near-optimality from simple reductions.
\newblock \emph{Advances in Neural Information Processing Systems}, 37:\penalty0 59174--59215, 2024.

\bibitem[Barber \& Duchi(2014)Barber and Duchi]{barber2014privacy}
Rina~Foygel Barber and John~C. Duchi.
\newblock Privacy and statistical risk: Formalisms and minimax bounds.
\newblock \emph{arXiv preprint arXiv:1412.4451}, 2014.

\bibitem[Bassily et~al.(2014)Bassily, Smith, and Thakurta]{bassily2014private}
Raef Bassily, Adam Smith, and Abhradeep Thakurta.
\newblock Private empirical risk minimization: Efficient algorithms and tight error bounds.
\newblock In \emph{2014 IEEE 55th Annual Symposium on Foundations of Computer Science}, pp.\  464--473. IEEE, 2014.

\bibitem[Bassily et~al.(2019)Bassily, Feldman, Talwar, and Thakurta]{bassily2019private}
Raef Bassily, Vitaly Feldman, Kunal Talwar, and Abhradeep Thakurta.
\newblock Private stochastic convex optimization with optimal rates.
\newblock \emph{arXiv preprint arXiv:1908.09970}, 2019.

\bibitem[Biswas et~al.(2007)Biswas, Datta, Fine, and Segal]{biswas2007statistical}
Atanu Biswas, Sujay Datta, Jason~P Fine, and Mark~R Segal.
\newblock \emph{Statistical advances in the biomedical science}.
\newblock Wiley Online Library, 2007.

\bibitem[Boyle \& Dykstra(1986)Boyle and Dykstra]{boyle1986method}
James~P Boyle and Richard~L Dykstra.
\newblock A method for finding projections onto the intersection of convex sets in hilbert spaces.
\newblock In \emph{Advances in order restricted statistical inference}, pp.\  28--47. Springer, 1986.

\bibitem[Brunel \& Avella-Medina(2020)Brunel and Avella-Medina]{brunel2020propose}
Victor-Emmanuel Brunel and Marco Avella-Medina.
\newblock Propose, test, release: Differentially private estimation with high probability.
\newblock \emph{arXiv preprint arXiv:2002.08774}, 2020.

\bibitem[Bun \& Steinke(2016)Bun and Steinke]{bun2016concentrated}
Mark Bun and Thomas Steinke.
\newblock Concentrated differential privacy: Simplifications, extensions, and lower bounds.
\newblock In \emph{Theory of Cryptography Conference}, pp.\  635--658. Springer, 2016.

\bibitem[Chaudhuri et~al.(2011)Chaudhuri, Monteleoni, and Sarwate]{chaudhuri2011differentially}
Kamalika Chaudhuri, Claire Monteleoni, and Anand~D Sarwate.
\newblock Differentially private empirical risk minimization.
\newblock \emph{Journal of Machine Learning Research}, 12\penalty0 (3), 2011.

\bibitem[Chen et~al.(2023)Chen, Ng, and Wang]{chen2023quantizing}
Junren Chen, Michael~K Ng, and Di~Wang.
\newblock Quantizing heavy-tailed data in statistical estimation:(near) minimax rates, covariate quantization, and uniform recovery.
\newblock \emph{IEEE Transactions on Information Theory}, 70\penalty0 (3):\penalty0 2003--2038, 2023.

\bibitem[Ding et~al.(2024)Ding, Lei, Zhu, Wang, Wang, and Xu]{ding2024revisiting}
Meng Ding, Mingxi Lei, Liyang Zhu, Shaowei Wang, Di~Wang, and Jinhui Xu.
\newblock Revisiting differentially private relu regression.
\newblock \emph{Advances in Neural Information Processing Systems}, 37:\penalty0 55470--55506, 2024.

\bibitem[Ding et~al.(2025)Ding, Lei, Wang, Zheng, Wang, and Xu]{dingnearly}
Meng Ding, Mingxi Lei, Shaowei Wang, Tianhang Zheng, Di~Wang, and Jinhui Xu.
\newblock Nearly optimal differentially private relu regression.
\newblock In \emph{The 41st Conference on Uncertainty in Artificial Intelligence}, 2025.

\bibitem[Dwork et~al.(2006)Dwork, McSherry, Nissim, and Smith]{dwork2006calibrating}
Cynthia Dwork, Frank McSherry, Kobbi Nissim, and Adam Smith.
\newblock Calibrating noise to sensitivity in private data analysis.
\newblock In \emph{Theory of cryptography conference}, pp.\  265--284. Springer, 2006.

\bibitem[Dykstra(1983)]{dykstra1983algorithm}
Richard~L Dykstra.
\newblock An algorithm for restricted least squares regression.
\newblock \emph{Journal of the American Statistical Association}, 78\penalty0 (384):\penalty0 837--842, 1983.

\bibitem[Feldman \& Vondrak(2019)Feldman and Vondrak]{feldman2019high}
Vitaly Feldman and Jan Vondrak.
\newblock High probability generalization bounds for uniformly stable algorithms with nearly optimal rate.
\newblock In \emph{Conference on Learning Theory}, pp.\  1270--1279. PMLR, 2019.

\bibitem[Feldman et~al.(2020)Feldman, Koren, and Talwar]{feldman2020private}
Vitaly Feldman, Tomer Koren, and Kunal Talwar.
\newblock Private stochastic convex optimization: optimal rates in linear time.
\newblock In \emph{Proceedings of the 52nd Annual ACM SIGACT Symposium on Theory of Computing}, pp.\  439--449, 2020.

\bibitem[Feldman et~al.(2022)Feldman, McMillan, and Talwar]{feldman2022hiding}
Vitaly Feldman, Audra McMillan, and Kunal Talwar.
\newblock Hiding among the clones: A simple and nearly optimal analysis of privacy amplification by shuffling.
\newblock In \emph{2021 IEEE 62nd Annual Symposium on Foundations of Computer Science (FOCS)}, pp.\  954--964. IEEE, 2022.

\bibitem[Hardt et~al.(2015)Hardt, Recht, and Singer]{hardt2015train}
Moritz Hardt, Benjamin Recht, and Yoram Singer.
\newblock Train faster, generalize better: Stability of stochastic gradient descent.
\newblock \emph{arXiv e-prints}, pp.\  arXiv--1509, 2015.

\bibitem[Hu et~al.(2022)Hu, Ni, Xiao, and Wang]{hu2022high}
Lijie Hu, Shuo Ni, Hanshen Xiao, and Di~Wang.
\newblock High dimensional differentially private stochastic optimization with heavy-tailed data.
\newblock In \emph{Proceedings of the 41st ACM SIGMOD-SIGACT-SIGAI Symposium on Principles of Database Systems}, pp.\  227--236, 2022.

\bibitem[Huai et~al.(2020)Huai, Wang, Miao, Xu, and Zhang]{huai2020pairwise}
Mengdi Huai, Di~Wang, Chenglin Miao, Jinhui Xu, and Aidong Zhang.
\newblock Pairwise learning with differential privacy guarantees.
\newblock In \emph{Proceedings of the AAAI Conference on Artificial Intelligence}, volume~34, pp.\  694--701, 2020.

\bibitem[Ibragimov et~al.(2015)Ibragimov, Ibragimov, and Walden]{ibragimov2015heavy}
Marat Ibragimov, Rustam Ibragimov, and Johan Walden.
\newblock \emph{Heavy-tailed distributions and robustness in economics and finance}, volume 214.
\newblock Springer, 2015.

\bibitem[Kamath et~al.(2020)Kamath, Singhal, and Ullman]{kamath2020private}
Gautam Kamath, Vikrant Singhal, and Jonathan Ullman.
\newblock Private mean estimation of heavy-tailed distributions.
\newblock In \emph{Proceedings of 33rd Conference on Learning Theory (COLT)}, pp.\  2204--2235, 2020.

\bibitem[Kamath et~al.(2021)Kamath, Liu, and Zhang]{kamath2021improved}
Gautam Kamath, Xingtu Liu, and Huanyu Zhang.
\newblock Improved rates for differentially private stochastic convex optimization with heavy-tailed data.
\newblock \emph{arXiv preprint arXiv:2106.01336}, 2021.

\bibitem[Kamath et~al.(2022)Kamath, Liu, and Zhang]{kamath2022improved}
Gautam Kamath, Xingtu Liu, and Huanyu Zhang.
\newblock Improved rates for differentially private stochastic convex optimization with heavy-tailed data.
\newblock In \emph{International Conference on Machine Learning}, pp.\  10633--10660. PMLR, 2022.

\bibitem[Karimi et~al.(2016)Karimi, Nutini, and Schmidt]{karimi2016linear}
Hamed Karimi, Julie Nutini, and Mark Schmidt.
\newblock Linear convergence of gradient and proximal-gradient methods under the polyak-{\l}ojasiewicz condition.
\newblock In \emph{Joint European Conference on Machine Learning and Knowledge Discovery in Databases}, pp.\  795--811. Springer, 2016.

\bibitem[Kasiviswanathan \& Jin(2016)Kasiviswanathan and Jin]{kasiviswanathan2016efficient}
Shiva~Prasad Kasiviswanathan and Hongxia Jin.
\newblock Efficient private empirical risk minimization for high-dimensional learning.
\newblock In \emph{International Conference on Machine Learning}, pp.\  488--497, 2016.

\bibitem[Kifer et~al.(2012)Kifer, Smith, and Thakurta]{kifer2012private}
Daniel Kifer, Adam Smith, and Abhradeep Thakurta.
\newblock Private convex empirical risk minimization and high-dimensional regression.
\newblock In \emph{Conference on Learning Theory}, pp.\  25--1, 2012.

\bibitem[Koren \& Levy(2015)Koren and Levy]{koren2015fast}
Tomer Koren and Kfir~Y Levy.
\newblock Fast rates for exp-concave empirical risk minimization.
\newblock In \emph{NIPS}, pp.\  1477--1485, 2015.

\bibitem[Liu et~al.(2018)Liu, Zhang, Zhang, Jin, and Yang]{liu2018fast}
Mingrui Liu, Xiaoxuan Zhang, Lijun Zhang, Rong Jin, and Tianbao Yang.
\newblock Fast rates of erm and stochastic approximation: Adaptive to error bound conditions.
\newblock \emph{arXiv preprint arXiv:1805.04577}, 2018.

\bibitem[Liu et~al.(2021)Liu, Kong, Kakade, and Oh]{liu2021robust}
Xiyang Liu, Weihao Kong, Sham Kakade, and Sewoong Oh.
\newblock Robust and differentially private mean estimation.
\newblock \emph{arXiv preprint arXiv:2102.09159}, 2021.

\bibitem[Lowy \& Razaviyayn(2023)Lowy and Razaviyayn]{lowy2023private}
Andrew Lowy and Meisam Razaviyayn.
\newblock Private stochastic optimization with large worst-case lipschitz parameter: Optimal rates for (non-smooth) convex losses and extension to non-convex losses.
\newblock In \emph{International Conference on Algorithmic Learning Theory}, pp.\  986--1054. PMLR, 2023.

\bibitem[McSherry(2009)]{mcsherry2009privacy}
Frank~D McSherry.
\newblock Privacy integrated queries: an extensible platform for privacy-preserving data analysis.
\newblock In \emph{Proceedings of the 2009 ACM SIGMOD International Conference on Management of data}, pp.\  19--30, 2009.

\bibitem[Ramdas \& Singh(2012)Ramdas and Singh]{ramdas2012optimal}
Aaditya Ramdas and Aarti Singh.
\newblock Optimal rates for first-order stochastic convex optimization under tsybakov noise condition.
\newblock \emph{arXiv preprint arXiv:1207.3012}, 2012.

\bibitem[Ramdas \& Singh(2013)Ramdas and Singh]{ramdas2013algorithmic}
Aaditya Ramdas and Aarti Singh.
\newblock Algorithmic connections between active learning and stochastic convex optimization.
\newblock In \emph{International Conference on Algorithmic Learning Theory}, pp.\  339--353. Springer, 2013.

\bibitem[Smith et~al.(2017)Smith, Thakurta, and Upadhyay]{smith2017interaction}
Adam Smith, Abhradeep Thakurta, and Jalaj Upadhyay.
\newblock Is interaction necessary for distributed private learning?
\newblock In \emph{2017 IEEE Symposium on Security and Privacy (SP)}, pp.\  58--77. IEEE, 2017.

\bibitem[Su et~al.(2022)Su, Hu, and Wang]{su2022faster}
Jinyan Su, Lijie Hu, and Di~Wang.
\newblock Faster rates of private stochastic convex optimization.
\newblock In \emph{International Conference on Algorithmic Learning Theory}, pp.\  995--1002. PMLR, 2022.

\bibitem[Su et~al.(2023)Su, Zhao, and Wang]{su2023differentially}
Jinyan Su, Changhong Zhao, and Di~Wang.
\newblock Differentially private stochastic convex optimization in (non)-euclidean space revisited.
\newblock In \emph{Uncertainty in Artificial Intelligence}, pp.\  2026--2035. PMLR, 2023.

\bibitem[Su et~al.(2024)Su, Hu, and Wang]{su2024faster}
Jinyan Su, Lijie Hu, and Di~Wang.
\newblock Faster rates of differentially private stochastic convex optimization.
\newblock \emph{Journal of Machine Learning Research}, 25\penalty0 (114):\penalty0 1--41, 2024.

\bibitem[Tao et~al.(2022{\natexlab{a}})Tao, Wu, Cheng, and 0015]{tao2022private}
Youming Tao, Yulian Wu, Xiuzhen Cheng, and Di~Wang 0015.
\newblock Private stochastic convex optimization and sparse learning with heavy-tailed data revisited.
\newblock In \emph{IJCAI}, pp.\  3947--3953, 2022{\natexlab{a}}.

\bibitem[Tao et~al.(2022{\natexlab{b}})Tao, Wu, Zhao, and Wang]{tao2022optimal}
Youming Tao, Yulian Wu, Peng Zhao, and Di~Wang.
\newblock Optimal rates of (locally) differentially private heavy-tailed multi-armed bandits.
\newblock In \emph{International Conference on Artificial Intelligence and Statistics}, pp.\  1546--1574. PMLR, 2022{\natexlab{b}}.

\bibitem[Tian et~al.(2025)Tian, Ding, Tao, Xiang, and Wang]{tian2025differentially}
Xizhi Tian, Meng Ding, Touming Tao, Zihang Xiang, and Di~Wang.
\newblock Differentially private sparse linear regression with heavy-tailed responses.
\newblock \emph{arXiv preprint arXiv:2506.06861}, 2025.

\bibitem[van Erven et~al.(2015)van Erven, Gr{\"u}nwald, Mehta, Reid, and Williamson]{van2015fast}
Tim van Erven, Peter~D Gr{\"u}nwald, Nishant~A Mehta, Mark~D Reid, and Robert~C Williamson.
\newblock Fast rates in statistical and online learning.
\newblock \emph{Journal of Machine Learning Research}, 16:\penalty0 1793--1861, 2015.

\bibitem[Wang \& Xu(2022)Wang and Xu]{wang2022differentially}
Di~Wang and Jinhui Xu.
\newblock Differentially private $\ell_1$-norm linear regression with heavy-tailed data.
\newblock In \emph{2022 IEEE International Symposium on Information Theory (ISIT)}, pp.\  1856--1861. IEEE, 2022.

\bibitem[Wang \& Xu(2025)Wang and Xu]{wang2025private}
Di~Wang and Jinhui Xu.
\newblock Private least absolute deviations with heavy-tailed data.
\newblock \emph{Theoretical Computer Science}, 1030:\penalty0 115071, 2025.

\bibitem[Wang et~al.(2017)Wang, Ye, and Xu]{wang2017differentially}
Di~Wang, Minwei Ye, and Jinhui Xu.
\newblock Differentially private empirical risk minimization revisited: Faster and more general.
\newblock In \emph{Advances in Neural Information Processing Systems}, pp.\  2722--2731, 2017.

\bibitem[Wang et~al.(2018)Wang, Gaboardi, and Xu]{wang2018empirical}
Di~Wang, Marco Gaboardi, and Jinhui Xu.
\newblock Empirical risk minimization in non-interactive local differential privacy revisited.
\newblock In \emph{Advances in Neural Information Processing Systems}, pp.\  965--974, 2018.

\bibitem[Wang et~al.(2019{\natexlab{a}})Wang, Chen, and Xu]{wang2019differentially}
Di~Wang, Changyou Chen, and Jinhui Xu.
\newblock Differentially private empirical risk minimization with non-convex loss functions.
\newblock In \emph{International Conference on Machine Learning}, pp.\  6526--6535, 2019{\natexlab{a}}.

\bibitem[Wang et~al.(2019{\natexlab{b}})Wang, Smith, and Xu]{wang2019noninteractive}
Di~Wang, Adam Smith, and Jinhui Xu.
\newblock Noninteractive locally private learning of linear models via polynomial approximations.
\newblock In \emph{Algorithmic Learning Theory}, pp.\  897--902, 2019{\natexlab{b}}.

\bibitem[Wang et~al.(2020)Wang, Xiao, Devadas, and Xu]{wang2020differentially}
Di~Wang, Hanshen Xiao, Srinivas Devadas, and Jinhui Xu.
\newblock On differentially private stochastic convex optimization with heavy-tailed data.
\newblock In \emph{International Conference on Machine Learning}, pp.\  10081--10091. PMLR, 2020.

\bibitem[Woolson \& Clarke(2011)Woolson and Clarke]{woolson2011statistical}
Robert~F Woolson and William~R Clarke.
\newblock \emph{Statistical methods for the analysis of biomedical data}, volume 371.
\newblock John Wiley \& Sons, 2011.

\bibitem[Wu et~al.(2017)Wu, Li, Kumar, Chaudhuri, Jha, and Naughton]{wu2017bolt}
Xi~Wu, Fengan Li, Arun Kumar, Kamalika Chaudhuri, Somesh Jha, and Jeffrey Naughton.
\newblock Bolt-on differential privacy for scalable stochastic gradient descent-based analytics.
\newblock In \emph{Proceedings of the 2017 ACM International Conference on Management of Data}, pp.\  1307--1322. ACM, 2017.

\bibitem[Wu et~al.(2023)Wu, Zhou, Chowdhury, and Wang]{wu2023differentially}
Yulian Wu, Xingyu Zhou, Sayak~Ray Chowdhury, and Di~Wang.
\newblock Differentially private episodic reinforcement learning with heavy-tailed rewards.
\newblock In \emph{International Conference on Machine Learning}, pp.\  37880--37918. PMLR, 2023.

\bibitem[Wu et~al.(2024)Wu, Zhou, Tao, and Wang]{wu2024private}
Yulian Wu, Xingyu Zhou, Youming Tao, and Di~Wang.
\newblock On private and robust bandits.
\newblock \emph{Advances in Neural Information Processing Systems}, 36, 2024.

\bibitem[Xu et~al.(2017)Xu, Lin, and Yang]{xu2017stochastic}
Yi~Xu, Qihang Lin, and Tianbao Yang.
\newblock Stochastic convex optimization: Faster local growth implies faster global convergence.
\newblock In \emph{International Conference on Machine Learning}, pp.\  3821--3830. PMLR, 2017.

\bibitem[Xue et~al.(2021)Xue, Yang, Huai, and Wang]{xue2021differentially}
Zhiyu Xue, Shaoyang Yang, Mengdi Huai, and Di~Wang.
\newblock Differentially private pairwise learning revisited.
\newblock In \emph{30th International Joint Conference on Artificial Intelligence, IJCAI 2021}, pp.\  3242--3248. International Joint Conferences on Artificial Intelligence Organization, 2021.

\bibitem[Yang et~al.(2018)Yang, Li, and Zhang]{yang2018simple}
Tianbao Yang, Zhe Li, and Lijun Zhang.
\newblock A simple analysis for exp-concave empirical minimization with arbitrary convex regularizer.
\newblock In \emph{International Conference on Artificial Intelligence and Statistics}, pp.\  445--453. PMLR, 2018.

\end{thebibliography}
\bibliographystyle{tmlr}

\appendix
\section{Experimental Results}
% l4r regression
% the first graph
\begin{figure}[htbp]
    \centering
    \begin{subfigure}{0.45\textwidth}
        \centering
        \includegraphics[width=\textwidth]{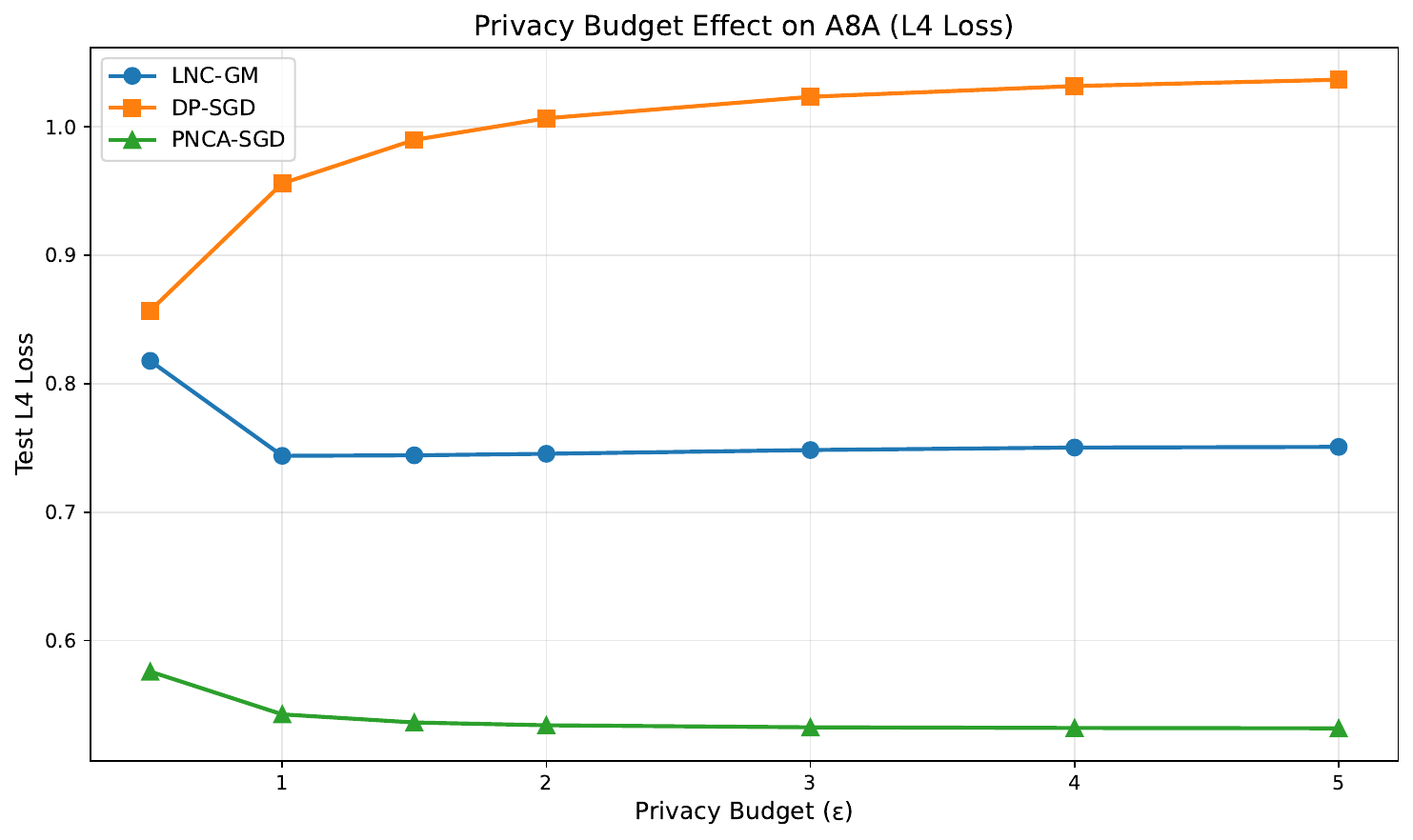}
        \caption{Results of $\ell_4$-norm  linear regression with different privacy budget $\varepsilon$ on  a8a}
        \label{fig:a8aa}
    \end{subfigure}
    \hfill
    \begin{subfigure}{0.45\textwidth}
        \centering
        \includegraphics[width=\textwidth]{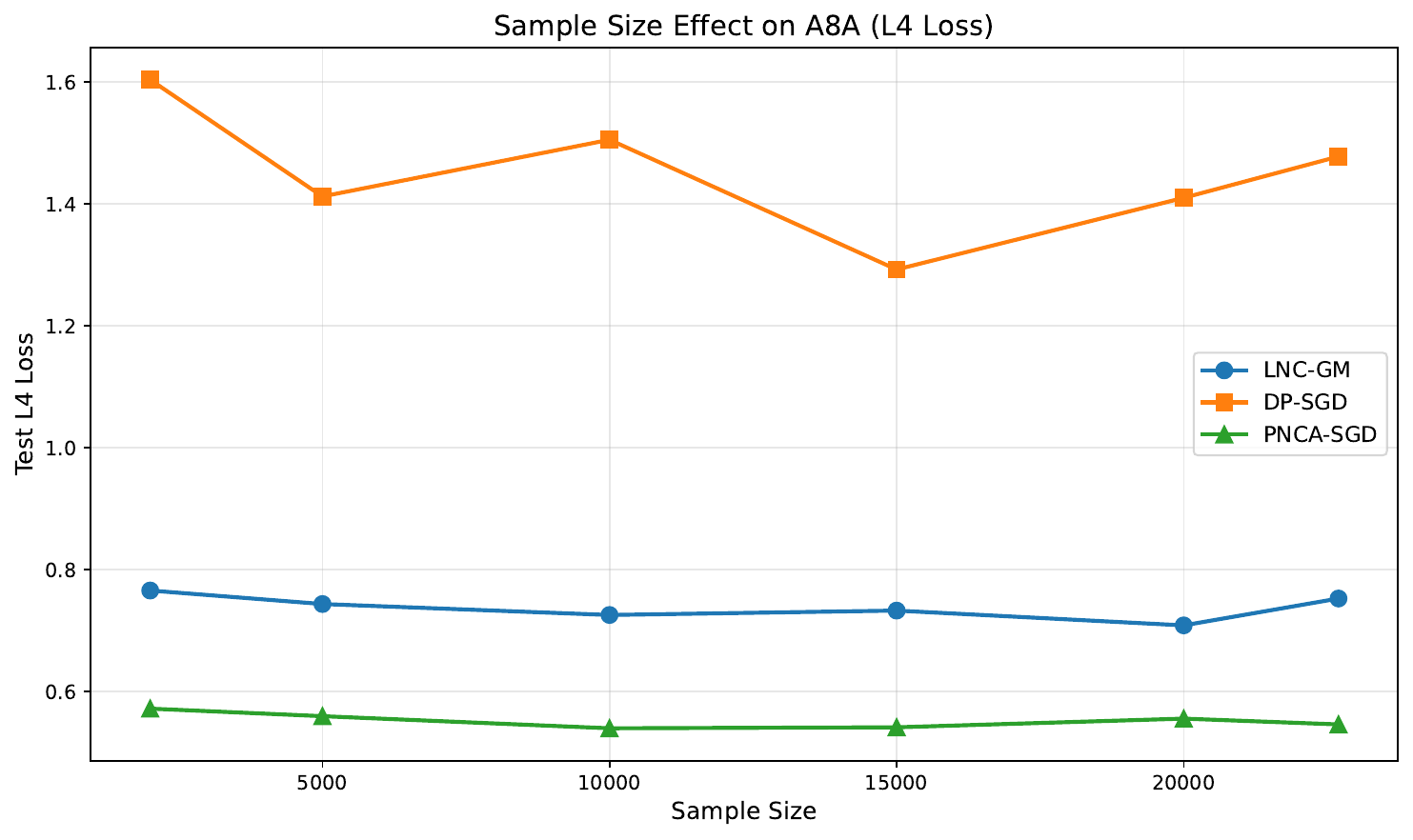}
        \caption{Results of $\ell_4$-norm linear regression with different training sample size on  a8a}
        \label{fig:a8ab}
    \end{subfigure}
    \caption{Two experiments on $\ell_4$-norm linear regression with a8a }
    \label{fig:a8a}
\end{figure}

% the second graph
\begin{figure}[htbp]
    \centering
    \begin{subfigure}{0.45\textwidth}
        \centering
        \includegraphics[width=\textwidth]{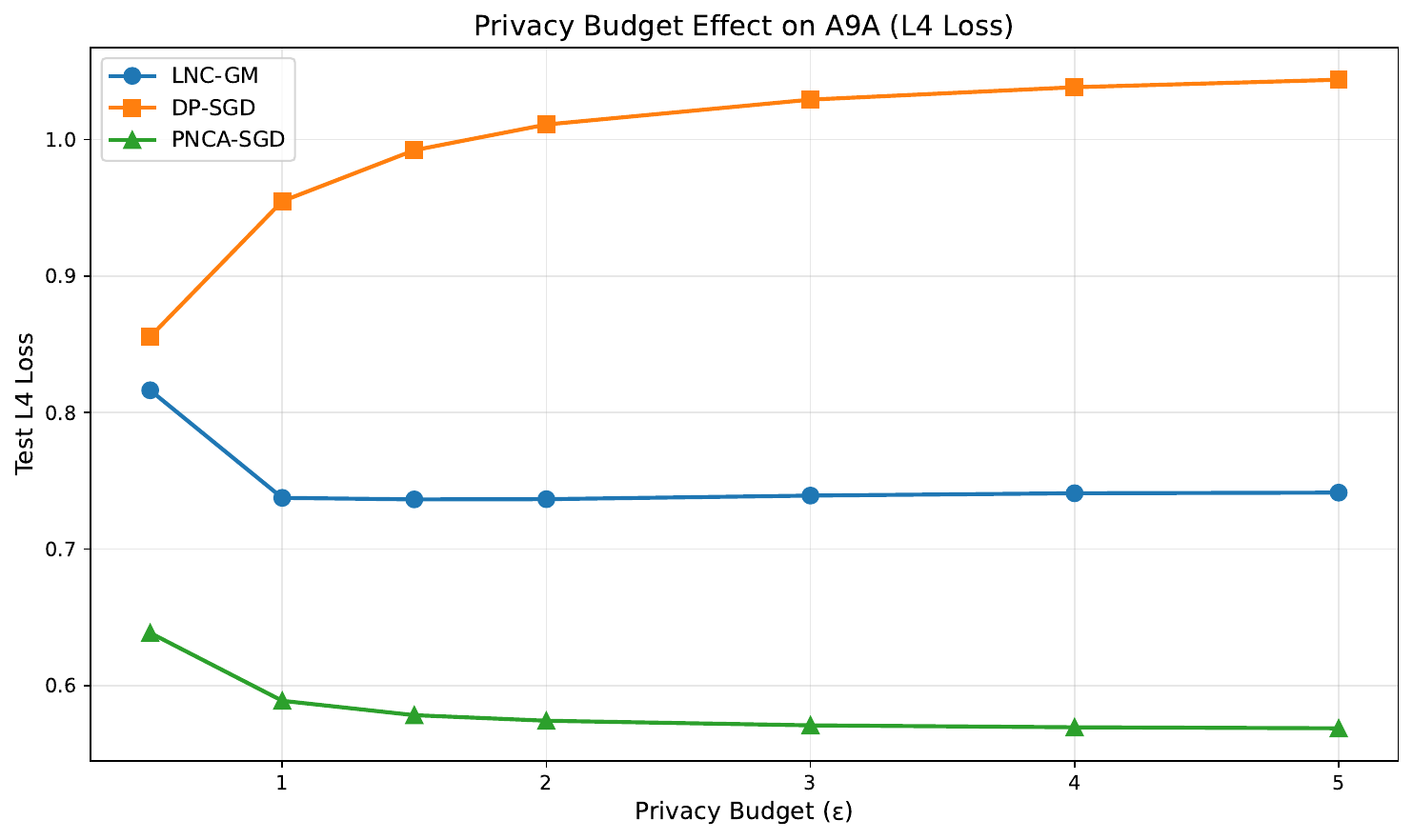}
        \caption{Results of $\ell_4$-norm linear regression with different privacy budget $\varepsilon$ on  a9a}
        \label{fig:a9aa}
    \end{subfigure}
    \hfill
    \begin{subfigure}{0.45\textwidth}
        \centering
        \includegraphics[width=\textwidth]{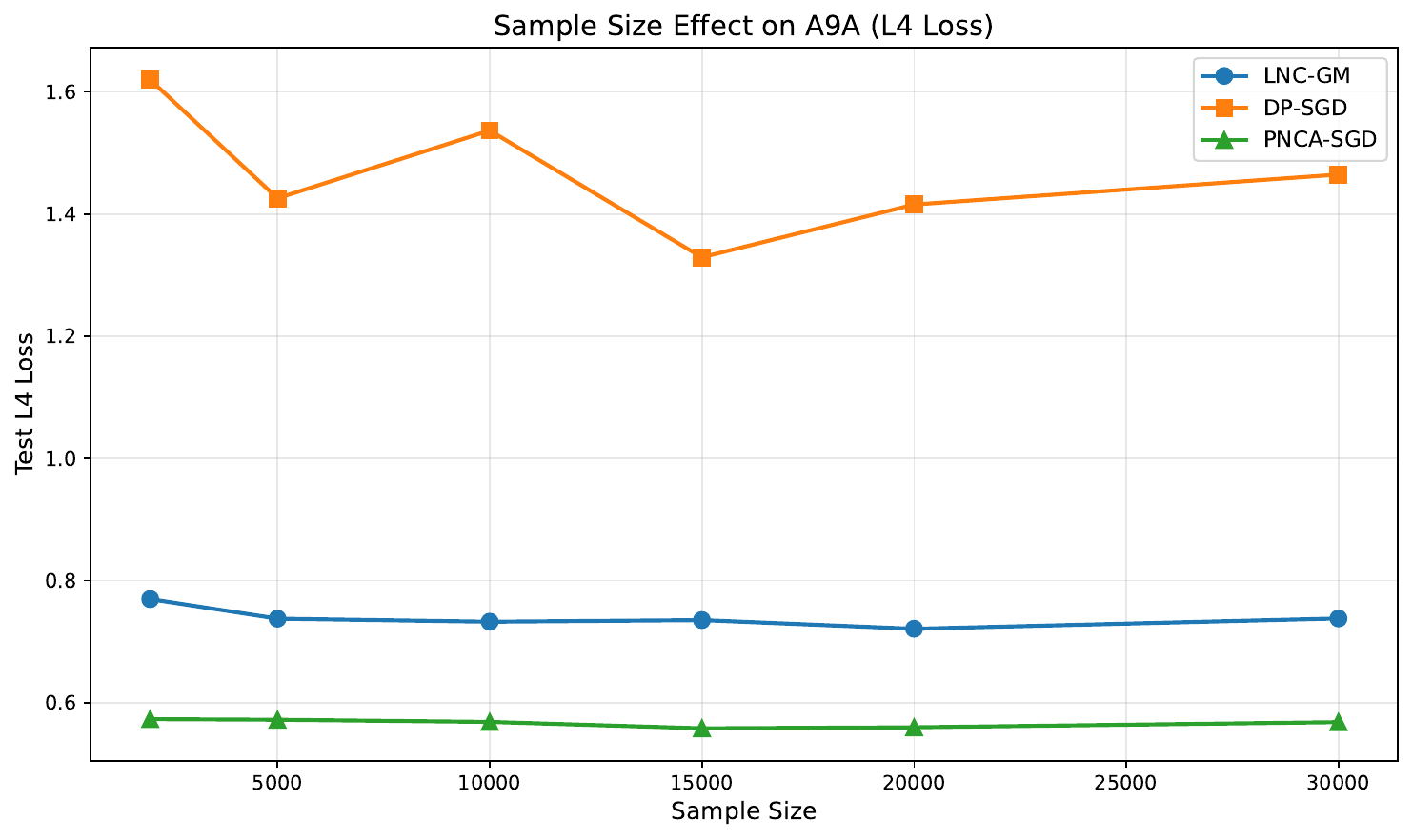}
        \caption{Results of $\ell_4$-norm linear regression with different training sample size on a9a}
        \label{fig:a9ab}
    \end{subfigure}
    \caption{Two experiments on $\ell_4$-norm  linear regression with a9a }
    \label{fig:a9a}
\end{figure}

% the third graph
\begin{figure}[htbp]
    \centering
    \begin{subfigure}{0.45\textwidth}
        \centering
        \includegraphics[width=\textwidth]{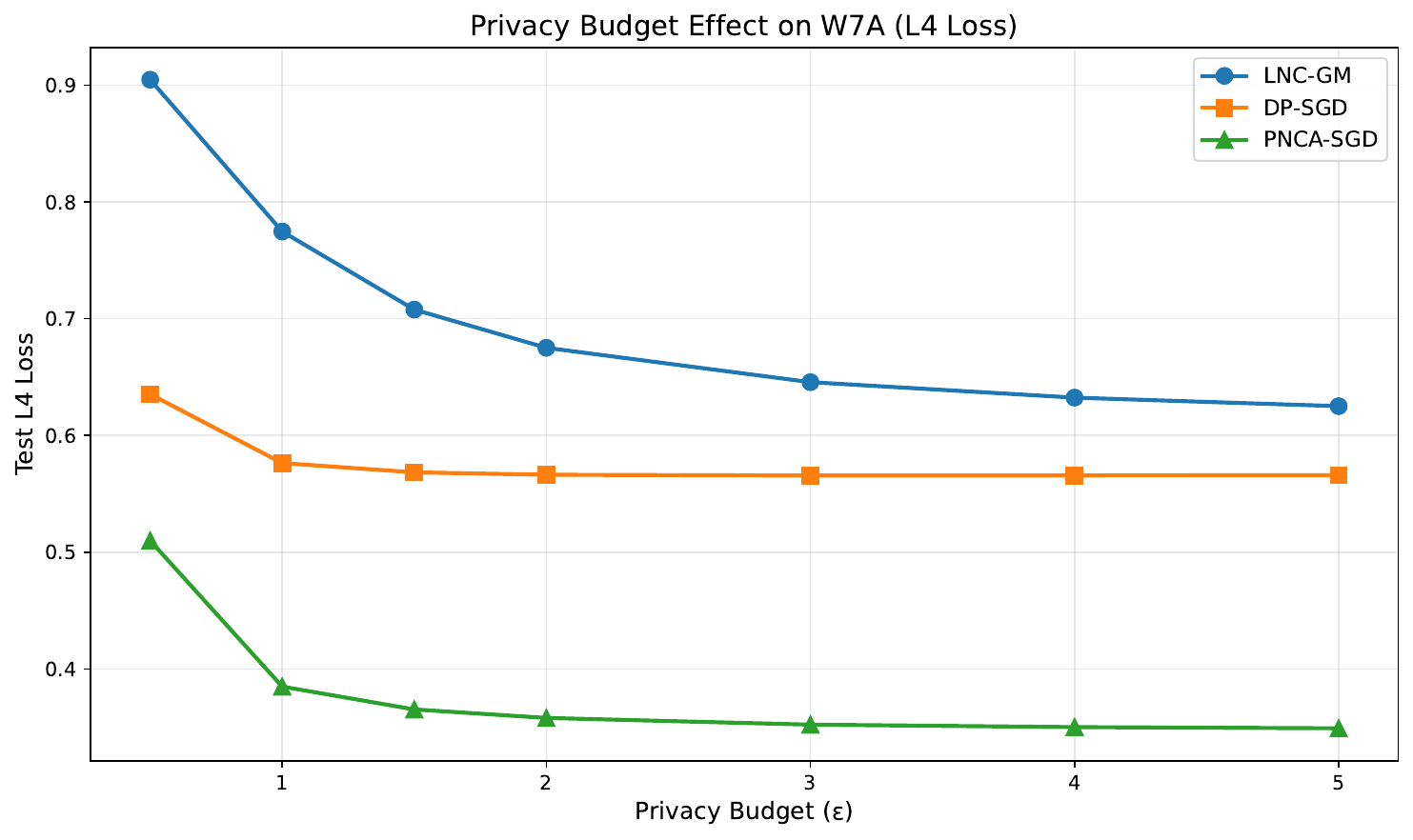}
        \caption{Results of $\ell_4$-norm linear regression with different privacy budget $\varepsilon$ on  w7a}
        \label{fig:w7aa}
    \end{subfigure}
    \hfill
    \begin{subfigure}{0.45\textwidth}
        \centering
        \includegraphics[width=\textwidth]{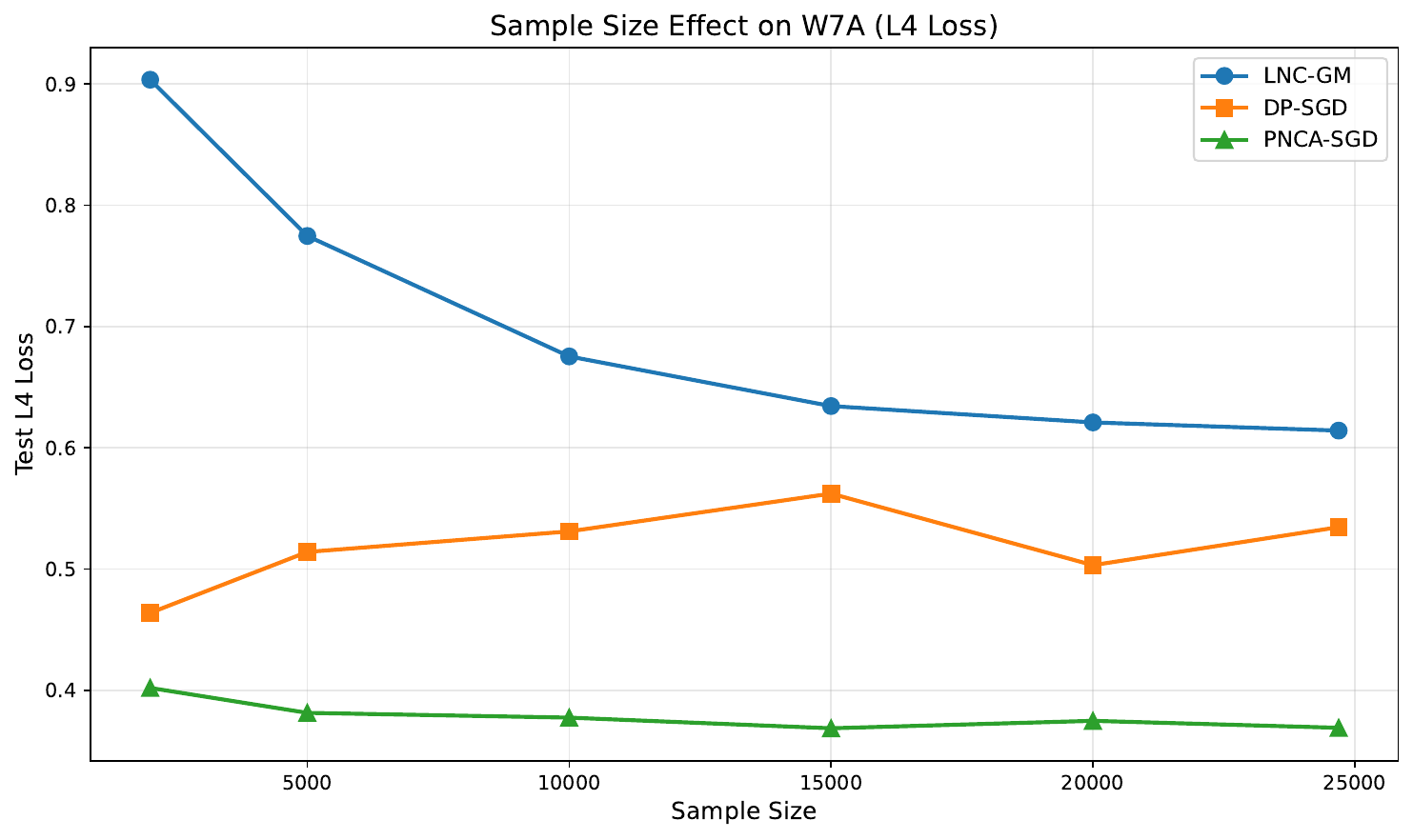}
        \caption{Results of $\ell_4$-norm linear regression with different training sample size on  w7a}
        \label{fig:w7ab}
    \end{subfigure}
    \caption{Two experiments on $\ell_4$-norm linear regression with w7a }
    \label{fig:w7a}
\end{figure}

% logistic regression:

% the second graph

% the third graph
\section{Omitted Proof}
\begin{proof}[{\bf Proof of Lemma \ref{lem:2}}]
    Let $l\in \mathbb{N}$, and $n=2^l$ and consider 
    \begin{equation}\nonumber
        \begin{aligned}
            \widehat{r}_{n}(X)^{(k)}& =\frac{1}{n}\sup_{w}\left(\sum_{i=1}^{n/2}\|\nabla f(w,x_i)\|^k+\sum_{i=n/2+1}^{n}\|\nabla f(w,x_i)\|^k\right)  \\
            &\leqslant\frac{1}{n}\left(\sup_{w}\sum_{i=1}^{n/2}\|\nabla f(w,x_{i})\|^{k}+\sup_{w}\sum_{i=n/2+1}^{n}\|\nabla f(w,x_{i})\|^{k}\right).
            \end{aligned}
    \end{equation}
    Taking expectations over the random draw of $X\sim \mathcal{D}^n$ and we have $\widetilde{e}_n^{(k)} \leqslant \widetilde{e}_{n/2}^{(k)}$. Thus, $\widetilde{R}_{k,n} \leqslant \widetilde{r}_k.$ 
\end{proof}
\begin{proof}[{\bf Proof of Theorem \ref{thm:1}}] 

{\bf Privacy.} Since in each epoch of  Algorithm \ref{alg:3}  we use a disjoint dataset, it is sufficient for us to show each $w_i$ is $(\varepsilon, \delta)$-DP. 

  Since the batches $\{B_i\}_{i=1}^l$ are disjoint, it suffices (by parallel composition in \citep{mcsherry2009privacy} to show that $w_i$ (produced by $T_i$ iterations of Algorithm 2 in line 6 of Algorithm 3) is $\frac{\varepsilon^2}{2}$-zCDP for all $i \in [l]$, hence by Proposition 1.3 in \citep{bun2016concentrated}, then it is $(2 \varepsilon \sqrt{\log(1/\delta)},\delta)$-DP

With clip threshold $C_i$ and batch size $n_i$, the $\ell_2$ sensitivity of the clipped subgradient update is bounded by

\begin{equation}
\Delta = \sup_{w,x\sim x'} \left\| \frac{1}{n_i} \sum_{j=1}^{n_i} \Pi_{C_i} (\nabla f(w, x_j)) - \Pi_{C_i} (\nabla f(w, x'_j)) \right\| = \frac{1}{n_i} \sup_{w,x,x'} \| \Pi_{C_i} (\nabla f(w, x)) - \Pi_{C_i} (\nabla f(w, x')) \| \leq \frac{2C_i}{n_i}.
\end{equation}

Note that the terms arising from regularization cancel out. Thus,  conditional on the previous updates $w_{1:i}$, the $(i+1)$-st update in line 3 of Algorithm 2 satisfies $\frac{\varepsilon^2}{2T_i}$-zCDP. Hence, Lemma 2.3 in \citep{bun2016concentrated} implies that $w_i$ (in line 6 of Algorithm 3) is $\frac{\varepsilon^2}{2}$-zCDP, hence $(2 \varepsilon \sqrt{\log(1/\delta)},\delta)$-DP. By the assumption that $\varepsilon \leqslant \sqrt{\log (1/\delta)}$, the mechanism is $(\varepsilon^2,\delta)$-DP, by taking $\varepsilon' = \epsilon^2$, we can claim that the algorithm is $(\varepsilon',\delta)$-DP.

{\bf Excess risk:} We finish our proof through several parts. We first recall the following lemma. 
    \begin{lemma}\label{7} [\citep{feldman2019high}]
    Assume $\operatorname{diam}_2(\mathcal{X}) \leqslant D$. Let $\mathcal{S}=\left(S_1, \ldots, S_n\right)$ where $S_1^n \stackrel{\text { iid }}{\sim} P$ and $f(w,x)$ is L-Lipschitz and $\lambda$-strongly convex for all $x \in \mathcal{X}$. Let $\hat{x}=\operatorname{argmin}_{x \in \mathcal{X}} \bar{F}(w)$ be the empirical minimizer. For $0<\beta \leqslant 1 / n$, with probability at least $1-\beta$
    $$
    F(\hat{x})-F\left(x^{\star}\right) \leqslant \frac{c L^2 \log (n) \log (1 / \beta)}{\lambda n}+\frac{c L D \sqrt{\log (1 / \beta)}}{\sqrt{n}} .
    $$
\end{lemma}
\begin{theorem}
    We have the following bound for $\left\| w_T-\hat{w} \right\| ^2$ for $T$ iterations:
    \begin{equation}\nonumber
        \left\| w_T-\hat{w} \right\| ^2 \leqslant \exp\{-\frac{\lambda \eta T}{2} \}\left\| w_0-\widehat{w} \right\|^2 + \frac{8\eta \hat{r}_n^2(x)}{\lambda }+8\eta\lambda D^2+\frac{20 \widehat{B}^2}{\lambda ^2}.
    \end{equation}
\end{theorem}
\begin{proof}
    % Suppose that we could find some $M_F$ to bound the gradient, i.e. $\|\widetilde{\nabla  }F_{\lambda }\| \leqslant M_F$ and similarly, $\left\| w_k-\widehat{w} \right\| \leqslant M_D$ with probability at least $1-\theta $.
Detailly, 
\begin{equation}\nonumber
\begin{aligned}
\left\|\tilde{\nabla} F_\lambda\left(w_t\right)\right\|^2 & \leqslant 2\left(\left\|\nabla \widehat{F}_\lambda\left(w_t\right)\right\|^2+\left\|b_t\right\|^2\right) \\
& \leqslant 2\left(2 \widehat{r}_n(X)^2+2 \lambda^2 D^2+\hat{B}^2\right),
\end{aligned}
\end{equation}
And also, by Young's inequality,
\begin{equation}\nonumber
    \left| \langle b_t,w_t-\hat{w} \rangle  \right|  \leqslant \frac{\widehat{B}^2}{\lambda }+\frac{\lambda }{4}\left\| w_t-\hat{w} \right\| ^2.
\end{equation}

Set $\widetilde{\nabla} F_\lambda\left(w_t\right)=\nabla \hat{F}_\lambda\left(w_t\right)+b_t=$ $\frac{1}{n} \sum_{i=1}^n \Pi_C\left(\nabla f\left(w, x_i\right)\right)+\lambda\left(w-w_0\right)$ as the biased, noisy subgradients of the regularized empirical loss in Algorithm \ref{alg:3} , with $N_t \sim \mathcal{N}\left(0, \sigma^2 \mathbf{I}_d\right)$ and $b_t=\frac{1}{n} \sum_{i=1}^n \Pi_C\left(\nabla f\left(w_t, x_i\right)\right)-$ $\frac{1}{n} \sum_{i=1}^n \nabla f\left(w_t, x_i\right)$. Denote $y_{t+1}=w_t-\eta \tilde{\nabla} F_\lambda\left(w_t\right)$, so that $w_{t+1}=\Pi_{\mathcal{W}}\left(y_{t+1}\right)$. For now, by strong convexity, we have
$$
\begin{aligned}
\widehat{F}_\lambda\left(w_t\right)-\widehat{F}_\lambda(\hat{w}) \leqslant & \left\langle\nabla \widehat{F}_\lambda\left(w_t\right), w_t-\hat{w}\right\rangle-\frac{\lambda}{2}\left\|w_t-\hat{w}\right\|^2 \\
= & \left\langle\widetilde{\nabla} F_\lambda\left(w_t\right), w_t-\hat{w}\right\rangle-\frac{\lambda}{2}\left\|w_t-\hat{w}\right\|^2+\left\langle\nabla \hat{F}_\lambda\left(w_t\right)-\widetilde{\nabla} F_\lambda\left(w_t\right), w_t-\hat{w}\right\rangle \\
= & \frac{1}{2 \eta}\left(\left\|w_t-\hat{w}\right\|^2+\left\|w_t-y_{t+1}\right\|^2-\left\|y_{t+1}-\hat{w}\right\|^2\right)-\frac{\lambda}{2}\left\|w_t-\hat{w}\right\|^2 \\
& +\left\langle\nabla \hat{F}_\lambda\left(w_t\right)-\tilde{\nabla} F_\lambda\left(w_t\right), w_t-\hat{w}\right\rangle \\
= & \frac{1}{2 \eta}\left(\left\|w_t-\hat{w}\right\|^2(1-\lambda \eta)-\left\|y_{t+1}-\hat{w}\right\|^2\right)+\frac{\eta}{2}\left\|\widetilde{\nabla} F_\lambda\left(w_t\right)\right\|^2 \\
& +\left\langle\nabla \widehat{F}_\lambda\left(w_t\right)-\tilde{\nabla} F_\lambda\left(w_t\right), w_t-\hat{w}\right\rangle \\
\leqslant & \frac{1}{2 \eta}\left(\left\|w_t-\hat{w}\right\|^2(1-\lambda \eta)-\left\|w_{t+1}-\hat{w}\right\|^2\right)+\frac{\eta}{2}\left\|\widetilde{\nabla} F_\lambda\left(w_t\right)\right\|^2-\left\langle b_t, w_t-\hat{w}\right\rangle,
\end{aligned}
$$
where we used non-expansiveness of projection and the definition of $\tilde{\nabla} F_\lambda\left(w_t\right)$ in the last line. Now, re-arranging this inequality,

\begin{equation}\nonumber
    \begin{aligned}
        \left\|w_{t+1}-\hat{w}\right\|^2 & \leqslant \left\|w_t-\hat{w}\right\|^2(1-\lambda \eta)+\eta ^2 \left\| \widetilde{\nabla }F_{\lambda }(w_t) \right\|^2-2 \eta \langle b_t,w_t-w\rangle-2 \eta  (\widehat{F}_{\lambda }(w_t)-\widehat{F}_{\lambda }(\widehat{w}))\\
        &\leqslant \left\|w_t-\hat{w}\right\|^2(1-\lambda \eta)+\eta ^2 \left\| \widetilde{\nabla }F_{\lambda }(w_t) \right\|^2-2 \eta \langle b_t,w_t-w\rangle\\
        &\leqslant \left\|w_t-\hat{w}\right\|^2(1-\frac{\lambda \eta}{2})+\eta ^2 \cdot  2(2\hat{r}_n^2(x)+2\lambda ^2D^2+\widehat{B}^2) +\frac{2 \eta \widehat{B}^2}{\lambda }\\
        \leqslant &\left\|w_t-\hat{w}\right\|^2(1-\frac{\lambda \eta}{2})+4\eta ^2(\hat{r}_n^2(x)+\lambda ^2D^2+\widehat{B}^2) +\frac{2 \eta \widehat{B}^2}{\lambda },
    \end{aligned}
\end{equation}
where $\widehat{B}$ is defined as below,
\begin{equation}\nonumber
    \widehat{B}=\sup _{t \in T} \left\| b_t \right\| \leqslant \frac{\widehat{r}_n(X)^{(k)}}{(k-1)C^{k-1}}.
\end{equation}
% Integrating all the above, we have
% \begin{equation}\nonumber
%     \left\| w_{t+1}-\widehat{w} \right\| ^2 \leqslant (1-\frac{\lambda \eta}{2} )\left\|w_t-\hat{w}\right\|^2+\eta ^2M_F+2 \eta \widehat{B}\left\|w_t-\hat{w}\right\|
% \end{equation}
Thus, iterating the above equation, we get 

\begin{equation}\nonumber
    \begin{aligned}
     \left\|w_T-\hat{w}\right\|^2 &\leqslant (1-\frac{\lambda \eta}{2} )^T\left\|w_0-\hat{w}\right\|^2+(4\eta ^2(\hat{r}_n^2(x)+\lambda ^2D^2+\widehat{B}^2)+\frac{2 \eta \widehat{B}^2}{\lambda })\sum\limits_{t=1}^{T-1} (1-\frac{\lambda \eta }{2})^t\\
     & \leqslant (1-\frac{\lambda \eta}{2})^T\left\|w_0-\hat{w}\right\|^2+(4\eta ^2(\hat{r}_n^2(x)+\lambda ^2D^2+\widehat{B}^2)+\frac{2 \eta \widehat{B}^2}{\lambda })\frac{2}{\lambda  \eta }\\
     & = (1-\frac{\lambda \eta}{2} )^T\left\|w_0-\hat{w}\right\|^2+\frac{8\eta}{\lambda }(\hat{r}_n^2(x)+\lambda ^2D^2+\widehat{B}^2)+\frac{4 \widehat{B}^2}{\lambda ^2}\\
     &\leqslant \exp\{-\frac{\lambda \eta T}{2} \}\left\| w_0-\widehat{w} \right\|^2 + \frac{8\eta \hat{r}_n^2(x)}{\lambda }+8\eta\lambda D^2+\frac{8 \eta \widehat{B}^2}{\lambda }+\frac{4 \widehat{B}^2}{\lambda ^2}\\
     &\leqslant \exp\{-\frac{\lambda \eta T}{2} \}\left\| w_0-\widehat{w} \right\|^2 + \frac{8\eta \hat{r}_n^2(x)}{\lambda }+8\eta\lambda D^2+\frac{20 \widehat{B}^2}{\lambda ^2}.
    \end{aligned}
 \end{equation}
 The last inequality holds due to the assumption that $\eta \leqslant \frac{2}{\lambda }$.
\end{proof}

\begin{theorem}
    We have the following bound for $ f(w_l)-f(\hat{w}_l) $:
    \begin{equation}\nonumber
         F(w_l)-F(\hat{w}_l) \leqslant  \widetilde{O}\left(\frac{\widetilde{r}_{2k,n_l}}{\widetilde{R}_{2k,n}}\cdot \frac{DL_f }{  \sqrt{n}}\right).
    \end{equation}
    % where $M_F=\min \left\{\frac{16LD_l}{n^2},\left(L_f+\widetilde{r}_{2k}(\frac{\sqrt[]{d}}{\varepsilon })^{\frac{k-1}{k}}\right)\right\}$. 
\end{theorem}

\begin{proof}
    Firstly, the choise of $D_i$ ensures that $\hat{w}_i \in \mathcal{W}_i$.

    Then by the above lemma, and choosing specific $T_i$,
    \begin{equation}\nonumber
        \left\| w_i-\hat{w}_i \right\| ^2 \leqslant \exp\{-\frac{\lambda_i \eta_i T_i }{2} \}\left\| w_{i-1}-\widehat{w}_i \right\|^2 + \frac{8\eta_i  \hat{r}_{n_i}^2(B_i)^{(2)}}{\lambda_i  }+8\eta_i \lambda_i  D_i ^2+\frac{20 \hat{r}_{n_i}^2(B_i)^{(2)}}{\lambda _i^2(k-1)C_i^{k-1}}.
    \end{equation}
    \begin{equation}\label{ite_w}
        \left\| w_i-\hat{w}_i \right\| ^2 \lesssim \frac{\eta _i }{\lambda _i }L_f^2+\frac{\widetilde{r}_{n_i}^{(2k)}}{\lambda _i^2C_i^{2k-2}4^i} \lesssim \frac{\eta ^2 n}{16^i 4^i}(L_f^2+\frac{n \widetilde{r}_{n_i}^{(2k)}}{C_i^{2k-2}4^i}).
    \end{equation}
    Then by setting $L=\sup _{w \in \mathcal{W}}\left\| \nabla F(w) \right\|  \leqslant r$. Therefore,
    \begin{equation}\nonumber
        \begin{aligned}
            F(w_l)-F(\hat{w}_l)\leqslant &\sqrt[]{\left\| w_l-\hat{w}_l \right\| ^2}\\
            \leqslant &L \sqrt[]{\eta _l^2(L_f^2+\frac{ \widetilde{e}_{n_i}^{(2k)}}{C_l^{2k-2}4^i})}\\
            \lesssim & L \frac{\eta }{n^2}(L_f+\frac{\widetilde{r}_{2k}^k}{C_l^{k-1}})\\
            \lesssim & L \frac{\eta }{n^2}(L_f+\frac{\widetilde{r}_{2k}^k}{C_l^{k-1}})\\
            \leqslant & L \frac{\eta}{n^2}\left(L_f+\widetilde{r}_{2k}(\frac{\sqrt[]{d}}{\varepsilon })^{\frac{k-1}{k}}\right).
        \end{aligned}
    \end{equation}

    We know that $\xi _i \sim \mathcal{N}(0,\sigma _i^2)$ and $\xi $ is sub-Gaussian, thus, we can derive that
\begin{equation}\nonumber
    \mathbb{P}\{\left\| \xi _i \right\| \geqslant t\ \sqrt[]{d}\}\leqslant 2 \exp \{-\frac{t^2}{16 \sigma _i^2}\}.
\end{equation}

Here there shall be some confusion about the lower index, where $k$ is equivalent to $l$ as above, not the original $k$ here.
Therefore, with probability $1-\beta $, $\left\| \xi _i \right\| \leqslant 4\ \sqrt[]{d}\sigma _i \log_{ }(4/\beta) $. Thus, due to the choice of $\eta$, we have
\begin{equation}\nonumber
    \begin{aligned}
        F(w_l)-F(\hat{w}_l) \leqslant& 4L_f \sqrt[]{d}\sigma _l \log_{} (4/\beta )=4L_f \sqrt[]{d} \log_{} (4/\beta )\frac{8 C_l\ \sqrt[]{\log_{} (1/\delta) }}{n_l\lambda_l\varepsilon } \\
    =&32L_f \sqrt[]{d \log (1/\delta)} \log_{} (4/\beta )\frac{C_l \eta _l n_l^{p-1}}{\varepsilon }\\
    =&32L_f \sqrt[]{d  \log (1/\delta)} \log_{} (4/\beta )\widetilde{r}_{2k,n_l}\left(\frac{\varepsilon n_l}{\sqrt[]{d \log_{} (n)}}\right)^{\frac{1}{k}}\frac{\eta }{4^l}\frac{n^{p-1}}{(2^l)^{p-1}}\frac{1}{\varepsilon }\\
    \leqslant&\frac{\widetilde{r}_{2k,n_l}}{\widetilde{R}_{2k,n}}\cdot \frac{32DL \log_{} (1/\beta )}{\sqrt{n}\log_{}^{p+\frac{5}{2}} n }.
    %L C_l \frac{16}{2^{4l}}\frac{\sqrt[]{d}}{\varepsilon }\sqrt[]{\log_{} (1/\delta) }\log_{} (4/\beta )\eta=LC_l16^{-(l-1)}\frac{\sqrt[]{d}}{\varepsilon }\sqrt[]{\log_{} (1/\delta) }\log_{} (4/\beta )\eta\\
     %\leqslant & \frac{LD_l}{16^{l-1}}=\frac{16LD_l}{n^2}.
    \end{aligned}
\end{equation} 

% To make it simpler, we just need to find the smaller one. 
\end{proof}
Finally, we reach the upper bound for $F(w_l)-F(w^*)$:
\begin{theorem}
    Finally, we reach the upper bound for $F(w_l)-F(w^*)$:
    \begin{equation}\nonumber
        F(w_l)-F(w^*) \lesssim \widetilde{R}_{2k,n}D( \frac{1}{\sqrt[]{n}}+(\frac{\sqrt[]{d \log (1/\delta)\log_{} n}}{\varepsilon  n})^{\frac{k-1}{k}})+\frac{D\sqrt[]{\log_{} (1/\beta )}}{2^{p+1}\sqrt[]{n}}.%+\frac{L_f \sqrt[]{\log_{} n}}{\sqrt[]{n}}+\frac{cL_f D \sqrt[]{\log_{} (1/\beta )}}{\sqrt[]{n}}.
    \end{equation}
\end{theorem}

\begin{proof}
   
    Rewrite this term into summation of their differences,
    \begin{equation}\nonumber
        F(w_l)-F(w^*)=\sum\limits_{i=1}^{l} [f(\hat{w}_i)-f(\hat{w}_{i-1})]+[f(w_l)-f(\hat{w}_l)],
    \end{equation}
    By lemma \ref{7}, 
    \begin{equation}\nonumber
        F(\hat{w}_i)-F(\hat{w}_{i-1}) \leqslant \frac{cL^2 \log_{} n_i \log {(2/\beta )} }{\lambda _i n_i}+\frac{cL D_i \sqrt[]{\log_{} (2/\beta )}}{\sqrt[]{n_i}}+\frac{\lambda _i}{2}\left\| w_{i-1}-\hat{w}_{i-1} \right\| ^2.
    \end{equation}
For $\left\| w_i-\widehat{w} _i\right\| ^2 \leqslant \frac{\eta _i }{\lambda _i }L_f^2+\frac{\widetilde{r}_{n_i}^{(2k)}}{\lambda _i^2C_i^{2k-2}4^i} \leqslant O \left( \frac{\eta ^2 n}{16^i 4^i}(L_f^2+\frac{n \widetilde{r}_{n_i}^{(2k)}}{C_i^{2k-2}4^i})\right)$, then summing over $i$ from $1$ to $l$, we have with probability at least $1-\beta $, for some constant $C_0$

\begin{equation}\nonumber
    \begin{aligned}
        &f(w_l)-f(w^*)\\
         \leqslant & C_0\sum\limits_{i=1}^{l} \left\{\lambda _i \left\| \widehat{w}_{i-1} -w_{i-1}\right\| ^2+ \frac{cL_f^2 \log_{} n_i \log_{} (1/\beta )}{\lambda _i n_i}+\frac{cL_f D_i \sqrt[]{\log_{} (1/\beta )}}{\sqrt[]{n_i}}\right\}\\%第一行
        \leqslant & \lambda_1 \left\| \widehat{w}_0-w_0 \right\| ^2+ 
        \sum\limits_{i=2}^{l} \lambda _i \left\| \widehat{w}_{i-1}-w_{i-1}\right\|^2+\sum\limits_{i=1}^l\frac{L_f^2 \log n_i\log(1/\beta)}{\lambda_i n_i}
        + \sum\limits_{i=1}^l \frac{L_f D_i \sqrt{\log (1/\beta)}}{\sqrt{n_i}}\\%第二行
        \leqslant & \frac{D^2}{\eta n^{2p}}+ \sum\limits_{i=2}^{l} \lambda _i\left[\eta^2_i n_i^p L_f^2+\frac{\eta_i^2n_i^{2p}\widetilde{e}_{n_i}^{(2k)}}{C_i^{2k-2}}\right] 
        + \sum\limits_{i=1}^l\frac{L_f^2(\log n-\log2^i)\log (1/\beta)}{n_i}\eta_i n_i^{p}
        +\sum\limits_{i=1}^lL_f^2 \eta_i n_i^{p-\frac{1}{2}} \sqrt{\log (1/\beta)}\\%第三行，具体表达式
        \leqslant & \frac{D^2}{\eta n^{2p}}+ \sum\limits_{i=2}^{l} \left[\eta^2_i L_f^2 
        + \frac{\eta_i n_i^{p}\widetilde{e}_{n_i}^{(2k)}}{C_i^{2k-2}}\right] 
        + \sum\limits_{i=1}^lL_f^2 \eta_i n_i^{p-1}(\log n-\log2^i)\log (1/\beta)
        +\sum\limits_{i=1}^lL_f^2 \eta_i n_i^{p-\frac{1}{2}} \sqrt[]{\log (1/\beta)}\\%第四行
        \leqslant & \frac{D^2}{\eta n^{2p}}+ \eta \left(L_f^2+\widetilde{R}_{2k,n}n^p(\frac{d \log_{} n}{\varepsilon ^2 n^2})^{\frac{k-1}{k}}\right)
        + L_f^2\eta n^{p-1}\log (1/\beta)\sum\limits_{i=1}^{l} \frac{(\log_{} n -i)}{4^i \cdot (2^{p-1})^i}
        +L_f^2 \eta n^{p-\frac{1}{2}} \sqrt[]{\log (1/\beta)}\sum\limits_{i=1}^l (\frac{1}{2^{p+\frac{3}{2}}})^i\\  
        \leqslant & \frac{D^2}{\eta n^{2p}}+ \eta \left(L_f^2+\widetilde{R}_{2k,n}n^p(\frac{d \log_{} n}{\varepsilon ^2 n^2})^{\frac{k-1}{k}}\right)\\
        +&L_f^2 \eta n^{p-1} \log (1/\beta)\left(\frac{\log_{} n}{2^{p+1}}+\frac{1}{2^{p+1}\log_{} ^pn }\right)
        +L_f^2 \eta n^{p-\frac{1}{2}}\sqrt[]{\log (1/\beta)}\frac{1-\frac{1}{n^{p+\frac{3}{2}}}}{2^{p+\frac{3}{2}}-1}\\
        \leqslant & \frac{D^2}{\eta n^{2p}}+\eta (L_f^2+\widetilde{R}_{2k,n}n^p(\frac{d \log_{} n}{\varepsilon ^2 n^2})^{\frac{k-1}{k}})+ 
        L_f^2 \eta n^{p-1}\log (1/\beta) \log_{} n \cdot 2^{-(p+1)}
        +L_f^2 \eta n^{p-\frac{1}{2}} \sqrt[]{\log (1/\beta)} \cdot 2^{-(p+\frac{3}{2})}.\\
    \end{aligned}
\end{equation}
Assume that $\exists p$ s.t. $L_f \leqslant O\left(n^{p/2}\widetilde{R}_{2k,n}(\frac{1}{\sqrt[]{n}}+(\frac{\sqrt[]{d \log (1/\delta)\log_{} n}}{\varepsilon  n})^{\frac{k-1}{k}})\right)$ and 
take  $\eta =\frac{D}{n^{\frac{p}{2}}}\min \{\frac{1}{L_f},\frac{1}{\widetilde{R}_{2k,n}n^{\frac{p+1}{2}}}(\frac{\varepsilon n}{\sqrt[]{d \log (1/\delta)\log_{} n}})^{\frac{k-1}{k}}$, $ \frac{1}{n^{\frac{p-1}{2}} L_f^2 \sqrt[]{\log_{} n \log_{} (1/\beta )}}\}$, then the above can be reduced to 
\begin{equation}\nonumber
    f(w_l)-f(w^*) \leqslant O \left( \widetilde{R}_{2k,n}D( \frac{1}{\sqrt[]{n}}+(\frac{\sqrt[]{d \log (1/\delta)\log_{} n}}{\varepsilon  n})^{\frac{k-1}{k}})+\frac{D\sqrt[]{\log_{} (1/\beta )}}{2^{p+1}\sqrt[]{n}}\right),%+\frac{L_f \sqrt[]{\log_{} n}}{\sqrt[]{n}}+\frac{cL_f D \sqrt[]{\log_{} (1/\beta )}}{\sqrt[]{n}}.
\end{equation}
which holds with probability at least $1-\beta $.

\end{proof}
\end{proof}
\begin{proof}[{\bf Proof of Thorem \ref{thm:2}}]
    \begin{theorem}
    Assume that loss function $F(\cdot )$ is $(\theta ,\lambda )$-TNC and $f(\cdot ,x )$ is convex, $\alpha  $-smooth and $L_f$-Lipschitz for each $x$. Then algorithm \ref{alg:4} is $(\varepsilon ,\delta )$-DP based on different stepsizes $\{\gamma_k \}_{k=1}^m$ and noises if $\gamma _k \leqslant \frac{1}{\alpha }$. Then for sufficiently large $n $ and $(\varepsilon ,\delta )$-DP, with probability at least $1-\beta $, we have
    \begin{equation}\nonumber
        F\left(\hat{w}_m\right)-F\left(w^*\right) \leqslant O\left(\frac{1}{\lambda^{\frac{1}{\theta-1}}} \cdot\left(\widetilde{R}_{2k,n}( \frac{\sqrt[]{\log_{} n}}{\sqrt[]{n}}+\left(\frac{\sqrt[]{d \log (1/\delta) \log_{}^3 n}}{\varepsilon  n}\right)^{\frac{k-1}{k}})+\frac{\sqrt[]{\log_{} n \log_{} (1/\beta )}}{2^{p+1}\sqrt[]{n}}\right)^{\frac{\theta}{\theta-1}}\right).
    \end{equation}
\end{theorem}

\begin{proof}
    
   % For convenience here we only show the proof of $(\varepsilon, \delta)$-DP. %The proof of $\varepsilon$-DP is almost the same by replacing the term $\left(\frac{1}{\sqrt{n}}+\frac{\sqrt{d \log (1 / \delta)}}{\varepsilon n}\right)$ to $\left(\frac{1}{\sqrt{n}}+\frac{d}{n \varepsilon}\right)$ in the following proof.

  The guarantee of $(\varepsilon, \delta)$-DP is just followed by Theorem \ref{thm:1}. 
  
  For simplicity, we denote $a(n)= O\left( \widetilde{R}_{2k,n}( \frac{1}{\sqrt[]{n}}+(\frac{\sqrt[]{d \log_{} n}}{\varepsilon  n})^{\frac{k-1}{k}})+\frac{\sqrt[]{\log_{} (1/\beta )}}{2^{p+1}\sqrt[]{n}}\right)$. We set $\mu_0=2 R_0^{1-\theta} a\left(n_0\right), \mu_k=$ $2^{(\theta-1) k} \mu_0$ and $R_k=\frac{R_0}{2^k}$, where $k=1, \cdots, m$.
  
  Then we have $\mu_k \cdot R_k^\theta=2^{-k} \mu_0 R_0^\theta$. We can also assume that $\lambda \leqslant \frac{L}{R_0^{\theta-1}}$, otherwise we can set $\lambda=\frac{L}{R_0^{\theta-1}}$, which makes TNC still hold.
  Recall that $m=\left\lfloor\frac{1}{2} \log _2 \frac{2 n}{\log _2 n}\right\rfloor-1$, when $n \geq 256$, it follows that
  $$
  0<\frac{1}{2} \log _2 \frac{2 n}{\log _2 n}-2 \leqslant m \leqslant \frac{1}{2} \log _2 \frac{2 n}{\log _2 n}-1 \leqslant \frac{1}{2} \log _2 n .
  $$
  Thus, we have $2^m \geq \frac{1}{4} \sqrt{\frac{2 n}{\log _2 n}}$.(if we pick specific $m$ such that $2^m \geq \frac{1}{4} \sqrt{\frac{2 n}{\log _2 n}} \cdot \frac{1}{\log_{}  n_0\ \sqrt[]{\log_{} (1/\beta )} } $ )
  Thus
  $$
  \begin{aligned}
  \mu_m=2^{(\theta-1) m} \mu_0 & \geq 2^m \mu_0 \\
  & \geq \frac{1}{4} \sqrt{\frac{2 n}{\log _2 n}}\frac{1}{\log_{}  n_0\ \sqrt[]{\log_{} (1/\beta )} } \cdot 2 \cdot R_0^{1-\theta} a\left(n_0\right) \\
  & = \frac{5 \cdot  R_0^{1-\theta}}{\log_{}  n_0\ \sqrt[]{\log_{} (1/\beta )} }\sqrt{\frac{2 n}{\log _2 n}}\left(  \widetilde{R}_{2k,n_0}( \frac{1}{\sqrt[]{n_0}}+(\frac{\sqrt[]{d \log_{} n_0}}{\varepsilon  n_0})^{\frac{k-1}{k}})+\frac{\sqrt[]{\log_{} (1/\beta )}}{2^{p+1}\sqrt[]{n_0}}\right)\\%\left(\frac{1}{\sqrt{\frac{n}{m}}}+\frac{\sqrt{d \log (n/m)}}{\varepsilon \cdot \frac{n}{m}}\right) \\
  & \geq 5 \cdot \widetilde{R}_{2k,n_0} R_0^{1-\theta} \sqrt{\frac{2 n}{\log _2 n}}\left(\frac{1}{\sqrt{\frac{2 n}{\log _2 2 n-\log _2 \log _2 n-4}}}\right)  \\
  & =5 \cdot \widetilde{R}_{2k,n_0} R_0^{1-\theta} \sqrt{\frac{\log _2 2 n-\log _2 \log _2 n-4}{\log _2 n}}  \cdot \log_{}  n_0\ \sqrt[]{\log_{} (1/\beta )} \\
  & \geq \widetilde{R}_{2k,n_0} R_0^{1-\theta}\left(\text { Since } 5 \cdot \sqrt{\frac{\log _2 2 n-\log _2 \log _2 n-4}{\log _2 n}} \geq 1 \text { when } n \geq 256\right) \\
  & \geq \lambda(\text { By assumption }) .
  \end{aligned}
  $$
  where the third inequality is given by throwing away the $(\frac{\sqrt[]{d \log_{} n_0}}{\varepsilon  n_0})^{\frac{k-1}{k}}$ and $\frac{\sqrt[]{\log_{} (1/\beta )}}{2^{p+1}\sqrt[]{n_0}}$  term and substituting $m$ in term $\frac{1}{\sqrt{\frac{n}{m}}}$ with $\frac{1}{2} \log _2 \frac{2 n}{\log _2 n}-2$.
  Below, we consider the following two cases.\\
  \textbf{Case 1} If $\lambda \geq \mu_0$, then $\mu_0 \leqslant \lambda \leqslant \mu_m$. We have the following lemma.

  % lemma 7
  \begin{lemma}
   Let $k^*$ satisfies $\mu_{k^*} \leqslant \lambda \leqslant 2^{\theta-1} \mu_{k^*}$, then for any $1 \leqslant k \leqslant k^*$, the points $\left\{\hat{w}_k\right\}_{k=1}^m$ generated by Algorithm \ref{alg:4} satisfy
  \begin{equation}\label{lemma_eq1}
   \left\|\hat{w}_{k-1}-w^*\right\|_2 \leqslant R_{k-1}=2^{-(k-1)} \cdot R_0,
  \end{equation}
  \begin{equation}\label{lemma_eq2}
   F\left(\hat{w}_k\right)-F\left(w^*\right) \leqslant \mu_k R_k^\theta=2^{-k} \mu_0 R_0^\theta .
  \end{equation}
  Moreover, for $k \geq k^*$, we have
 \begin{equation}\label{lemma_eq3}
   F\left(\hat{w}_k\right)-F\left(\hat{w}_{k^*}\right) \leqslant \mu_{k^*} R_{k^*}^\theta .
 \end{equation}
  \end{lemma}
\begin{proof}
We prove (\ref{lemma_eq1}), (\ref{lemma_eq2}) by induction. Note that (\ref{lemma_eq1}) holds for $k=1$. Assume (\ref{lemma_eq1}) is true for some $k>1$, then we have
  $$
  \begin{aligned}
  F\left(\hat{w}_k\right)-F\left(w^*\right) & \leqslant  R_{k-1} \cdot \left(\widetilde{R}_{2k,n_0}( \frac{1}{\sqrt[]{n_0}}+(\frac{\sqrt[]{d \log_{} n_0}}{\varepsilon  n_0})^{\frac{k-1}{k}})+\frac{\sqrt[]{\log_{} (1/\beta )}}{2^{p+1}\sqrt[]{n_0}}\right) \\
  & =R_{k-1} a\left(n_0\right) \\
  & =\frac{1}{2} \mu_k 2^{(1-\theta) k} R_0^{\theta-1} R_{k-1} \\
  & =\mu_k R_k^\theta
  \end{aligned}
  $$
  Which is (\ref{lemma_eq2}). By the definition of TNC, we have
  $$
  \begin{aligned}
  \left\|\hat{w}_k-w^*\right\|_2^\theta & \leqslant \frac{1}{\lambda}\left(F\left(\hat{w}_k\right)-F\left(w^*\right)\right) \\
  & \leqslant \frac{F\left(\hat{w}_k\right)-F\left(w^*\right)}{\mu_{k^*}} \\
  & \leqslant \frac{\mu_k R_k^\theta}{\mu_{k^*}} \leqslant R_k^\theta
  \end{aligned}
  $$
  Thus (\ref{lemma_eq1}) is true for $k+1$.
  Now we prove (\ref{lemma_eq3}). Referring to Theorem \ref{thm:1} , we know that
  $$
  \begin{aligned}
F\left(\hat{w}_k\right)-F\left(\hat{w}_{k-1}\right) & \leqslant R_{k-1} \cdot a\left(n_0\right) \\
  & =2^{k^*-k} R_{k^*-1} a\left(n_0\right) \\
  & =2^{k^*-k} \mu_{k^*} R_{k^*}^\theta \\
  & =\mu_k R_k^\theta
  \end{aligned}
  $$
  Thus, for $k>k^*$,
  $$
  \begin{aligned}
  F\left(\hat{w}_k\right)-F\left(\hat{w}_{k^*}\right) & =\sum_{j=k^*+1}^k\left(F\left(\hat{w}_j\right)-F\left(\hat{w}_{j-1}\right)\right) \\
  & \leqslant \sum_{j=k^*+1}^k 2^{k^*-j} \mu_{k^*} R_{k^*}^\theta \\
  & =\left(1-2^{k^*-k}\right) \mu_{k^*} R_{k^*}^\theta \\
  & \leqslant \mu_{k^*} R_{k^*}^\theta
  \end{aligned}
  $$
\end{proof}
    
Here completes the proof of the lemma. Now we proceed to prove Theorem \ref{thm:1} in this case.
  
  $$
  \begin{aligned}
  F\left(\hat{w}_m\right)-F\left(w^*\right) & =\left(F\left(\hat{w}_m\right)-F\left(\hat{w}_{k^*}\right)\right)+\left(F\left(\hat{w}_{k^*}\right)-F\left(w^*\right)\right) \\
  & \leqslant 2 \mu_{k^*} R_{k^*}^\theta \\
  & \leqslant 4\left(\frac{\mu_{k^*}}{\lambda}\right)^{\frac{1}{\theta-1}} \mu_{k^*} R_{k^*}^\theta\left(\operatorname{Since}\left(\frac{\mu_{k^*}}{\lambda}\right)^{\frac{1}{\theta-1}} \geq \frac{1}{2}\right) \\
  & =4\left(\frac{2^{(\theta-1) k^*} \mu_0}{\lambda}\right)^{\frac{1}{\theta-1}} \mu_{k^*} R_{k^*}^\theta \\
  & =4\left(2^{k^*} \mu_{k^*} R_{k^*}^\theta \mu_0^{\frac{1}{\theta-1}}\left(\frac{1}{\lambda}\right)^{\frac{1}{\theta-1}}\right) \\
  & =4\left(\mu_0 R_0^\theta \mu_0^{\frac{1}{\theta-1}}\left(\frac{1}{\lambda}\right)^{\frac{1}{\theta-1}}\right) \\
  & =4\left(R_0^\theta \mu_0^{\frac{\theta}{\theta-1}}\left(\frac{1}{\lambda}\right)^{\frac{1}{\theta-1}}\right) \\
  & =4 \cdot\left(\left(2 \cdot a\left(n_0\right)\right)^{\frac{\theta}{\theta-1}}\left(\frac{1}{\lambda}\right)^{\frac{1}{\theta-1}}\right) \\
  & =4 \cdot\left(\frac{1}{\lambda}\right)^{\frac{1}{\theta-1}} \cdot 2\left( \widetilde{R}_{2k,n_0}( \frac{1}{\sqrt[]{n_0}}+(\frac{\sqrt[]{d \log_{} n_0}}{\varepsilon  n_0})^{\frac{k-1}{k}})+\frac{\sqrt[]{\log_{} (1/\beta )}}{2^{p+1}\sqrt[]{n_0}} \right)^{\frac{\theta}{\theta-1}}
  \end{aligned}
  $$
  where $m=O\left(\log _2 n\right) (  $ Recall that $ m \leqslant \frac{1}{2} \log _2 n)$.\\
  \textbf{Case 2} If $\lambda<\mu_0$, then
  $$
  \begin{aligned}
  F\left(\hat{w}_1\right)-F\left(w^*\right) & \leqslant R_0 a\left(n_0\right) \\
  & =\left(\frac{2}{\mu_0}\right)^{\frac{1}{\theta-1}} \cdot a\left(n_0\right)^{\frac{\theta}{\theta-1}} \\
  & <\left(\frac{2}{\lambda}\right)^{\frac{1}{\theta-1}} \cdot a\left(n_0\right)^{\frac{\theta}{\theta-1}}
  \end{aligned}
  $$
  Also, we have
  $$
  \begin{aligned}
  F\left(\hat{w}_m\right)-F\left(\hat{w}_1\right) & =\sum_{j=2}^m\left(F\left(\hat{w}_j\right)-F\left(\hat{w}_{j-1}\right)\right) \\
  & \leqslant \sum_{j=2}^m R_{j-1} \cdot a\left(n_0\right) \\
  & =\sum_{j=2}^m 2^{-(j-1)} R_0 \cdot a\left(n_0\right) \\
  & =\left(1-(1 / 2)^{m-1}\right) R_0 \cdot a\left(n_0\right)<R_0 \cdot a\left(n_0\right)
  \end{aligned}
  $$
  
  By a similar argument process as in Case 1, we have
  $$
  \begin{aligned}
  F\left(\hat{w}_m\right)-F\left(w^*\right) & =\left(F\left(\hat{w}_m\right)-F\left(\hat{w}_1\right)\right)+\left(F\left(\hat{w}_1\right)-F\left(w^*\right)\right) \\
  & \leqslant 2 R_0 a\left(n_0\right) \leqslant 2\left(\frac{2}{\lambda}\right)^{\frac{1}{\theta-1}} \cdot a\left(n_0\right)^{\frac{\theta}{\theta-1}} \\
  & =2 \cdot\left(\frac{2}{\lambda}\right)^{\frac{1}{\theta-1}} \cdot \left(\widetilde{R}_{2k,n_0}( \frac{1}{\sqrt[]{n_0}}+(\frac{\sqrt[]{d \log_{} n_0}}{\varepsilon  n_0})^{\frac{k-1}{k}})+\frac{\sqrt[]{\log_{} (1/\beta )}}{2^{p+1}\sqrt[]{n_0}}\right)^{\frac{\theta }{\theta -1}}
  \end{aligned}
  $$
  Combining the two cases, we conclude that with probability at least $1-\beta $,
  $$
  F\left(\hat{w}_m\right)-F\left(w^*\right) \leqslant O\left(\frac{1}{\lambda^{\frac{1}{\theta-1}}} \cdot\left(\widetilde{R}_{2k,n}( \frac{\sqrt[]{\log_{} n}}{\sqrt[]{n}}+\left(\frac{\sqrt[]{d  \log (1/\delta) \log_{}^3 n}}{\varepsilon  n}\right)^{\frac{k-1}{k}})+\frac{\sqrt[]{\log_{} n \log_{} (1/\beta )}}{2^{p+1}\sqrt[]{n}}\right)^{\frac{\theta}{\theta-1}}\right).
  $$
  \end{proof}
\end{proof}

\begin{proof}[{\bf Proof of Theorem \ref{thm:3}}]
    \begin{proof}

The guarantee of $(\varepsilon, \delta)$-DP is just followed by Theorem \ref{thm:1} and the parallel theorem of Differential Privacy. In the following we  focus on the utility.

Since $k=\left\lfloor\left(\log \log _{\bar{\theta}} 2\right) \cdot \log \log n\right\rfloor$, then $k \leqslant\left(\log _{\bar{\theta}} 2\right) \cdot \log \log n$, namely $2^k \leqslant(\log n)^{\log 2}$ and $\frac{2^k-1}{(\log n)^{\log _{\theta^2}}} \leqslant 1$. Observe that the total sample number used in the algorithm is $\sum_{i=1}^k n_i \leqslant\sum_{i=1}^k \frac{2^{i-1} n}{(\log n)^{\log_{\bar{\theta }} 2}}=\frac{\left(2^k-1\right) n}{(\log n)^{\log_{\bar{\theta }} 2}} \leqslant n$.

For the output of phase $i$, denote $\Delta_i=F\left(w_i\right)-F\left(w^*\right)$, and let $D_i^\theta=\left\|w_i-w^*\right\|_2^\theta$. The assumption of TNC implies that $F\left(w_i\right)-F\left(w^*\right) \geq \lambda\left\|w_i-w^*\right\|_2^\theta$, which is $F\left(w_i\right)-F\left(w^*\right) \geq$ $\lambda \left\|w_i-w^*\right\|_2^\theta$ when we take expectations at both sides, namely
\begin{equation}\label{simplified}
    \Delta_i \geq \lambda D_i^\theta .
\end{equation}

Thus, we have
\begin{equation}\label{eq6}
    \begin{aligned}
        \Delta_i &\leqslant c\widetilde{R}_{2k,n}D_{i-1}( \frac{1}{\sqrt[]{n_i}}+(\frac{\sqrt[]{d \log_{} n_i}}{\varepsilon  n_i})^{\frac{k-1}{k}})+\frac{cD_{i-1}\sqrt[]{\log_{} (1/\beta )}}{2^{p+1}\sqrt[]{n_i}}\\
         &\stackrel{(\ref{simplified})}{\leqslant} \left(\frac{\Delta_{i-1}}{\lambda}\right)^{\frac{1}{\theta }}\left(c\widetilde{R}_{2k,n}( \frac{1}{\sqrt[]{n_i}}+(\frac{\sqrt[]{d \log_{} n_i}}{\varepsilon  n_i})^{\frac{k-1}{k}})+\frac{c\sqrt[]{\log_{} (1/\beta )}}{2^{p+1}\sqrt[]{n_i}} \right),
    \end{aligned}
\end{equation}
where the first inequality comes from Theorem \ref{thm:1} and the second inequality uses (\ref{simplified}). Denote $E_i=\frac{c^\theta  }{\lambda}\left(\widetilde{R}_{2k,n}( \frac{1}{\sqrt[]{n_i}}+(\frac{\sqrt[]{d \log_{} n_i}}{\varepsilon  n_i})^{\frac{k-1}{k}})+\frac{\sqrt[]{\log_{} (1/\beta )}}{2^{p+1}\sqrt[]{n_i}} \right)^\theta$. Then (\ref{eq6}) can be simplified as
\begin{equation}\label{eq7}
    \Delta_i \leqslant\left(\Delta_{i-1} E_i\right)^{\frac{1}{\theta}} .
\end{equation}

Notice that $n_i / n_{i-1}=2$, then $\frac{E_{i-1}}{E_i} \leqslant\left(\frac{n_i}{n_{i-1}}\right)^\theta=2^\theta$, namely:
\begin{equation}\label{eq8}
    E_i \geq 2^{-\theta} E_{i-1}.
\end{equation}

Then we can rearrange the above inequality as
\begin{equation}\label{eq9}
    \frac{\Delta_i}{E_i^{\frac{1}{\theta -1}}} \leqslant \frac{\left(\Delta_{i-1} E_i\right)^{\frac{1}{\theta}}}{E_i^{\frac{1}{\theta -1}}} \leqslant 2^{\frac{1}{\theta -1}}\left(\frac{\Delta_{i-1}}{E_{i-1}^{\frac{1}{\theta -1}}}\right)^{\frac{1}{\theta }},
\end{equation}
where the first inequality uses (\ref{eq7}) and the second inequality applies (\ref{eq8}).

It can be verified that (\ref{eq9}) is equivalent to
\begin{equation}\nonumber
    \frac{\Delta_i}{2^{\frac{\theta}{(\theta-1)^2}} E_i^{\frac{1}{\theta-1}}} \leqslant\left(\frac{\Delta_{i-1}}{2^{\frac{\theta}{(\theta-1)^2}} E_{i-1}^{\frac{1}{\theta-1}}}\right)^{\frac{1}{\theta}} \leqslant\left(\frac{\Delta_1}{2^{\frac{\theta}{(\theta-1)^2}} E_1^{\frac{1}{\theta-1}}}\right)^{\frac{1}{\theta^{i-1}}} .
\end{equation}
According to Lemma \ref{lemma:1}, $\Delta_1 \leqslant\left(L^\theta \lambda^{-1}\right)^{\frac{1}{\theta -1}}$. Also observe that
\begin{equation}\nonumber
    E_1=\frac{c^\theta  }{\lambda}\left(\widetilde{R}_{2k,n}( \frac{1}{\sqrt[]{n_1}}+(\frac{\sqrt[]{d \log_{} n_1}}{\varepsilon  n_1})^{\frac{k-1}{k}})+\frac{\sqrt[]{\log_{} (1/\beta )}}{2^{p+1}\sqrt[]{n_1}} \right)^\theta \geq \frac{ c^\theta\widetilde{R}_{2k,n}^\theta}{\lambda} \frac{1}{\left(\sqrt{n_1}\right)^\theta} \geq  \frac{c^\theta\widetilde{R}_{2k,n}^\theta}{\lambda} \frac{1}{n^\theta} .
\end{equation}
Let $c_1=c^{\frac{\theta}{\theta-1}} 2^{\frac{\theta}{(\theta-1)^2}}$, then $\frac{\Delta_1}{2^{\frac{\theta}{(\theta-1)^2}} E_1^{\frac{1}{\theta-1}}} \leqslant \frac{n^{\frac{\theta}{\theta-1}}}{c_1}$, which implies that for $l=\left\lfloor\left(\log _{\bar{\theta}} 2\right) \cdot \log \log n\right\rfloor$,
$$
\frac{\Delta_l}{2^{\frac{\theta}{(\theta-1)^2}} E_l^{\frac{1}{\theta-1}}} \leqslant\left(\frac{n^{\frac{\theta}{\theta-1}}}{c_1}\right)^{\frac{1}{\theta^{l-1}}} .
$$

Let $C_1=2^{\frac{\theta^3}{\theta-1}+\theta^2\left|\log c_1\right|}$. In the following we  prove that
$$
\left(\frac{n^{\frac{\theta}{\theta-1}}}{c_1}\right)^{\frac{1}{\theta^{l-1}}} \leqslant C_1 .
$$

Since $l+1 \geq\left(\log _{\bar{\theta}} 2\right) \log \log n \geq\left(\log _\theta 2\right) \log \log n$, it follows that
$$
(l-1) \log \theta+\log \log C_1 \geq \log \left(\frac{\theta}{\theta-1}+\left|\log c_1\right|\right)+\log \log n,
$$
which indicates
$$
\left(\frac{\theta}{\theta-1}+\left|\log c_1\right|\right) \log n \leqslant \theta^{l-1} \log C_1 .
$$

Thus we have $\frac{\theta}{\theta-1} \log n-\log c_1 \leqslant \theta^{l-1} \log C_1$, which is equivalent to our object $\left(\frac{n^{\frac{\theta}{\theta-1}}}{c_1}\right)^{\frac{1}{\theta^{k-1}}} \leqslant$ $C_1$.
Now we know
$$
\frac{\Delta_l}{2^{\frac{\theta^2}{(\theta-1)^2}} E_l^{\frac{1}{\theta-1}}} \leqslant\left(\frac{n^{\frac{\theta}{\theta-1}}}{c_1}\right)^{\frac{1}{\theta^{l-1}}} \leqslant C_1,
$$
which indicates that $\frac{\Delta_l}{E_l^{\frac{\theta}{\theta-1}}} \leqslant 2^{\frac{\theta}{(\theta-1)^2}} C_1=2^{\theta^2\left(\frac{\theta^2-\theta+1}{(\theta-1)^2}+\left|\log c_1\right|\right)}:=C$.
As a result, we hold a solution with error:
$$
\begin{aligned}
F\left(w_l\right)-F\left(w^*\right) \leqslant C E_l^{\frac{1}{\theta-1}} & =C\left(\frac{c^\theta }{\lambda}\right)^{\frac{1}{\theta-1}}\left(\widetilde{R}_{2k,n}( \frac{1}{\sqrt[]{n_l}}+(\frac{\sqrt[]{d \log_{} n_l}}{\varepsilon  n_l})^{\frac{k-1}{k}})+\frac{\sqrt[]{\log_{} (1/\beta )}}{2^{p+1}\sqrt[]{n_l}} \right)^{\frac{\theta}{\theta-1}} \\
%& \leqslant 2^{\frac{3 \theta}{2(\theta-1)}} \cdot C\left(\frac{c^\theta }{\lambda}\right)^{\frac{1}{\theta-1}}\left(\frac{1}{n}+\frac{d \log (1 / \delta)}{\varepsilon^2 n^2}\right)^{\frac{\theta}{2(\theta-1)}}
\end{aligned}
$$
\end{proof}
\end{proof}

\begin{proof}[{\bf Proof of Theorem \ref{thm:low1}}]
We first define the set of distributions $\{Q_v\}_{v\in \mathcal{V}}$. Specifically, by the standard Gilbert-Varshamov bound, there exists a set $\mathcal{V}\subset \{\pm\}^d$ such that: (1) $|\mathcal{V}|\geq 2^\frac{d}{20}$, (2) for all $v, v'\in \mathcal{V}$, $d_{ham}(v, v')\geq \frac{d}{8}$ \citep{acharya2021differentially}. For each $v\in \mathcal{V}$, we define $Q_v$ as
\begin{equation}
    X_v=\begin{cases}
        0, \text{ with probability } 1-p\\
        p^{-\frac{1}{k}}\frac{\tilde{r}_k}{2\sqrt{d}}v, \text{ with probability } p 
    \end{cases}
\end{equation}
We can see that for each $X_v\sim Q_v$, we always have $\|\mu_v=\mathbb{E}[X_v]\|_2=p^\frac{k-1}{k}\frac{\tilde{r}_k}{2}=\mu$. 

We then consider the loss function $f(w,x)=-\langle w, x\rangle+\frac{1}{\theta}\|w\|_2^\theta$, i.e., $F_P(w)=-\langle w, \mathbb{E}_P[x] \rangle+\frac{1}{\theta}\|w\|_2^\theta$ for distribution $P$. By \citep{ramdas2012optimal} we know it satisfies $(\theta, 1)$-TNC when $\theta\geq 2$. Moreover, for each $Q_v$ we have 
\begin{equation}
    \mathbb{E}[\sup_{w\in \mathcal{W}}\|\nabla f(w, x)\|_2^k ]=  \mathbb{E}[\sup_{w\in \mathcal{W}}\|\|w\|_2^{\theta-2}w-x \|_2^k ]\leqslant \mathbb{E}[\|2x\|_2^k]=\tilde{r}^k_k=\tilde{r}^{(k)},
\end{equation}
where the first inequality is due to the radius of $\mathcal{W}$ is $(\frac{p^{-\frac{1}{k}}\tilde{r}_k}{2})^\frac{1}{\theta-1}$. Thus we can see $F_P(w)$ satisfies Assumption \ref{1}. For convenience we denote $F_{Q_v}(w)=F_v(w)$. 

By the form the $F_{v}(w)$ we can also see that
\begin{equation}
    \nabla F_v(w^*)=0 \equiv \|w^*\|_2^{\theta-2} w^*=\mu_v.
\end{equation}
Thus the optimal solution $w^*_v=\frac{\mu_v}{\mu^\frac{\theta-2}{\theta-1}}\in \mathcal{W}$ by our assumption on $n$ and thus $p\leqslant 1$. In total we have 
\begin{align}
         &\mathcal{M}(\mathcal{W}, \mathcal{P}, \mathcal{F}_k^\theta( \mathcal{P}, \tilde{r}_k), \rho)\geq    \inf_{\mathcal{A}\in \mathcal{Q}(\rho) }\frac{1}{|\mathcal{V}|} \sum_{v\in \mathcal{V}}
        \mathbb{E}_{\mathcal{A}, D\in Q_v^n} [F_v(\mathcal{A}(D))-\min_{w\in \mathcal{W}} F_v(w)],\\
        & \geq  \inf_{\mathcal{A}\in \mathcal{Q}(\rho) } \frac{1}{|\mathcal{V}|} \sum_{v\in \mathcal{V}}  \mathbb{E}_{\mathcal{A}, D\in Q_v^n} \|\mathcal{A}(D)-w^*_v\|_2^\theta=\inf_{\mathcal{A}\in \mathcal{Q}(\rho) } \frac{1}{|\mathcal{V}|} \sum_{v\in \mathcal{V}}  \mathbb{E}_{\mathcal{A}, D\in Q_v^n} \|\mathcal{A}(D)-\frac{\mu_v}{\mu^\frac{\theta-2}{\theta-1}}\|_2^\theta. \label{eq:apped9}
\end{align}
Next, we recall the following private Fano's lemma: 
\begin{lemma}\label{fano2}[Theorem 1.4 in \citep{kamath2021improved}]
Let  $\mathcal{P}$ be  a class of distributions  over a data universe $\mathcal{X}$. For each distribution $p\in \mathcal{T}$, there is a deterministic function $\theta(p)\in \mathcal{T}$, where $\mathcal{T}$ is the parameter space. Let $\rho:\mathcal{T} \times \mathcal{T} :\mapsto \mathbb{R}_+ $ be  a semi-metric function on the space $\mathcal{T}$ and $\Phi: \mathbb{R}_+\mapsto \mathbb{R}_+$ be a non-decreasing function with $\Phi(0)=0$. We further assume that  $X=\{X_i\}_{i=1}^{n}$ are  $n$ i.i.d observations drawn according to some distribution $p\in \mathcal{P}$, and   $Q:\mathcal{X}^n\mapsto \Theta$ be some algorithm whose output $Q(X)$ is an estimator.  Consider a set of distributions $\mathcal{V}=\{p_1, p_2, \cdots, p_M\}\subseteq \mathcal{P}$ such that for all $i\neq j$, 
\begin{itemize}
    \item $\Phi(\rho(\theta(p_i), \theta_(p_j))\geq \alpha$, 
    \item $D_{KL}(p_i, p_j)\leqslant \beta$, where $D_{KL}$ is the KL-divergence,
    \item $D_{TV}(p_i, p_j)\leqslant \gamma$, 
\end{itemize}
then we have for any $\rho$-zCDP  mechanism $Q$.
\begin{align*}
  \frac{1}{M}\sum_{i\in [M]}\mathbb{E}_{X\sim p_i^n, Q}[\Phi(\rho(Q(X), \theta(p_i))]  \geq \frac{\alpha}{2}\max\{1-\frac{n\beta+\log 2}{\log M}, 1-\frac{\rho(n^2\gamma^2+n\gamma(1-\gamma))+\log 2 }{\log M}\}. 
\end{align*}
\end{lemma}
Now we will leverage the above lemma to lower bound \eqref{eq:apped9}. We can see in our set of probabilities $\{Q_v\}_{v\in \mathcal{V}}$, for any $v, v'\in \mathcal{V}$ we have $D_{TV}(Q_v, Q_{v'})\leqslant p$. And 
\begin{equation}
    \| \frac{\mu_v}{\mu^\frac{\theta-2}{\theta-1}}-\frac{\mu_{v'}}{\mu^\frac{\theta-2}{\theta-1}}\|_2^\theta = \frac{1}{\mu^\frac{\theta(\theta-2)}{\theta-1}}\|p^{\frac{k-1}{k}} \frac{\tilde{r}_k}{2\sqrt{d}}(v-v')\|_2^\theta \geq C\frac{p^{\frac{\theta(k-1)}{k}} }{\mu^\frac{\theta(\theta-2)}{\theta-1}}\tilde{r}_k^\theta =\Omega( \tilde{r}^\frac{\theta}{\theta-1}_k p^{\frac{k-1}{k}\frac{\theta}{\theta-1}}). 
\end{equation}
Taking $p=\frac{\sqrt{d}}{n\sqrt{\rho}}$ and by Lemma \ref{fano2} we have 
\begin{equation}
    \inf_{\mathcal{A}\in \mathcal{Q}(\rho) } \frac{1}{|\mathcal{V}|} \sum_{v\in \mathcal{V}}  \mathbb{E}_{\mathcal{A}, D\in Q_v^n} \|\mathcal{A}(D)-\frac{\mu_v}{\mu^\frac{\theta-2}{\theta-1}}\|_2^\theta\geq \Omega \left( (\tilde{r}_k (\frac{\sqrt{d}}{n\sqrt{\rho}})^\frac{k-1}{k})^\frac{\theta}{\theta-1}\right). 
\end{equation}
\end{proof}

\begin{proof}[{\bf Proof of Theorem \ref{thm:low2}}] The lower bound for non-private case follows the proof in \citep{asi2021private}. Here we extend to the heavy-tailed case. For the index set $\mathcal{V}$ we consider the same one as in the proof of Theorem \ref{thm:low1}. For each $v\in \mathcal{V}$ we define $X\sim P_v$ as 
\begin{equation}
   \text{ for } j\in [d], X_j=\begin{cases}
        v_j e_j \frac{\tilde{r}_k}{2\sqrt{d}}, \text{ with probability } \frac{1+\delta}{2},\\
         -v_j e_j \frac{\tilde{r}_k}{2\sqrt{d}}, \text{ with probability } \frac{1-\delta}{2}. 
    \end{cases}
\end{equation}
We can see that for each $X_v\sim Q_v$, we always have $\|\mu_v=\mathbb{E}[X_v]\|_2=\delta \frac{\tilde{r}_k}{2}=\mu$.  

We then consider the loss function $f(w,x)=-\langle w, x\rangle+\frac{1}{\theta}\|w\|_2^\theta$, i.e., $F_P(w)=-\langle w, \mathbb{E}_P[x] \rangle+\frac{1}{\theta}\|w\|_2^\theta$ for distribution $P$. By \citep{ramdas2012optimal} we know it satisfies $(\theta, 1)$-TNC when $\theta\geq 2$. Moreover, for each $Q_v$ we have 
\begin{equation}
    \mathbb{E}[\sup_{w\in \mathcal{W}}\|\nabla f(w, x)\|_2^k ]=  \mathbb{E}[\sup_{w\in \mathcal{W}}\|\|w\|_2^{\theta-2}w-x \|_2^k ]\leqslant \mathbb{E}[\|2x\|_2^k]=\tilde{r}^k_k=\tilde{r}^{(k)},
\end{equation}
where the first inequality is due to the radius of $\mathcal{W}$ is $(\frac{\tilde{r}_k}{2})^\frac{1}{\theta-1}$. Thus we can see $F_P(w)$ satisfies Assumption \ref{1}. For convenience we denote $F_{Q_v}(w)=F_v(w)$. 

By the form the $F_{v}(w)$ we can also see that
\begin{equation}
    \nabla F_v(w^*)=0 \equiv \|w^*\|_2^{\theta-2} w^*=\mu_v.
\end{equation}
Thus the optimal solution $w^*_v=\frac{\mu_v}{\mu^\frac{\theta-2}{\theta-1}}\in \mathcal{W}$ by our assumption on $n$ and thus $p\leqslant 1$. In total we have 
\begin{align}
         &\mathcal{M}(\mathcal{W}, \mathcal{P}, \mathcal{F}_k^\theta( \mathcal{P}, \tilde{r}_k), \rho)\geq    \inf_{\mathcal{A}\in \mathcal{Q}(\rho) }\frac{1}{|\mathcal{V}|} \sum_{v\in \mathcal{V}}
        \mathbb{E}_{\mathcal{A}, D\in Q_v^n} [F_v(\mathcal{A}(D))-\min_{w\in \mathcal{W}} F_v(w)],\\
        & \geq  \inf_{\mathcal{A}\in \mathcal{Q}(\rho) } \frac{1}{|\mathcal{V}|} \sum_{v\in \mathcal{V}}  \mathbb{E}_{\mathcal{A}, D\in Q_v^n} \|\mathcal{A}(D)-w^*_v\|_2^\theta=\inf_{\mathcal{A}\in \mathcal{Q}(\rho) } \frac{1}{|\mathcal{V}|} \sum_{v\in \mathcal{V}}  \mathbb{E}_{\mathcal{A}, D\in Q_v^n} \|\mathcal{A}(D)-\frac{\mu_v}{\mu^\frac{\theta-2}{\theta-1}}\|_2^\theta. 
\end{align}
We can see in our set of probabilities $\{Q_v\}_{v\in \mathcal{V}}$, for any $v, v'\in \mathcal{V}$ we have $D_{KL}(Q_v, Q_{v'})\leqslant \delta^2$. And  
\begin{equation}
    \| \frac{\mu_v}{\mu^\frac{\theta-2}{\theta-1}}-\frac{\mu_{v'}}{\mu^\frac{\theta-2}{\theta-1}}\|_2^\theta = \frac{1}{\mu^\frac{\theta(\theta-2)}{\theta-1}}\| \frac{\delta\tilde{r}_k}{2\sqrt{d}}(v-v')\|_2^\theta \geq C\frac{\delta^\theta}{\mu^\frac{\theta(\theta-2)}{\theta-1}}\tilde{r}_k^\theta =\Omega( \tilde{r}^\frac{\theta}{\theta-1}_k \delta^{\frac{\theta}{\theta-1}}). 
\end{equation}
Thus by Fano's lemma or Lemma \ref{fano2}, taking $\delta=\sqrt{\frac{d}{n}}$ we have the result. 
\end{proof}

\begin{proof}[{\bf Proof of Theorem \ref{thm:SC1}}]
{\bf Proof of Privacy.} We first recall the following lemma:
\begin{lemma}\citep{feldman2022hiding}
For a domain $\mathcal{D}$, let $\mathcal{R}^{(i)}: f\times \mathcal{D}\rightarrow \mathcal{S}^{(i)}$ for $i \in [n]$ be a sequence of algorithms such that $\mathcal{R}^{(i)}(z_{1:i-1},\cdot)$ is a $(\varepsilon_0,\delta_0)$-DP local randomizer for all values of auxiliary inputs $z_{1:i-1}\in \mathcal{S}^{(1)}\times \cdots \times \mathcal{S}^{(i-1)}$. Let $\mathcal{A}_{\mathcal{S}}:\mathcal{D}^n \rightarrow \mathcal{S}^{(1)}\times \cdots \times \mathcal{S}^{(n)}$ be the algorithm that given a dataset $x_{1:n\in\mathcal{D}^n}$, sample a uniformly random permutation $\pi$, then sequentially computes $z_i = \mathcal{R}^{(i)}(z_{1:i-1},x_{\pi(i)})$ for $i\in [n]$, and the outputs $z_{1:n}$. Then for any $\delta \in [0,1]$ such that $\varepsilon_0\leqslant \log \left(\frac{n}{16\log (2/\delta)}\right)$, $\mathcal{A}_{\mathcal{S}}$ is $(\varepsilon, \delta+O(e^{\varepsilon}\delta_0 n))$-DP where $\varepsilon = O\left((1-e^{-\varepsilon_0})\cdot (\frac{\sqrt{e^{\varepsilon_0}\log (1/\delta)}}{\sqrt{n}}+\frac{e^{\varepsilon_0}}{n})\right)$.
\end{lemma}
We know that for each $x\in \mathcal{B}_t$, we have $\mathcal{R}(\Pi_C(\nabla f(w, x)))=\Pi_C(\nabla f(w, x))+\zeta_x$, with $\zeta_x\sim \mathcal{N}(0, \sigma^2_1)$ and $\sigma_1^2=\frac{8C^2\log \frac{1}{\delta_0}}{\varepsilon_0^2})$ is an $(\varepsilon_0, \delta_0)$-LDP randomizer. As we randomly shuffled the data in the beginning, thus, the algorithm will be $(\hat{\varepsilon},\hat{\delta}+O(e^{\hat{\varepsilon}}\delta_0 n))$-DP where $\hat{\varepsilon}=O\left((1-e^{-\varepsilon_0})\cdot (\frac{\sqrt{e^{\varepsilon_0}\log (1/\hat{\delta})}}{\sqrt{n}}+\frac{e^{\varepsilon_0}}{n})\right)$.

Now, assume that $\varepsilon_0\leqslant \frac{1}{2}$, then $\exists c_1>0$, s.t.,
\begin{equation*}
\begin{aligned}
\hat{\varepsilon}&\leqslant c_1(1-e^{-\varepsilon_0})\cdot \left(\frac{\sqrt{e^{\varepsilon_0}\log(1/\hat{\delta})}}{\sqrt{n}}+\frac{e^{\varepsilon_0}}{n}\right)\\
& \leqslant c_1\cdot\left( (e^{\varepsilon_0/2}-e^{-\varepsilon_0/2})\cdot\sqrt{\frac{\log(1/\hat{\delta})}{n}}  +\frac{e^{\varepsilon_0}  -1}{n}  \right)\\
& \leqslant c_1 \cdot\left( \left((1+\varepsilon_0)-(1-\frac{\varepsilon_0}{2}  )\right) \cdot \sqrt{\frac{\log(1/\hat{\delta})}{n}}+\frac{(1+2\varepsilon_0)-1}{n}\right)\\
& =c_1 \cdot \varepsilon_0\cdot \left( \frac{3}{2}\sqrt{\frac{\log(1/\hat{\delta})}{n}}+\frac{2}{n}\right).
\end{aligned}
\end{equation*}
Set $\hat{\delta}=\frac{\delta}{2}$, $\delta_0 =c_2\cdot\frac{\delta}{e^{\hat{\varepsilon}}n}$ for some constant $c_2>0$ and replace $\varepsilon_0=\frac{2\sqrt{2}C \sqrt{\log \frac{1}{\delta_0}}}{\sigma_1})$:
\begin{equation*}
\begin{aligned}
\hat{\varepsilon}&\leqslant c_1\cdot\frac{2\sqrt{2}C \sqrt{\log \frac{1}{\delta_0}}}{\sigma_1})\cdot \left( \frac{3}{2}\sqrt{\frac{\log(1/\hat{\delta})}{n}}+\frac{2}{n}\right)\\
& \leqslant O\left(\frac{C \cdot \sqrt{\log(1/\delta)\log (e^{\hat{\varepsilon}}n/\delta)}}{\sigma_1 \sqrt{n}}\right).
\end{aligned}
\end{equation*}
For any $\varepsilon\leqslant 1$, if we set $\sigma =  O\left(\frac{C \cdot \sqrt{\log(1/\delta)\log (e^{\hat{\varepsilon}}n/\delta)}}{\varepsilon \sqrt{n}}\right)$, then we have $\hat{\varepsilon}\leqslant \varepsilon$. Furthermore, we need $\varepsilon_0 = \frac{2\sqrt{2}C \sqrt{\log \frac{1}{\delta_0}}}{\sigma_1}) \leqslant \frac{1}{2}$, which would be ensured if we set $\varepsilon = O\left(\sqrt{\frac{\log(n/\delta)}{n}}\right)$. This implies that for $\sigma_1 =  O\left(\frac{C \cdot \sqrt{\log(1/\delta)\log (e^{\hat{\varepsilon}}n/\delta)}}{\varepsilon \sqrt{n}}\right)$, algorithm \ref{alg:6} satisfies $(\varepsilon,\delta)$-DP as long as $\varepsilon=O\left(\sqrt{\frac{\log(n/\delta)}{n}}\right)$ if releasing $\mathcal{R}(\Pi_C(\nabla f(w, x)))$ for all $x$. Thus in step 6 we can see $\widetilde{\nabla} F_t\left(w_t^{m d}\right)=\frac{T}{n}\sum_{x\in \mathcal{B}_t}(\mathcal{R}(\Pi_C(\nabla f(w_t^{md}, x)))$ is $(\varepsilon, \delta)$-DP for each $t$. And since $\{B_t\}$ are disjoint, Algorithm \ref{alg:6} is $(\varepsilon, \delta)$-DP.

\begin{lemma}\label{lemma:5} 
    \citep{barber2014privacy} Let $\{z_i\}_{i=1}^s\sim\mathcal{D}^s$ be $\mathbb{R}^d$-valued random vectors with $\mathbb{E}z_i= \nu $ and $\mathbb{E}\left\| z_i \right\| ^k \leqslant r^{(k)}$ for some $k \geqslant 2$. Denote the noiseless average of clipped samples by $\widehat{\nu }:=\frac{1}{s}\sum\limits_{i=1}^{s} \prod_{C}^{}(z_i)$ and $\widetilde{\nu }:=\widehat{\nu }+N$. Then, % $\left\| \mathbb{E}\widetilde{\nu }-\nu  \right\|= \left\| \mathbb{E}\widehat{\nu }-\nu  \right\| \leqslant \mathbb{E}\left\| \widehat{\nu }-\nu  \right\| $ 
    $\lVert\mathbb{E}\widetilde{\nu}-\nu\rVert=\lVert\mathbb{E}\widehat{\nu}-\nu\rVert\leqslant\mathbb{E}\lVert\widehat{\nu}-\nu\rVert\leqslant \frac{r^{(k)}}{(k-1)C^{k-1}}$, and $\mathbb{E}\|\widetilde{\nu}-\mathbb{E}\widetilde{\nu}\|^{2}=\mathbb{E}\|\widetilde{\nu}-\mathbb{E}\widehat{\nu}\|^{2}\leqslant d\sigma^{2}+\frac{r^{(2)}}{s}.$ 
\end{lemma}

    claim: we can improve the noise to $\Sigma^2:=\sup_{t\in[T]}\mathbb{E}[\|N_t\|^2]\leqslant d\sigma^2+\frac{r^2T}{n}\approx \frac{dC^2T}{\varepsilon^2n^2}+\frac{r^2T}n$. 

Excess risk: Consider round $t \in[T]$ of Algorithm \ref{alg:6}, where Algorithm \ref{alg:1} is run on input data $\left\{\nabla f\left(w_t, x_i^t\right)\right\}_{i=1}^{n / T}$. Denote the bias of Algorithm 1 by $b_t:=\mathbb{E} \widetilde{\nabla} F_t\left(w_t\right)-\nabla F\left(w_t\right)$, where $\widetilde{\nabla} F_t\left(w_t\right)=\widetilde{\nu}$ in the notation of Algorithm 1. Also let $\hat{\nabla} F_t\left(w_t\right):=\hat{\mu}$ (in the notation of Lemma \ref{lemma:5}) and denote the noise by $N_t=\widetilde{\nabla} F_t\left(w_t\right)-\nabla F\left(w_t\right)-b_t=\widetilde{\nabla} F_t\left(w_t\right)-\mathbb{E} \widetilde{\nabla} F_t\left(w_t\right)$. Then we have $B:=\sup _{t \in[T]}\left\|b_t\right\| \leqslant \frac{r^{(k)}}{(k-1) C^{k-1}}$ and $\Sigma^2:=\sup _{t \in[T]} \mathbb{E}\left[\left\|N_t\right\|^2\right] \leqslant d \sigma^2+\frac{r^2 T}{n} \leqslant O\left(\frac{d C^2 T}{\varepsilon^2 n^2}+\frac{r^2 T}{n}\right)$, by Lemma 5. Plugging these estimates for $B$ and $\Sigma^2$ into Proposition 40 of \citep{lowy2023private} and setting $C=r\left(\frac{\varepsilon n}{\sqrt{d \log_{} (1/\delta )}}\right)^{1 / k}$, we get
$$
\begin{aligned}
\mathbb{E} F\left(w_T^{a g}\right)-F^* & \leqslant O\left( \frac{\beta D^2}{T^2}+\frac{D(\Sigma+B)}{\sqrt{T}}+B D\right) \\
& \leqslant O\left( \frac{\beta D^2}{T^2}+\frac{C D \sqrt{d\log_{} (1/\delta )}}{\varepsilon n}+\frac{r D}{\sqrt{n}}+\frac{r^{(k)} D}{C^{k-1}} \right)\\
& \leqslant O\left( \frac{\beta D^2}{T^2}+r D\left[\frac{1}{\sqrt{n}}+\left(\frac{\sqrt{d\log_{} (1/\delta ) }}{\varepsilon n}\right)^{(k-1) / k}\right]\right).
\end{aligned}
$$

Now, our choice of $T$
\begin{equation}\nonumber
        T = \min \{\sqrt[]{\frac{\beta D}{r}}\cdot \left(\frac{\varepsilon n}{\sqrt[]{d \log_{} (1/\delta )}}\right)^{\frac{k-1}{2k}}\!,\ \sqrt[]{\frac{\beta D}{r}}\cdot n^{1/4} \},
\end{equation}
    implies that $\frac{\beta D^2}{T^2} \leqslant r D\left[\frac{1}{\sqrt{n}}+\left(\frac{\sqrt{d\log_{} (1/\delta ) }}{\varepsilon n}\right)^{(k-1) / k}\right]$ and we get the result upon plugging in $T$.
\end{proof}

\begin{proof}[{\bf Proof of Theorem \ref{thm:SC2}}]
    Similar to the proof of Theorem \ref{thm:3}.
\end{proof}

\end{document}